\documentclass{article}

\usepackage{microtype}
\usepackage{graphicx}
\usepackage{subcaption}
\usepackage{booktabs} 

\usepackage{hyperref}
\usepackage{xurl}

\usepackage[accepted]{icml2026}

\usepackage{amsmath}
\usepackage{amssymb}
\usepackage{mathtools}
\usepackage{amsthm}

\usepackage[capitalize,noabbrev]{cleveref}

\usepackage{soul}
\sethlcolor{yellow}

\usepackage{array}
\newcolumntype{C}[1]{>{\centering\arraybackslash}m{#1}}
\usepackage{multirow}
\usepackage[table]{xcolor}
\usepackage{pgfplotstable}
\usepackage{tabularx}

\newcolumntype{Y}{>{\centering\arraybackslash}X}

\definecolor{headerbg1}{RGB}{222,228,240}
\definecolor{headerbg2}{RGB}{210,230,250}
\definecolor{groupbg}{RGB}{242,246,252}   
\definecolor{bestbg}{HTML}{CFE2FF}    
\definecolor{secondbg}{HTML}{E8F1FF}  

\DeclareRobustCommand{\shadeword}[2]{%
  \begingroup
  \setlength{\fboxsep}{0.2ex}
  \setlength{\fboxrule}{0pt}%
  \raisebox{0pt}[\ht\strutbox][\dp\strutbox]{%
    \colorbox{#1}{\kern0.3em #2\kern0.3em}%
  }%
  \endgroup
}

\hypersetup{
    colorlinks=true,
    linkcolor=red,
    citecolor=cyan,
    filecolor=magenta,      
    urlcolor=magenta,
    }

\usepackage{xspace}
\DeclareRobustCommand{\ssibench}{{\ttfamily\bfseries SSI-Bench}\xspace}

\theoremstyle{plain}

\theoremstyle{definition}

\theoremstyle{remark}

\usepackage[textsize=tiny]{todonotes}

\icmltitlerunning{Thinking in Structures: Evaluating Spatial Intelligence in Constraint-Governed Spaces}

\begin{document}

\twocolumn[
  \icmltitle{Thinking in Structures: Evaluating Spatial Intelligence \\ in Constraint-Governed Spaces}



  \icmlsetsymbol{equal}{*}

  \begin{icmlauthorlist}
    \icmlauthor{Chen Yang}{thu}
    \icmlauthor{Guanxin Lin}{thu}
    \icmlauthor{Youquan He}{thu}
    \icmlauthor{Peiyao Chen}{thu}
    \icmlauthor{Guanghe Liu}{thu}
    \icmlauthor{Yufan Mo}{thu}
    \icmlauthor{Zhouyuan Xu}{thu}
    \icmlauthor{Linhao Wang}{thu}
    \icmlauthor{Guohui Zhang}{thu}
    \icmlauthor{Zihang Zhang}{thu}
    \icmlauthor{Shenxiang Zeng}{thu}
    \icmlauthor{Chen Wang}{thu}
    \icmlauthor{Jiansheng Fan}{thu}
  \end{icmlauthorlist}

  \icmlaffiliation{thu}{Tsinghua University}

  \icmlcorrespondingauthor{Chen Wang}{chwang@tsinghua.edu.cn}
  \icmlcorrespondingauthor{Jiansheng Fan}{fanjsh@tsinghua.edu.cn}

  \icmlkeywords{Machine Learning, ICML}
    
    \vspace{-0.125in}
    \begin{center}
      \resizebox{\linewidth}{!}{
        \includegraphics{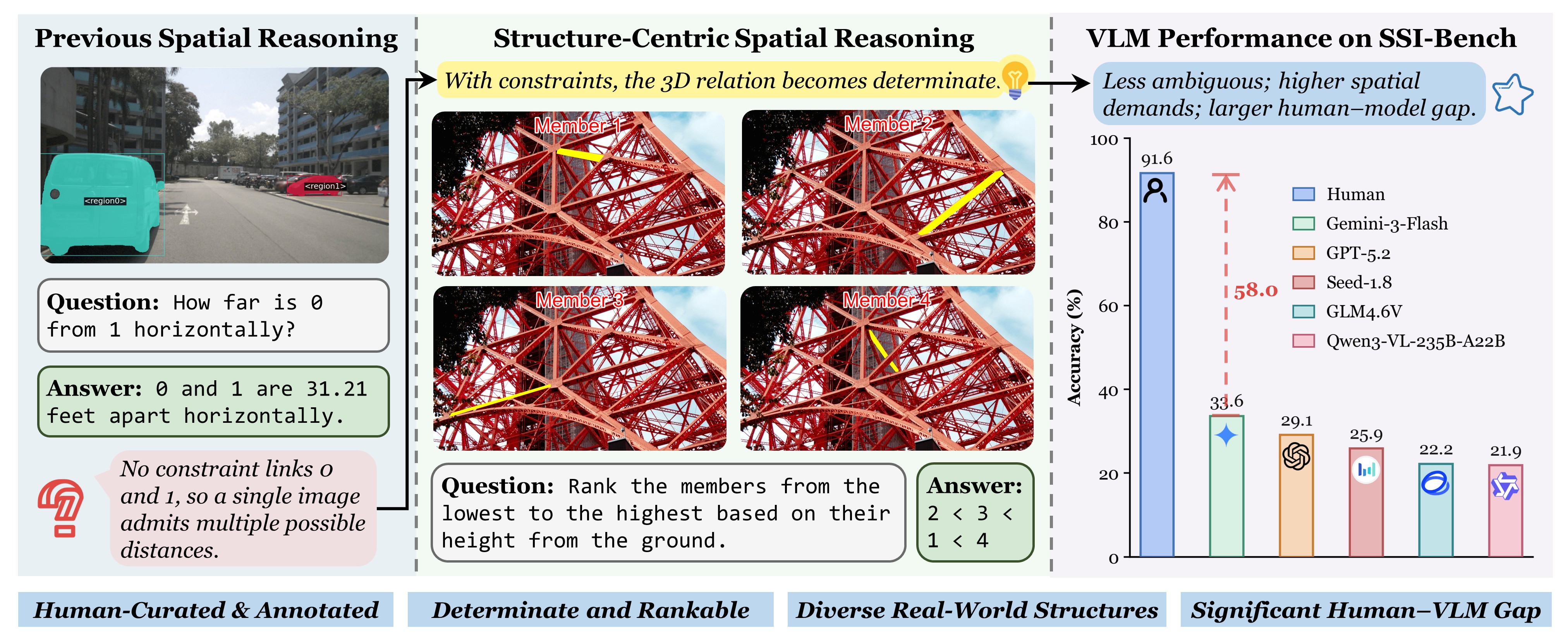}
        }
        \captionof{figure}{\ssibench is a VQA benchmark designed to evaluate models’ spatial reasoning under structural constraints on complex real-world 3D scenes. The bar chart illustrates the significant performance gap between state-of-the-art VLMs and human performance.}
    	\label{fig:teaser}
    \end{center}
    \vspace{0.1in}
]



\printAffiliationsAndNotice{}  

\begin{abstract}
Spatial intelligence is crucial for vision--language models (VLMs), yet many scene-centric benchmarks evaluate unconstrained environments where a single image may admit multiple plausible 3D interpretations. We introduce \ssibench, a VQA benchmark for \emph{Structure-Centric Spatial Reasoning (SCSR)} in constraint-governed spaces. Built from complex real-world 3D structures, it uses structural constraints from geometry, topology, and physical feasibility to make component relations more determinate from visual evidence. The benchmark contains 1,000 ranking questions spanning geometric and topological reasoning, where correct ordering requires resolving all candidate-wise 3D relations, imposing stronger demands on spatial understanding. It is created through a fully human-centered pipeline with over 400 researcher-hours of image curation, component annotation, and question design. Evaluating 31 VLMs reveals a large gap to humans: the best open-source model achieves $22.2\%$ accuracy and the strongest closed-source model reaches $33.6\%$, while humans score $91.6\%$. Further results show that chain-of-thought reasoning brings only marginal gains, and error analysis reveals fundamental limitations in current models' spatial understanding within constraint-governed spaces. Project page: \url{https://ssi-bench.github.io}.
\end{abstract}

\vspace{-0.2in}

\section{Introduction}

Vision--language models (VLMs) have made rapid progress in multimodal understanding and reasoning \cite{vteam2026glm45vglm41vthinkingversatilemultimodal, bai2025qwen3vltechnicalreport, li2025benchmark, liu2024improved}. For real-world deployment, however, a central open problem is spatial intelligence: inferring 3D relations and latent structure from visual input and using them to answer geometric and relational queries \cite{yang2025thinking, yin2025spatial}. Recent benchmarks have expanded along multiple axes, including single-view vs.\ multi-view inference \cite{cheng2024spatialrgpt, yang2025mmsi}, images vs.\ videos \cite{ma20253dsrbench, lin2025ost}, and automated vs.\ human annotation \cite{li2025sti, lin2025mmsi}. These efforts have been valuable for measuring progress, but they provide limited resolution on constraint-governed spaces, where 3D configurations are restricted by structure and feasibility.

We study this regime through \emph{Structure-Centric Spatial Reasoning (SCSR)}: spatial reasoning in which the underlying 3D state is inferred from structural elements and restricted by structural constraints, such as geometric regularities, topological connectivity, and physical feasibility. This perspective reveals an evaluation gap in existing benchmarks. Most spatial benchmarks focus on scene-centric reasoning in largely unconstrained environments, such as indoor navigation and everyday object arrangements, where object configurations are weakly governed by feasibility constraints and can vary almost arbitrarily \cite{yang2025thinking}. In single-image settings, this makes many 3D relations underdetermined. For example, an object may appear smaller because it is physically smaller or because it is farther away. Multiple 3D configurations can therefore remain consistent with the same 2D observation, making strict relational queries depend on assumptions, appearance priors, or dataset-specific regularities rather than a uniquely recoverable 3D state.

To address this limitation, we introduce \ssibench (\textbf{S}tructure-centric \textbf{S}patial \textbf{I}ntelligence Benchmark, Figure~\ref{fig:teaser}), a VQA benchmark designed to evaluate SCSR on complex real-world 3D structures. Such structures instantiate constraint-governed spaces: their components follow geometric regularities and connectivity rules, while their realizability is further restricted by physics-based feasibility. These structural constraints reduce single-image ambiguity by making component relations more determinate from visual evidence, enabling precise ranking questions over 3D geometric and topological criteria. Solving these problems requires constraint-consistent 3D understanding under viewpoint variation, clutter, and self-occlusion, together with spatial operations such as mental rotation, cross-sectional inference, occlusion reasoning, and force-path reasoning \cite{slim20253dcompat++, collins2022abo, chen2021geoqa}.

Importantly, this benchmark is not designed to test domain-specific engineering expertise, but to serve as a structure-centric complement to existing scene-centric spatial benchmarks. Its images come from everyday photography sources and cover common real-world structures such as roofs, stairs, bridges, towers, frames, and railings, rather than specialized industrial blueprints. This focus isolates a core aspect of spatial intelligence that is often entangled with object semantics and layout priors in unconstrained scenes. Thus, it complements existing evaluations by testing whether models can recover and reason over coherent 3D structure from real-world visual observations.

Because structural constraints make candidate relations more determinate, ranking provides a well-defined way to evaluate whether models can recover the relative 3D relations required by an explicit geometric or topological criterion. The benchmark contains 1,000 multiple-choice ranking questions, each presenting 3 or 4 candidates, namely members or groups, and requiring selection of the correct permutation. Unlike binary or standard multiple-choice formats, correct ranking requires resolving all relative 3D relations among the candidates, imposing stronger demands on spatial understanding. We organize tasks into two families. Geometric tasks include Ground Height, Ground Angle, Dimension, Relative Distance, Area, and Volume, while Topological tasks evaluate graph-based relations such as hop distance and cycle length. We also include a Multi-View subset to test cross-view correspondence relative to a reference member.

Constructing a diverse real-world benchmark for SCSR is challenging because existing structural datasets rarely provide the explicit spatial metadata needed for automated question generation \cite{yang2025thinking, lin2025ost, li2025sti}. We therefore develop the benchmark through a fully human-centered pipeline. Ten researchers devoted over 400 hours to reviewing approximately 20,000 structure-related images from multiple sources and selecting over 2,000 candidates that collectively cover nearly all common structure forms, including but not limited to space frames \cite{xu2023analysis}, steel towers \cite{bezas2022design}, cable-stayed bridges \cite{qi2024first}, timber trusses \cite{vollmecke2025assessment}, reinforcement frameworks, and pipeline systems \cite{yang2022novel}. Candidate sets are curated so that the correct ordering is not reliably recoverable from simple 2D pixel rankings and remains unambiguous under the intended structural interpretation. Each question is further checked by independent reviewers to ensure unambiguity and appropriate challenge.

We evaluate 31 widely used VLMs on the benchmark. The best open-source model achieves $22.2\%$ accuracy and the strongest closed-source model reaches $33.6\%$, while humans score $91.6\%$. Chain-of-thought reasoning improves performance only marginally, and error analysis suggests that the gap is primarily driven by limitations in structural grounding and constraint-consistent 3D reasoning.

Overall, our contributions are threefold. First, we introduce \ssibench, a human-curated benchmark for SCSR in constraint-governed spaces via ranking-based geometric and topological tasks. Second, we evaluate 31 widely used VLMs and human performance, revealing a substantial gap. Third, we provide an error analysis that identifies dominant failure modes and suggests directions for improving structure-centric spatial reasoning.

\begin{table*}[t]
  \caption{Task taxonomy of structure-centric spatial reasoning in \ssibench.}
  \label{tab:ssi-taxonomy}
  \vspace{-4pt}
  \centering
  \setlength{\tabcolsep}{4pt}
  \renewcommand{\arraystretch}{1.08}
  {\fontsize{8pt}{9.2pt}\selectfont
  \begin{tabularx}{\textwidth}{@{}p{1.9cm} p{2.4cm} X p{1.6cm}@{}}
    \toprule
    \textbf{Category} & \textbf{Sub-Category} & \textbf{Criterion} & \textbf{Candidates} \\
    \midrule
    \multirow[t]{7}{*}{Geometric}
      & Ground Height     & Rank members by centroid height relative to the ground plane. & 4 members \\
      & Ground Angle      & Rank members by the angle between their principal direction and the ground plane. & 4 members \\
      & Dimension         & Rank members by their length along the principal direction. & 4 members \\
      & Relative Distance & Rank member groups by the minimum distance between their principal-axis lines. & 3 groups \\
      & Area              & Rank node groups by planar convex-hull area in the reference plane. & 3 groups \\
      & Volume            & Rank node groups by 3D convex-hull volume enclosing the nodes. & 3 groups \\
      & Multi-View        & Fuse two views to rank target members by geometric relations to a reference member. & 3 groups \\
    \midrule
    \multirow[t]{3}{*}{Topological}
      & Hop Distance      & Rank member groups by shortest-path hop count on the structural connectivity graph. & 3 groups \\
      & Cycle Length      & Rank member groups by the minimum cycle length with the specified members. & 3 groups \\
      & Multi-View        & Fuse two views to rank target members by topological relations to a reference member. & 3 groups \\
    \bottomrule
  \end{tabularx}
  }
  \vskip -0.1in
\end{table*}

\begin{figure}[H]
  \centering
  \includegraphics[width=0.7\linewidth]{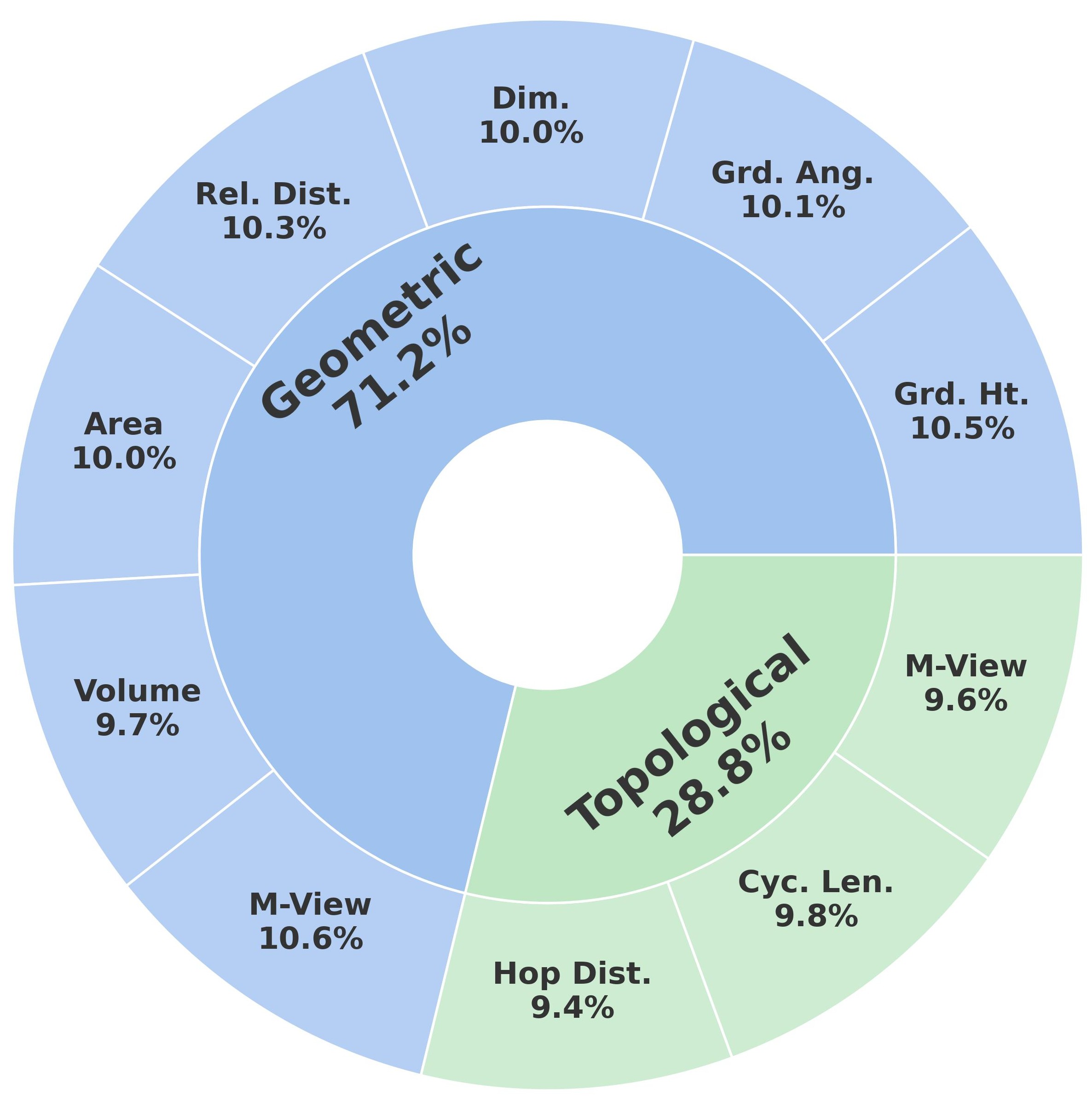}
  \caption{Distribution of task categories in \ssibench.}
  \label{fig:category_distribution}
  \vskip -0.15in
\end{figure}

\section{Related Work}

\textbf{Spatial Intelligence Benchmarks.}
Benchmarks for spatial intelligence in VLMs have developed along several complementary directions. Early benchmarks are primarily single-view and image-based, emphasizing local metric cues such as depth and distance, \emph{e.g.}, SpatialRGPT \cite{cheng2024spatialrgpt} and SpatialVLM \cite{chen2024spatialvlm}. Video-based benchmarks extend evaluation to spatio-temporal understanding with object--object and object--camera relations, including VSI-Bench \cite{yang2025thinking}, SPAR-Bench \cite{zhang2025flatland}, OST-Bench \cite{lin2025ost}, and MMSI-Video-Bench \cite{lin2025mmsi}. More recent efforts increasingly emphasize multi-view or multi-image inference to recover 3D structure, as in MMSI-Bench \cite{yang2025mmsi}, ViewSpatial-Bench \cite{li2025viewspatial}, and MindCube \cite{yin2025spatial}. Another growing line targets dynamic spatial intelligence by stressing motion, trajectories, and evolving instance states, exemplified by STI-Bench \cite{li2025sti} and DSI-Bench \cite{zhang2025dsi}.
Beyond modality and task format, spatial benchmarks can also be viewed along an axis between ecological breadth and reasoning purity. At one end, benchmarks such as VSI-Bench and MMSI-Bench target broad spatial needs in everyday scenes and are closely aligned with embodied AI applications. At the other end, benchmarks such as Spatial457 \cite{wang2025spatial457} and parts of OmniSpatial \cite{jia2025omnispatial} emphasize more diagnostic spatial reasoning with simplified shapes or synthetic objects. Despite this progress, most benchmarks either focus on unconstrained scene-centric settings, where single-image 3D relations can be underdetermined, or on abstract settings that reduce real-world visual complexity.
This leaves a complementary gap for benchmarks that retain real-world visual complexity while making 3D relations more determinate through geometric, topological, and physical constraints. Such benchmarks can support precise ranking-based evaluation of structure-centric spatial reasoning in constraint-governed spaces. Detailed quantitative comparisons are provided in Appendix~\ref{app:ssi_comparison}.

\textbf{Structural Understanding Benchmarks.}
Structural understanding is commonly evaluated via benchmarks that emphasize part structure, precise geometry, or explicit geometric reasoning. For part-centric structure, PartNet~\cite{mo2019partnet} provides large-scale 3D objects with fine-grained hierarchical part annotations. 3DCoMPaT++~\cite{slim20253dcompat++} further supports part-based and compositional learning with part-instance-level supervision in a multimodal setting. For geometry-centric structure, ABC \cite{koch2019abc} offers large CAD models with analytic representations and rich ground truth, and datasets such as ABO \cite{collins2022abo} provide real product assets for studying object geometry with multimodal signals. Beyond object datasets, GeoQA \cite{chen2021geoqa} and VQA benchmarks such as CLEVR \cite{johnson2017clevr} and GQA \cite{hudson2019gqa} evaluate structural reasoning through diagrammatic geometry problems or structured compositional queries, where CLEVR is synthetic and GQA is grounded in real images. Compared with these benchmarks, which often assess structure understanding explicitly via part labels or geometric outputs, our benchmark targets structural understanding implicitly from the perspective of spatial intelligence: models are not asked to output structure directly, but to answer spatial-relation queries that require recovering the underlying 3D structure. It further focuses on real-world 3D structures, introducing richer geometric complexity, occlusion, clutter, and viewpoint variation than curated or simulated settings.

\begin{figure*}[t]
  \centering
  \includegraphics[width=\textwidth]{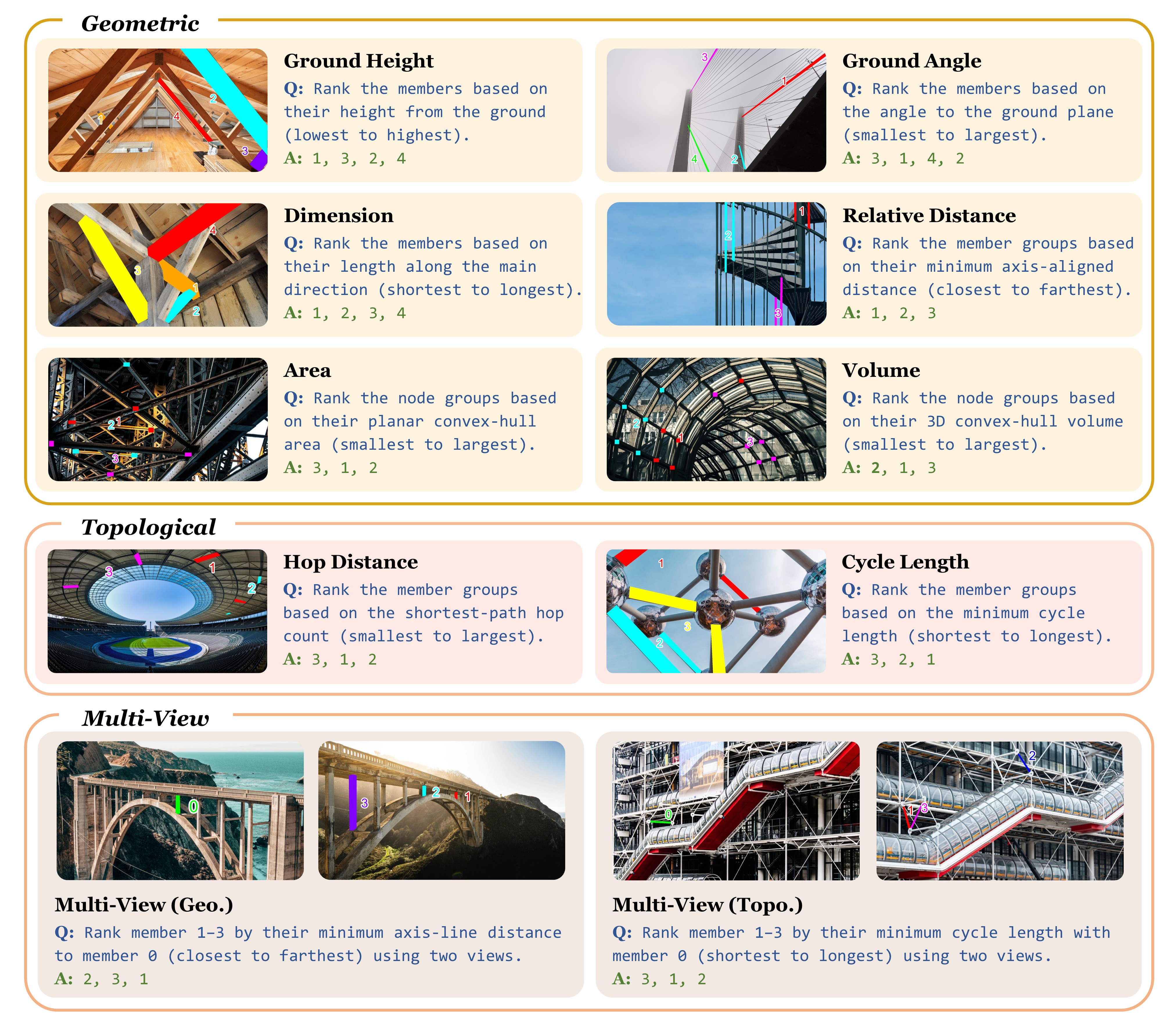}
  \caption{Representative \ssibench samples from each category. For visualization, we overlay all candidates in one image; the benchmark provides separately annotated images per option. Ties use the smaller index first. Full questions are in Appendix~\ref{app:samples}.}
  \label{fig:samples}
  \vskip -0.05in
\end{figure*}

\section{\ssibench}

This section introduces \ssibench, a benchmark for evaluating spatial intelligence under structural constraints. In this regime, solving visual questions requires recovering an underlying 3D geometric and relational structure that is consistent with strong feasibility constraints. We begin by formalizing SCSR in Section~\ref{subsec:SCSR_formal}, then outline the benchmark design and task taxonomy in Section~\ref{overview}. Section~\ref{construction} details the benchmark construction pipeline.

\begin{figure*}[t]
  \centering
  \includegraphics[width=\textwidth]{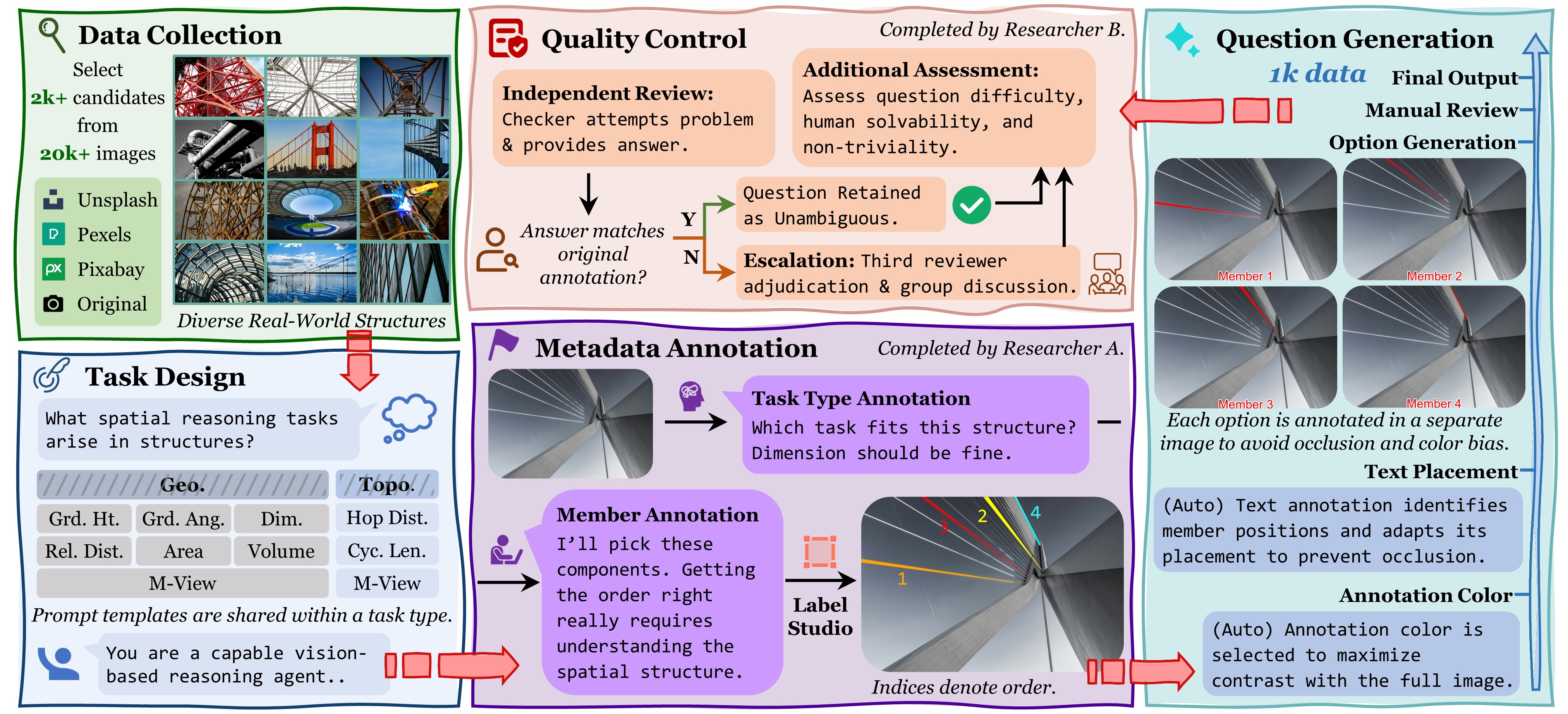}
  \caption{Illustration of the \ssibench construction pipeline.}
  \label{fig:construction}
  \vskip -0.05in
\end{figure*}

\subsection{Problem Formulation}
\label{subsec:SCSR_formal}

SCSR studies spatial reasoning in which the latent 3D state is restricted to a feasible set defined by explicit constraints. We represent a structural scene as
\begin{equation}
\mathbf{s} \;=\; (V, E, \mathbf{G}, \mathbf{A}),
\end{equation}
where $V$ and $E$ denote nodes and members, with connectivity graph $\mathcal{G}=(V,E)$. The variable $\mathbf{G}$ collects geometric degrees of freedom (\emph{e.g.}, node coordinates, member directions/lengths, or other latent 3D parameters), and $\mathbf{A}$ captures discrete attributes (\emph{e.g.}, component types or grouping metadata). Feasibility constraints restrict admissible states to a constrained manifold---or more generally, a constrained feasible set---given by
\begin{equation}
\mathcal{M} \;=\; \Bigl\{\mathbf{s}\;:\;\mathbf{c}(\mathbf{s})=\mathbf{0},\;\mathbf{h}(\mathbf{s})\le \mathbf{0}\Bigr\},
\label{eq:SCSR_manifold}
\end{equation}
where $\mathbf{c}$ encodes equality constraints (\emph{e.g.}, geometric compatibility or connectivity-induced relations) and $\mathbf{h}$ encodes inequality constraints (\emph{e.g.}, non-intersection, support conditions, or physics-based feasibility). Given an observation $x$ (one or more images), SCSR focuses on inferring relations that remain consistent with both the evidence in $x$ and the feasible set $\mathcal{M}$, rather than relying on unconstrained 2D correlations.

\textbf{Instantiation of structural constraints.}
In \ssibench, the structural constraints in Eq.~(\ref{eq:SCSR_manifold}) are instantiated by three categories: 
\emph{geometric regularity}, \emph{topological and connectivity constraints}, and \emph{physical and semantic feasibility}. 
Geometric regularity and connectivity mainly define equality or discrete compatibility constraints in $\mathbf{c}(\mathbf{s})$ and the graph structure $\mathcal{G}$, while physical and semantic feasibility is naturally captured by inequality constraints in $\mathbf{h}(\mathbf{s})$. 
These constraints are not provided to models as symbolic inputs; instead, they are implicit in the visual scene and are used during benchmark construction to ensure that the queried 3D relations are well-defined and visually inferable. 
Further details are provided in Appendix~\ref{app:structural_constraints}.

\textbf{Ranking formulation.}
Each \ssibench instance specifies a task type $\tau$ and a candidate set $\mathcal{C}=\{c_i\}_{i=1}^K$ (members or groups, with $K\in\{3,4\}$), together with a criterion function $f_\tau(\mathbf{s},c)$ that assigns a task-relevant value to each candidate. The ground-truth answer is the permutation $\pi^\star$ that orders candidates by this task-defined value:
\begin{equation}
\pi^\star \;=\; \operatorname*{argsort}_{\pi \in S_K}\;\Bigl(f_\tau(\mathbf{s},c_{\pi(1)}),\,\ldots,\,f_\tau(\mathbf{s},c_{\pi(K)})\Bigr).
\label{eq:SCSR_ranking}
\end{equation}

\subsection{Overview of \ssibench}
\label{overview}

Real-world spatial reasoning often hinges on structure. 
Geometric regularities and connectivity rules restrict how components can be arranged, making structural scenes representative constraint-governed spaces. 
\ssibench is designed to evaluate SCSR in such spaces.

\ssibench adopts a structural view of each scene. 
Each instance is defined by nodes and members as primitives, their geometric attributes, and their topological organization captured by a connectivity graph. 
Building on this representation, we design two task families---Geometric and Topological---and instantiate them as ranking questions over members or groups. 
We also include a Multi-View setting that fuses two views to infer relations relative to a reference member and to test cross-view structural consistency. 
Table~\ref{tab:ssi-taxonomy} and Figure~\ref{fig:category_distribution} provide a compact overview of the task taxonomy and category distribution.
Representative examples from all categories are shown in Figure~\ref{fig:samples}.

The benchmark contains 1{,}000 multiple-choice ranking questions. 
Each question presents 3 or 4 candidates and asks for the correct permutation under a specified criterion. 
Every instance includes a detailed prompt and localization annotations for the referenced members or groups, and is curated to have a unique correct answer with deterministic tie-breaking when necessary. 
More detailed dataset statistics are reported in Appendix~\ref{app:data_statistics}.

\subsection{Benchmark Construction Process}
\label{construction}

Figure~\ref{fig:construction} summarizes our benchmark construction pipeline. \ssibench is built through a fully human-centered process: ten researchers with interdisciplinary expertise in AI and structural engineering devoted over 400 hours to its construction. Additional details are provided in Appendix~\ref{app:ssi_constr_details}.

\textbf{Data collection.} To target spatial reasoning under structural constraints, we curate real-world images of 3D structures with strong geometric and topological constraints. Most images come from three royalty-free sources---Unsplash \cite{unsplash-website}, Pexels \cite{pexels-website}, and Pixabay \cite{pixabay-website}---and are filtered for diversity in structure types, scenes, illumination, and viewpoints. After reviewing roughly 20{,}000 structure-related images, we retain over 2{,}000 candidates. Because multi-view structural imagery is scarce, part of the Multi-View subset is sourced from our own photography. All selected images are high-resolution and uniformly compressed so that the longer side is at most 1920 pixels.

\textbf{Task design.} We define ten task categories across two families, Geometric and Topological, guided by the spatial judgments that naturally arise in complex structures under geometric and connectivity constraints. Ground Height, Ground Angle, and Dimension are member-level tasks with four candidates, while the remaining tasks are group-level with three candidates. The Multi-View subset provides two images per question: one highlights a reference member (Member 0) and the other highlights target members, requiring cross-view correspondence. For each task, we use a task-specific prompt template aligned with its criterion and generate options by permuting candidates into multiple-choice permutations. When ties occur, we apply a consistent ordering rule (placing the smaller index first) to avoid ambiguity.

\textbf{Metadata annotation.} We annotate the information needed for ranking using Label Studio. Annotators record an ascending order under the specified criterion and explicitly mark ties. We also provide localization annotations for referenced members or groups. Highlighting polygons are drawn to tightly fit target components while respecting occlusion. Candidate sets are curated so that correct ordering is not reliably recoverable from 2D pixel layouts alone, but instead requires reasoning about the underlying 3D geometry and topology constraints.

\textbf{Question generation.} Using the annotated metadata, we instantiate ranking questions with the corresponding images and prompts. To reduce occlusion and color-induced bias, we provide a separately annotated image for each option rather than marking all options in a single image. Highlight colors are automatically selected from a predefined palette to maintain contrast, and text labels are placed near targets with adaptive positioning and then verified by human reviewers for legibility. This process yields 1{,}000 questions. The metadata annotation and question generation stages can be extended to support semi-automatic data scaling by expanding candidate annotations and sampling new ranking questions, as detailed in Appendix~\ref{app:semi_auto_expansion}.

\textbf{Quality control.} Each question is independently reviewed by a checker who attempts the problem. Questions with disagreements are escalated to a third reviewer for adjudication. Reviewers also assess whether questions are human-solvable yet challenging, and whether at least one option is non-trivial (i.e., not inferable from superficial 2D cues). We additionally assign a difficulty label to every question. Further details are provided in Appendix~\ref{app:difficulty_annotation}.

\begin{table*}[t]
    \caption{Evaluation on \ssibench (Taskwise Accuracy). We highlight the \shadeword{bestbg}{best} and \shadeword{secondbg}{second best} results within each category (Proprietary or Open-source Models). Pairwise Accuracy results are reported in Appendix~\ref{app:pairwise_results}. Abbreviations: Grd. Ht. = Ground Height; Grd. Ang. = Ground Angle; Dim. = Dimension; Rel. Dist. = Relative Distance; Area = Area; Volume = Volume; M-View = Multi-View; Hop Dist. = Hop Distance; Cyc. Len = Cycle Length.}
  \label{tab:geom-topo}
  \vspace{-9pt}
  \setlength{\fboxsep}{1.45pt}

  \setlength{\tabcolsep}{0pt}
  \renewcommand{\arraystretch}{1}

  \begin{center}
    \begin{footnotesize}
      \begin{sc}
        \begin{tabular}{ l *{11}{>{\centering\arraybackslash}p{0.07\textwidth}} }
        \toprule
        \multirow{2}{*}[-0.5ex]{\textbf{Models}} & \multicolumn{7}{c}{\textbf{Geometric}} & \multicolumn{3}{c}{\textbf{Topological}} & \multirow{2}{*}[-0.5ex]{\textbf{Avg.}} \\
        \cmidrule(lr){2-8} \cmidrule(lr){9-11} & {\scriptsize\normalfont\textbf{Grd. Ht.}} & {\scriptsize\normalfont\textbf{Grd. Ang.}} & {\scriptsize\normalfont\textbf{Dim.}} & {\scriptsize\normalfont\textbf{Rel. Dist.}} & {\scriptsize\normalfont\textbf{Area}} & {\scriptsize\normalfont\textbf{Volume}} & {\scriptsize\normalfont\textbf{M-View}} & {\scriptsize\normalfont\textbf{Hop Dist.}} & {\scriptsize\normalfont\textbf{Cyc. Len.}} & {\scriptsize\normalfont\textbf{M-View}} & \\
        \midrule

          \rowcolor{groupbg}
          \multicolumn{12}{l}{\textit{Proprietary Models}} \\
          Gemini-3-Pro        & 25.71 & \colorbox{secondbg}{37.62} & 28.00 & 33.01 & 24.00 & \colorbox{bestbg}{27.84} & \colorbox{secondbg}{31.13} & \colorbox{bestbg}{35.11} & 30.61 & 21.88 & \colorbox{secondbg}{29.50} \\
          Gemini-3-Flash      & \colorbox{bestbg}{37.14} & \colorbox{bestbg}{38.61} & \colorbox{bestbg}{35.00} & \colorbox{secondbg}{41.75} & 27.00 & 25.77 & \colorbox{bestbg}{33.96} & \colorbox{secondbg}{32.98} & \colorbox{bestbg}{34.69} & 28.13 & \colorbox{bestbg}{33.60} \\
          Gemini-2.5-Pro      & 20.95 & 31.68 & 23.00 & 33.98 & 19.00 & 22.68 & \colorbox{secondbg}{31.13} & 22.34 & 27.55 & 28.13 & 26.10 \\
          Gemini-2.5-Flash    & 20.00 & 24.75 & 21.00 & 21.36 & 23.00 & 12.37 & 26.42 & 26.60 & 24.49 & 22.92 & 22.30 \\
          GPT-5.2             & \colorbox{secondbg}{29.52} & 30.69 & \colorbox{secondbg}{32.00} & \colorbox{secondbg}{41.75} & \colorbox{bestbg}{30.00} & 21.65 & 29.25 & 24.47 & 30.61 & 19.79 & 29.10 \\
          GPT-5 mini          & 19.05 & 30.69 & 29.00 & \colorbox{bestbg}{43.69} & 26.00 & 17.53 & 20.75 & 21.28 & 19.39 & \colorbox{bestbg}{31.25} & 25.90 \\
          GPT-4.1             & 17.14 & 16.83 & 25.00 & 21.36 & \colorbox{bestbg}{30.00} & 22.68 & 16.04 & 23.40 & 25.51 & 27.08 & 22.40 \\
          GPT-4o              & 19.05 & 20.79 & 20.00 & 26.21 & \colorbox{secondbg}{29.00} & \colorbox{secondbg}{26.80} & 17.92 & 17.02 & 22.45 & 27.08 & 22.60 \\
          Claude-Sonnet-4.5   & 8.57  & 12.87 & 16.00 & 30.10 & 21.00 & 19.59 & 24.53 & 20.21 & 19.39 & 27.08 & 19.90 \\
          Seed-1.8            & 19.05 & 24.75 & 22.00 & 37.86 & 19.00 & 24.74 & 24.53 & 24.47 & \colorbox{secondbg}{33.67} & \colorbox{secondbg}{29.17} & 25.90 \\

          \midrule
        \rowcolor{groupbg}\multicolumn{12}{l}{\textit{Open-source Models}} \\
        GLM-4.6V & 9.52 & \colorbox{secondbg}{21.78} & 16.00 & \colorbox{secondbg}{30.10} & 28.00 & \colorbox{secondbg}{21.65} & \colorbox{bestbg}{26.42} & 25.53 & 20.41 & 22.92 & \colorbox{bestbg}{22.20} \\
        GLM-4.6V-Flash & \colorbox{secondbg}{13.33} & 16.83 & 12.00 & 25.24 & 26.00 & \colorbox{bestbg}{25.77} & 16.98 & 26.60 & 21.43 & 28.13 & 21.10 \\
        GLM-4.5V & \colorbox{bestbg}{17.14} & \colorbox{bestbg}{23.76} & \colorbox{secondbg}{17.00} & 28.16 & 25.00 & 15.46 & 19.81 & 20.21 & 23.47 & 23.96 & 21.40 \\
        Qwen3-VL-235B-A22B & \colorbox{secondbg}{13.33} & 16.83 & \colorbox{bestbg}{22.00} & 26.21 & 26.00 & 20.62 & 21.70 & \colorbox{bestbg}{29.79} & 19.39 & 23.96 & \colorbox{secondbg}{21.90} \\
        Qwen3-VL-30B-A3B & 5.71 & 7.92 & 12.00 & 28.16 & 29.00 & 17.53 & 22.64 & \colorbox{secondbg}{28.72} & 28.57 & 27.08 & 20.60 \\
        Qwen3-VL-8B & 9.52 & 5.94 & 12.00 & 28.16 & 18.00 & 18.56 & 18.87 & 24.47 & \colorbox{secondbg}{29.59} & 28.13 & 19.20 \\
        Qwen3-VL-4B & 6.67 & 7.92 & 11.00 & \colorbox{bestbg}{33.01} & 29.00 & 16.49 & 20.75 & 25.53 & \colorbox{bestbg}{30.61} & 27.08 & 20.70 \\
        Qwen3-VL-2B & 5.71 & 7.92 & 11.00 & 28.16 & \colorbox{bestbg}{35.00} & 16.49 & 18.87 & \colorbox{secondbg}{28.72} & \colorbox{secondbg}{29.59} & 11.46 & 19.20 \\
        InternVL3.5-241B-A28B & 3.81 & 6.93 & 6.00 & 28.16 & 25.00 & 16.49 & 21.70 & \colorbox{secondbg}{28.72} & 23.47 & 23.96 & 18.30 \\
        InternVL3.5-30B-A3B & 6.67 & 7.92 & 12.00 & \colorbox{secondbg}{30.10} & 32.00 & 16.49 & 18.87 & 27.66 & 28.57 & 28.13 & 20.70 \\
        InternVL3.5-38B & 6.67 & 5.94 & 16.00 & 22.33 & 29.00 & 15.46 & 22.64 & 27.66 & 18.37 & 27.08 & 19.00 \\
        InternVL3.5-14B & 5.71 & 7.92 & 11.00 & 16.50 & 29.00 & 17.53 & 23.58 & 19.15 & 19.39 & \colorbox{bestbg}{30.21} & 17.90 \\
        InternVL3.5-8B & 5.71 & 6.93 & 10.00 & 24.27 & \colorbox{secondbg}{33.00} & 16.49 & \colorbox{secondbg}{24.53} & 25.53 & 28.57 & 28.13 & 20.20 \\
        InternVL3.5-4B & 8.57 & 6.93 & 7.00 & 27.18 & 28.00 & 14.43 & 15.09 & 17.02 & 24.49 & 19.79 & 16.80 \\
        InternVL3.5-2B & 4.76 & 3.96 & 11.00 & 22.33 & 25.00 & 17.53 & 4.72 & 11.70 & 10.20 & 0.00 & 11.10 \\
        Llama-4-Scout-17B-16E & 9.52 & 19.80 & \colorbox{secondbg}{17.00} & 20.39 & 29.00 & 16.49 & 21.70 & 21.28 & 23.47 & 28.13 & 20.60 \\
        Gemma-3-27B & 7.62 & 7.92 & 10.00 & 26.21 & \colorbox{secondbg}{33.00} & 16.49 & 22.64 & 26.60 & 28.57 & 27.08 & 20.50 \\
        Gemma-3-12B & 8.57 & 4.95 & 8.00 & 16.50 & 18.00 & 17.53 & 22.64 & 19.15 & \colorbox{secondbg}{29.59} & \colorbox{secondbg}{29.17} & 17.30 \\
        Gemma-3-4B & 6.67 & 6.93 & 12.00 & 22.33 & 32.00 & 18.56 & 23.58 & 26.60 & \colorbox{secondbg}{29.59} & 19.79 & 19.70 \\
        LLaVA-Onevision-72B & 4.76 & 6.93 & 4.00 & 20.39 & 31.00 & 16.49 & 23.58 & 19.15 & 23.47 & 22.92 & 17.20 \\
        LLaVA-Onevision-7B & 9.52 & 5.94 & 3.00 & 19.42 & 19.00 & 19.59 & \colorbox{secondbg}{24.53} & 23.40 & 22.45 & 18.75 & 16.50 \\

          \midrule
          \rowcolor{groupbg}
          \multicolumn{12}{l}{\textit{Baselines}} \\
          Random Guessing     & 4.17  & 4.17  & 4.17  & 16.67 & 16.67 & 16.67 & 16.67 & 16.67 & 16.67 & 16.67 & 12.85 \\
        \textbf{Human Performance}
        & \textbf{94.29} & \textbf{91.09} & \textbf{92.00} & \textbf{93.20}
        & \textbf{98.00} & \textbf{91.75} & \textbf{89.62}
        & \textbf{92.55} & \textbf{89.80} & \textbf{83.33} & \textbf{91.60} \\

          \bottomrule
        \end{tabular}
      \end{sc}
    \end{footnotesize}
  \end{center}
  \vskip -0.2in
\end{table*}

\section{Experiments and Analysis}

\subsection{Evaluation Settings}
\label{evel_setting}

\textbf{VLM evaluation.} We evaluate \ssibench on 31 VLMs, including 10 proprietary models from four families and 21 open-source models from six families. All models are run with temperature 0. Following Enact~\cite{wang2025enact}, input images are resized so that the longer side is at most 512 pixels, and we use a unified prompt template for each QA type. Models are instructed to output a parsable Python list encoding a permutation of indices. We report two complementary metrics: \emph{Taskwise Accuracy} (exact-match accuracy on the full permutation) and \emph{Pairwise Accuracy} (pairwise ordering consistency).

\textbf{Human evaluation.} We recruit six independent evaluators with basic structural knowledge who were not involved in data annotation. They answer the entire benchmark under the same instructions as the models. Their average performance serves as a proxy for human-level capability on \ssibench.

\textbf{Random baseline.} We report the expected accuracy of uniformly random ranking.
Implementation details and metric definitions are provided in Appendices~\ref{app:eval_metrics}--\ref{app:human_eval_setup}.

\begin{figure*}[t]
  \centering
  \includegraphics[width=\textwidth]{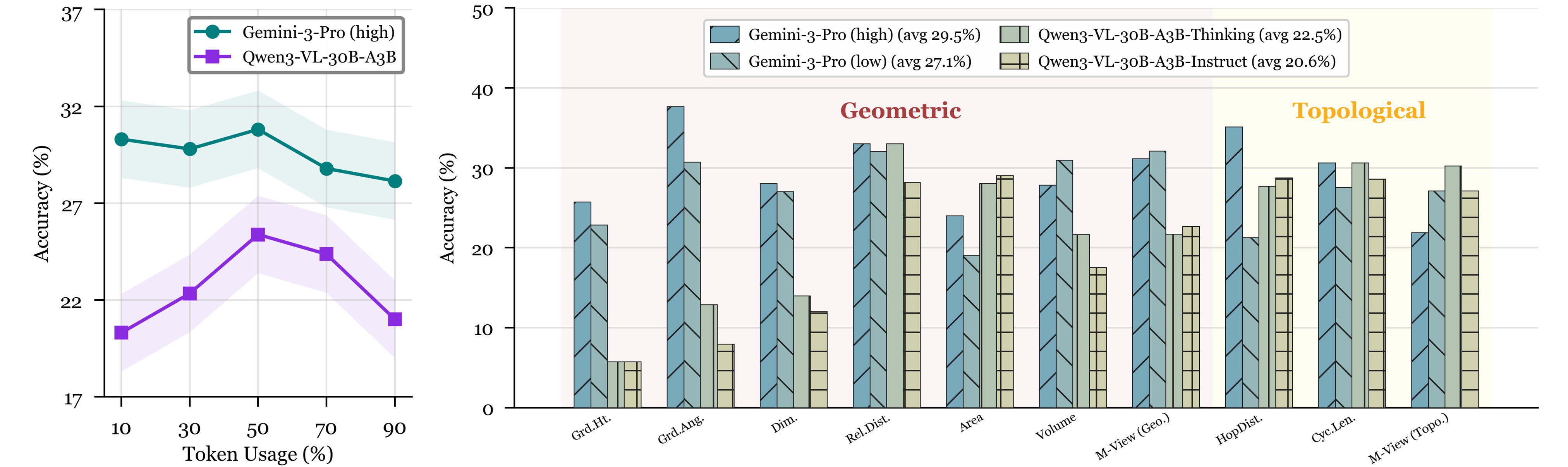}
  \caption{(\emph{Left}) Relationship between thinking-token usage and accuracy; (\emph{Right}) Sub-category level effects of thinking on SCSR.}
  \label{fig:think_anal}
  \vskip -0.05in
\end{figure*}

\subsection{Main Results}

Table~\ref{tab:geom-topo} presents the primary results on \ssibench in terms of Taskwise Accuracy. We summarize the main findings below. More results are shown in Appendices~\ref{app:generality_analysis} and \ref{app:pairwise_results}.

\textbf{Current VLMs struggle with SCSR.} \ssibench yields substantially lower accuracies than prior spatial benchmarks in comparable modalities but largely unconstrained settings, indicating that SCSR is harder and less amenable to 2D shortcut cues. Even strong VLMs remain far from human performance. The best proprietary model, Gemini-3-Flash, reaches $33.60\%$ average Taskwise Accuracy, and the best open-source model, GLM-4.6V, reaches $22.20\%$, while humans achieve $91.60\%$. The random-ranking baseline is $12.85\%$, and several models remain close to this level, suggesting that \ssibench cannot be solved reliably by weak heuristics and that robust SCSR is still missing in today’s VLMs.

\textbf{Advanced open-source models still trail proprietary counterparts.} A consistent gap appears between open-source and proprietary models across both geometric and topological tasks. Proprietary systems dominate the top of the leaderboard (\emph{e.g.}, Gemini-3-Flash at $33.60\%$, Gemini-3-Pro at $29.50\%$, and GPT-5.2 at $29.10\%$), whereas leading open-source models remain lower (\emph{e.g.}, GLM-4.6V at $22.20\%$ and Qwen3-VL-235B-A22B at $21.90\%$). This gap suggests that current open-weight models are less reliable at inferring the latent 3D state and applying 
structural constraints for correct rankings, while the absolute performance of proprietary models also indicates that SCSR remains broadly unsolved.

\textbf{Progress over time is visible but remains incremental, and scaling yields limited and inconsistent gains.} Across major model lineages, performance improves gradually over successive generations, but the gains are uneven and often modest. Within Gemini, average accuracy increases from Gemini-2.5-Flash ($22.30\%$) and Gemini-2.5-Pro ($26.10\%$) to Gemini-3-Pro ($29.50\%$) and Gemini-3-Flash ($33.60\%$). A similar pattern appears in the GPT series, where GPT-4.1 ($22.40\%$) and GPT-4o ($22.60\%$) improve to GPT-5 mini ($25.90\%$) and GPT-5.2 ($29.10\%$). In contrast, progress within GLM is smaller, with GLM-4.5V ($21.40\%$) and GLM-4.6V ($22.20\%$) remaining close. Larger or more recent variants do not always improve substantially within the same family, suggesting that scaling alone is insufficient to close the gap to human-level structural reasoning. These trends point to the need for improvements beyond scaling, such as stronger structure-aware training signals and better coverage of constrained 3D structural configurations.

\begin{figure*}[t]
  \centering
  \includegraphics[width=\textwidth]{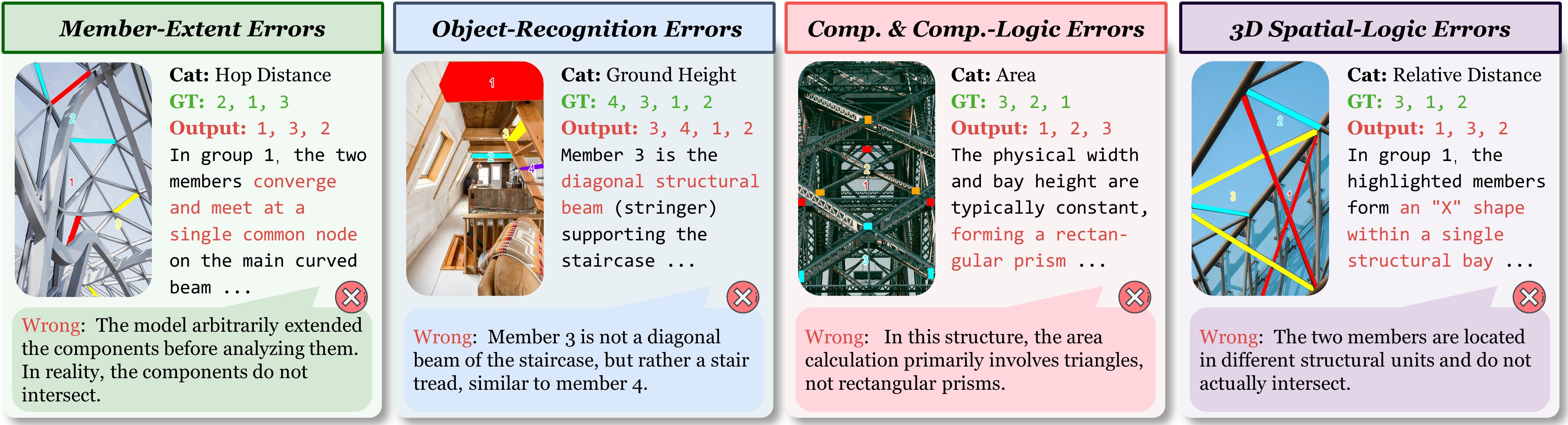}
  \caption{Illustration of four error types identified in VLM spatial reasoning on \ssibench.}
  \label{fig:error_anal}
  \vskip -0.05in
\end{figure*}

\subsection{Impact of Thinking on SCSR}
\label{sec:thinking_SCSR}

We study how explicit ``thinking'' affects SCSR by comparing two representative VLMs under the same evaluation protocol as Section~\ref{evel_setting}: Gemini-3-Pro with two thinking levels (\textsc{high} vs.\ \textsc{low}) and Qwen3-VL-30B-A3B with two variants (\textsc{Thinking} vs.\ \textsc{Instruct}). Each setting is evaluated on the full \ssibench benchmark. Figure~\ref{fig:think_anal} shows the results. Additional details are provided in Appendix~\ref{app:thinking_results}.

\textbf{Thinking improves performance, but only modestly.}
Stronger thinking consistently increases Taskwise Accuracy, yet the gains are small relative to the overall difficulty of SCSR. Gemini-3-Pro improves from $27.1\%$ (\textsc{low}) to $29.5\%$ (\textsc{high}), and Qwen3-VL-30B-A3B improves from $20.6\%$ (\textsc{Instruct}) to $22.5\%$ (\textsc{Thinking}). These results indicate that ``thinking'' provides incremental benefits rather than resolving the dominant failure modes on \ssibench.

\textbf{Token usage is a weak proxy for effective reasoning.}
We bin questions by token usage (fraction of the maximum thinking-token count) and compute accuracy per bucket (Figure~\ref{fig:think_anal}, \emph{Left}). For both models, accuracy is non-monotonic: it peaks at moderate usage and drops at very low or very high usage. This suggests that extra tokens often signal uncertainty rather than improved constraint-consistent inference. High usage frequently corresponds to prolonged deliberation over incorrect structural hypotheses (\emph{e.g.}, mislocalization or cross-view mismatch), which amplifies errors; the decline at the highest usage bucket is more pronounced for Qwen3-VL-30B-A3B, while Gemini-3-Pro remains comparatively stable.

\textbf{Benefits are task-dependent and can be negative on 3D-consistency bottlenecks.}
Per-task results (Figure~\ref{fig:think_anal}, \emph{Right}) show that thinking does not yield uniform gains. For Gemini-3-Pro, \textsc{high} thinking improves several single-view geometric/topological criteria (\emph{e.g.}, Ground Angle and HopDist.), but it is mixed---and sometimes worse---on tasks that hinge on globally consistent 3D reconstruction, particularly the Multi-View setting and Volume. Overall, these patterns suggest that additional deliberation helps mainly when the evidence supporting the task criterion is stable, but it can amplify errors when structural grounding or cross-view correspondence is uncertain, leading to longer reasoning over incorrect 3D hypotheses.

\subsection{Error Analysis}
\label{sec:error-analysis}

To diagnose the bottlenecks of current VLMs on SCSR, we perform an error analysis with Gemini-3-Pro as a representative model. Following the evaluation protocol in Section~\ref{evel_setting}, we collect its reasoning traces and randomly sample 100 questions from \ssibench for manual inspection. We identify four dominant failure modes; Figure~\ref{fig:error_anal} presents representative examples. More complete cases and analyses are provided in Appendix~\ref{app:additional_case}.

\textbf{Member-extent errors.}
The model frequently misestimates the spatial extent of a highlighted member, implicitly extending it beyond its true endpoints or truncating it to a visible fragment. This failure is particularly prevalent under occlusion and clutter: when only a portion of a member is visible, the model often treats the fragment as the entire component. Such extent errors directly corrupt geometric comparisons that rely on correct endpoints and principal directions, leading to unreliable rankings in tasks such as \textit{Dimension} and \textit{Relative Distance}.

\textbf{Object-recognition errors.}
We observe two recurring patterns. First, the model misidentifies components or nodes (\emph{e.g.}, confusing stair treads with diagonal braces), with errors increasing when targets are small, thin, or heavily occluded. Second, it misjudges coarse orientation, such as interpreting a slanted member as horizontal or a near-vertical member as perfectly vertical. These recognition and orientation failures most directly degrade \textit{Ground Angle}, and they also propagate to downstream comparisons that depend on correct component identity and orientation.

\textbf{Computational and comparison-logic errors.}
Even when the relevant components are localized correctly, the model may apply an incorrect comparison rule. For \textit{Area} and \textit{Volume}, it sometimes optimizes the wrong quantity (\emph{e.g.}, reasoning about 2D projected area when the criterion is 3D convex-hull volume) or adopts invalid simplifications (\emph{e.g.}, substituting an oblique height with an orthogonal height). For length-related criteria, the model occasionally falls back on coarse, experience-based heuristics rather than computing the task-defined relation under the implied 3D structure, producing plausible yet incorrect orderings.

\textbf{3D spatial-logic errors.}
These failures reflect limitations in reconstructing and reasoning over a globally consistent 3D state under structural constraints. The model often shows weak depth reasoning, confusing near--far relations under perspective and foreshortening (common in cable, truss, and tower scenes). In Multi-View questions, it may fail to establish consistent cross-view correspondences, leading to incompatible interpretations across views. It also sometimes composes relations incorrectly; for example, from ``Member~1 is left of Member~2'' and ``Member~3 is above Member~1,'' it may wrongly infer ``Member~3 is left of Member~2,'' indicating unstable reference selection and non-robust relational composition.

Overall, these error modes suggest that the \ssibench performance gap arises not only from imperfect visual grounding, but also from limited 3D structural reconstruction and constraint-consistent spatial inference---capabilities central to SCSR. 
They also suggest two directions for future improvement: enhancing fine-grained structural perception through captions or annotations of component locations, extents, connectivity, and relations, and strengthening spatial reasoning through large-scale chain-of-thought data for structure-centric tasks.

\section{Conclusions}
In this work, we present \ssibench, a human-curated benchmark that targets spatial reasoning under structural constraints in vision--language models using complex real-world 3D structures. Evaluations across 31 VLMs expose a pronounced gap between current models and human performance, underscoring the difficulty of recovering constraint-consistent 3D structure from visual input. Our analyses further show that explicit thinking yields only modest improvements, with most errors tracing back to limitations in structural grounding and globally coherent 3D reasoning. We expect \ssibench to facilitate more fine-grained evaluation and to inform future advances in spatially grounded multimodal systems. Limitations are discussed in Appendix~\ref{app:limitations}.

\section*{Acknowledgements}
The authors gratefully acknowledge the financial support provided by the National Natural Science Foundation of China (Grant No. 52408188).

\section*{Impact Statement}
This work aims to advance the evaluation of spatial intelligence in vision--language models. 
While improved spatial reasoning can benefit applications such as robotics, accessibility, and infrastructure inspection, it may also have dual-use implications in sensitive contexts. 
\ssibench is an offline evaluation benchmark and does not provide action policies, control interfaces, or deployment data. 
We encourage its responsible use as a diagnostic tool for robust and transparent spatial reasoning.

\bibliography{example_paper}

@article{cheng2024spatialrgpt,
  title={Spatialrgpt: Grounded spatial reasoning in vision-language models},
  author={Cheng, An-Chieh and Yin, Hongxu and Fu, Yang and Guo, Qiushan and Yang, Ruihan and Kautz, Jan and Wang, Xiaolong and Liu, Sifei},
  journal={Advances in Neural Information Processing Systems},
  volume={37},
  pages={135062--135093},
  year={2024}
}

@inproceedings{chen2024spatialvlm,
  title={Spatialvlm: Endowing vision-language models with spatial reasoning capabilities},
  author={Chen, Boyuan and Xu, Zhuo and Kirmani, Sean and Ichter, Brain and Sadigh, Dorsa and Guibas, Leonidas and Xia, Fei},
  booktitle={Proceedings of the IEEE/CVF Conference on Computer Vision and Pattern Recognition},
  pages={14455--14465},
  year={2024}
}

@inproceedings{yang2025thinking,
  title={Thinking in space: How multimodal large language models see, remember, and recall spaces},
  author={Yang, Jihan and Yang, Shusheng and Gupta, Anjali W and Han, Rilyn and Fei-Fei, Li and Xie, Saining},
  booktitle={Proceedings of the Computer Vision and Pattern Recognition Conference},
  pages={10632--10643},
  year={2025}
}

@article{zhang2025flatland,
  title={From flatland to space: Teaching vision-language models to perceive and reason in 3d},
  author={Zhang, Jiahui and Chen, Yurui and Zhou, Yanpeng and Xu, Yueming and Huang, Ze and Mei, Jilin and Chen, Junhui and Yuan, Yu-Jie and Cai, Xinyue and Huang, Guowei and others},
  journal={arXiv preprint arXiv:2503.22976},
  year={2025}
}

@article{lin2025ost,
  title={Ost-bench: Evaluating the capabilities of mllms in online spatio-temporal scene understanding},
  author={Lin, JingLi and Zhu, Chenming and Xu, Runsen and Mao, Xiaohan and Liu, Xihui and Wang, Tai and Pang, Jiangmiao},
  journal={arXiv preprint arXiv:2507.07984},
  year={2025}
}

@article{xu2025multi,
  title={Multi-spatialmllm: Multi-frame spatial understanding with multi-modal large language models},
  author={Xu, Runsen and Wang, Weiyao and Tang, Hao and Chen, Xingyu and Wang, Xiaodong and Chu, Fu-Jen and Lin, Dahua and Feiszli, Matt and Liang, Kevin J},
  journal={arXiv preprint arXiv:2505.17015},
  year={2025}
}

@article{yang2025mmsi,
  title={MMSI-Bench: A Benchmark for Multi-Image Spatial Intelligence},
  author={Yang, Sihan and Xu, Runsen and Xie, Yiman and Yang, Sizhe and Li, Mo and Lin, Jingli and Zhu, Chenming and Chen, Xiaochen and Duan, Haodong and Yue, Xiangyu and others},
  journal={arXiv preprint arXiv:2505.23764},
  year={2025}
}

@article{li2025sti,
  title={Sti-bench: Are mllms ready for precise spatial-temporal world understanding?},
  author={Li, Yun and Zhang, Yiming and Lin, Tao and Liu, XiangRui and Cai, Wenxiao and Liu, Zheng and Zhao, Bo},
  journal={arXiv preprint arXiv:2503.23765},
  year={2025}
}

@article{zhang2025dsi,
  title={Dsi-bench: A benchmark for dynamic spatial intelligence},
  author={Zhang, Ziang and Wang, Zehan and Zhang, Guanghao and Dai, Weilong and Xia, Yan and Yan, Ziang and Hong, Minjie and Zhao, Zhou},
  journal={arXiv preprint arXiv:2510.18873},
  year={2025}
}

@inproceedings{yin2025spatial,
  title={Spatial mental modeling from limited views},
  author={Yin, Baiqiao and Wang, Qineng and Zhang, Pingyue and Zhang, Jianshu and Wang, Kangrui and Wang, Zihan and Zhang, Jieyu and Chandrasegaran, Keshigeyan and Liu, Han and Krishna, Ranjay and others},
  booktitle={Structural Priors for Vision Workshop at ICCV'25},
  year={2025}
}

@article{li2025viewspatial,
  title={ViewSpatial-Bench: Evaluating Multi-perspective Spatial Localization in Vision-Language Models},
  author={Li, Dingming and Li, Hongxing and Wang, Zixuan and Yan, Yuchen and Zhang, Hang and Chen, Siqi and Hou, Guiyang and Jiang, Shengpei and Zhang, Wenqi and Shen, Yongliang and others},
  journal={arXiv preprint arXiv:2505.21500},
  year={2025}
}

@article{lin2025mmsi,
  title={MMSI-Video-Bench: A Holistic Benchmark for Video-Based Spatial Intelligence},
  author={Lin, Jingli and Xu, Runsen and Zhu, Shaohao and Yang, Sihan and Cao, Peizhou and Ran, Yunlong and Hu, Miao and Zhu, Chenming and Xie, Yiman and Long, Yilin and others},
  journal={arXiv preprint arXiv:2512.10863},
  year={2025}
}

@article{zhu2025cvbench,
  title={CVBench: Benchmarking Cross-Video Synergies for Complex Multimodal Reasoning},
  author={Zhu, Nannan and Dong, Yonghao and Wang, Teng and Li, Xueqian and Deng, Shengjun and Wang, Yijia and Hong, Zheng and Geng, Tiantian and Niu, Guo and Huang, Hanyan and others},
  journal={arXiv preprint arXiv:2508.19542},
  year={2025}
}

@article{yeh2025seeing,
  title={Seeing from another perspective: Evaluating multi-view understanding in mllms},
  author={Yeh, Chun-Hsiao and Wang, Chenyu and Tong, Shengbang and Cheng, Ta-Ying and Wang, Ruoyu and Chu, Tianzhe and Zhai, Yuexiang and Chen, Yubei and Gao, Shenghua and Ma, Yi},
  journal={arXiv preprint arXiv:2504.15280},
  year={2025}
}

@article{he2025egoexobench,
  title={Egoexobench: A benchmark for first-and third-person view video understanding in mllms},
  author={He, Yuping and Huang, Yifei and Chen, Guo and Pei, Baoqi and Xu, Jilan and Lu, Tong and Pang, Jiangmiao},
  journal={arXiv preprint arXiv:2507.18342},
  year={2025}
}

@inproceedings{johnson2017clevr,
  title={Clevr: A diagnostic dataset for compositional language and elementary visual reasoning},
  author={Johnson, Justin and Hariharan, Bharath and Van Der Maaten, Laurens and Fei-Fei, Li and Lawrence Zitnick, C and Girshick, Ross},
  booktitle={Proceedings of the IEEE conference on computer vision and pattern recognition},
  pages={2901--2910},
  year={2017}
}

@inproceedings{hudson2019gqa,
  title={Gqa: A new dataset for real-world visual reasoning and compositional question answering},
  author={Hudson, Drew A and Manning, Christopher D},
  booktitle={Proceedings of the IEEE/CVF conference on computer vision and pattern recognition},
  pages={6700--6709},
  year={2019}
}

@inproceedings{koch2019abc,
  title={Abc: A big cad model dataset for geometric deep learning},
  author={Koch, Sebastian and Matveev, Albert and Jiang, Zhongshi and Williams, Francis and Artemov, Alexey and Burnaev, Evgeny and Alexa, Marc and Zorin, Denis and Panozzo, Daniele},
  booktitle={Proceedings of the IEEE/CVF conference on computer vision and pattern recognition},
  pages={9601--9611},
  year={2019}
}

@inproceedings{chen2021geoqa,
  title={Geoqa: A geometric question answering benchmark towards multimodal numerical reasoning},
  author={Chen, Jiaqi and Tang, Jianheng and Qin, Jinghui and Liang, Xiaodan and Liu, Lingbo and Xing, Eric and Lin, Liang},
  booktitle={Findings of the Association for Computational Linguistics: ACL-IJCNLP 2021},
  pages={513--523},
  year={2021}
}

@inproceedings{collins2022abo,
  title={Abo: Dataset and benchmarks for real-world 3d object understanding},
  author={Collins, Jasmine and Goel, Shubham and Deng, Kenan and Luthra, Achleshwar and Xu, Leon and Gundogdu, Erhan and Zhang, Xi and Vicente, Tomas F Yago and Dideriksen, Thomas and Arora, Himanshu and others},
  booktitle={Proceedings of the IEEE/CVF conference on computer vision and pattern recognition},
  pages={21126--21136},
  year={2022}
}

@inproceedings{mo2019partnet,
  title={Partnet: A large-scale benchmark for fine-grained and hierarchical part-level 3d object understanding},
  author={Mo, Kaichun and Zhu, Shilin and Chang, Angel X and Yi, Li and Tripathi, Subarna and Guibas, Leonidas J and Su, Hao},
  booktitle={Proceedings of the IEEE/CVF conference on computer vision and pattern recognition},
  pages={909--918},
  year={2019}
}

@article{slim20253dcompat++,
  title={3dcompat++: An improved large-scale 3d vision dataset for compositional recognition},
  author={Slim, Habib and Li, Xiang and Li, Yuchen and Ahmed, Mahmoud and Ayman, Mohamed and Upadhyay, Ujjwal and Abdelreheem, Ahmed and Prajapati, Arpit and Pothigara, Suhail and Wonka, Peter and others},
  journal={IEEE Transactions on Pattern Analysis and Machine Intelligence},
  year={2025},
  publisher={IEEE}
}

@article{li2025benchmark,
  title={Benchmark evaluations, applications, and challenges of large vision language models: A survey},
  author={Li, Zongxia and Wu, Xiyang and Du, Hongyang and Nghiem, Huy and Shi, Guangyao},
  journal={arXiv preprint arXiv:2501.02189},
  volume={1},
  year={2025}
}

@inproceedings{liu2024improved,
  title={Improved baselines with visual instruction tuning},
  author={Liu, Haotian and Li, Chunyuan and Li, Yuheng and Lee, Yong Jae},
  booktitle={Proceedings of the IEEE/CVF conference on computer vision and pattern recognition},
  pages={26296--26306},
  year={2024}
}

@article{xu2023analysis,
  title={Analysis and assessment of life-cycle carbon emissions of space frame structures},
  author={Xu, Xian and You, Jianzhou and Wang, Yafeng and Luo, Yaozhi},
  journal={Journal of Cleaner Production},
  volume={385},
  pages={135521},
  year={2023},
  publisher={Elsevier}
}

@article{bezas2022design,
  title={Design recommendations for the stability of transmission steel lattice towers},
  author={Bezas, Marios-Zois and Jaspart, Jean-Pierre and Vayas, Ioannis and Demonceau, Jean-Fran{\c{c}}ois},
  journal={Engineering Structures},
  volume={252},
  pages={113603},
  year={2022},
  publisher={Elsevier}
}

@inproceedings{qi2024first,
  title={The first engineering application of 10MN CFRP cables in cable-stayed bridge in China},
  author={Qi, Ligang and Bai, Jie and Wu, Hangzi and Xu, Guowen and Xiong, Hao and Yang, Yan},
  booktitle={Structures},
  volume={68},
  pages={107199},
  year={2024},
  organization={Elsevier}
}

@article{vollmecke2025assessment,
  title={Assessment of nailed connections in existing timber trusses},
  author={V{\"o}llmecke, Lars and Krenzer, Adrian and Seim, Werner},
  journal={Construction and Building Materials},
  volume={491},
  pages={142359},
  year={2025},
  publisher={Elsevier}
}

@article{yang2022novel,
  title={Novel leakage detection by ensemble 1DCNN-VAPSO-SVM in oil and gas pipeline systems},
  author={Yang, Dandi and Hou, Nan and Lu, Jingyi and Ji, Daan},
  journal={Applied Soft Computing},
  volume={115},
  pages={108212},
  year={2022},
  publisher={Elsevier}
}

@misc{pexels-website,
  author       = {{Pexels}},
  title        = {Pexels},
  howpublished = {\url{https://www.pexels.com/}},
  note         = {Accessed: 2026-01-10}
}

@misc{unsplash-website,
  author       = {{Unsplash}},
  title        = {Unsplash},
  howpublished = {\url{https://unsplash.com/}},
  note         = {Accessed: 2026-01-10}
}

@misc{pixabay-website,
  author       = {{Pixabay}},
  title        = {Pixabay},
  howpublished = {\url{https://pixabay.com/}},
  note         = {Accessed: 2026-01-10}
}

@inproceedings{duan2024vlmevalkit,
  title={Vlmevalkit: An open-source toolkit for evaluating large multi-modality models},
  author={Duan, Haodong and Yang, Junming and Qiao, Yuxuan and Fang, Xinyu and Chen, Lin and Liu, Yuan and Dong, Xiaoyi and Zang, Yuhang and Zhang, Pan and Wang, Jiaqi and others},
  booktitle={Proceedings of the 32nd ACM international conference on multimedia},
  pages={11198--11201},
  year={2024}
}

@misc{google-gemini3-website,
  author       = {{Google DeepMind}},
  title        = {A new era of intelligence with Gemini 3},
  howpublished = {\url{https://blog.google/products-and-platforms/products/gemini/gemini-3/}},
  year         = {2025},
  note         = {Accessed: 2026-01-10}
}

@misc{google-gemini25-thinking-updates-website,
  author       = {{Google DeepMind}},
  title        = {Gemini 2.5: Our most intelligent AI model},
  howpublished = {\url{https://blog.google/innovation-and-ai/models-and-research/google-deepmind/gemini-model-thinking-updates-march-2025/}},
  year         = {2025},
  note         = {Accessed: 2026-01-10}
}

@misc{openai-gpt5.2-website,
  author       = {{OpenAI}},
  title        = {Introducing GPT-5.2},
  howpublished = {\url{https://openai.com/index/introducing-gpt-5-2/}},
  year         = {2025},
  note         = {Accessed: 2026-01-10}
}

@misc{openai-gpt5-website,
  author       = {{OpenAI}},
  title        = {Introducing GPT-5},
  howpublished = {\url{https://openai.com/index/introducing-gpt-5/}},
  year         = {2025},
  note         = {Accessed: 2026-01-10}
}

@misc{openai-gpt4.1-website,
  author       = {{OpenAI}},
  title        = {Introducing GPT-4.1 in the API},
  howpublished = {\url{https://openai.com/index/gpt-4-1/}},
  year         = {2025},
  note         = {Accessed: 2026-01-10}
}

@misc{openai-gpt4o-website,
  author       = {{OpenAI}},
  title        = {Hello GPT-4o},
  howpublished = {\url{https://openai.com/index/hello-gpt-4o/}},
  year         = {2024},
  note         = {Accessed: 2026-01-10}
}

@misc{anthropic-claude-sonnet45-website,
  author       = {{Anthropic}},
  title        = {Introducing Claude Sonnet 4.5},
  howpublished = {\url{https://www.anthropic.com/news/claude-sonnet-4-5}},
  year         = {2025},
  note         = {Accessed: 2026-01-10}
}

@misc{bytedance-seed18-website,
  author       = {{ByteDance Seed}},
  title        = {Seed1.8},
  howpublished = {\url{https://seed.bytedance.com/seed1_8}},
  year         = {2025},
  note         = {Accessed: 2026-01-10}
}

@misc{bai2025qwen3vltechnicalreport,
      title={Qwen3-VL Technical Report}, 
      author={Shuai Bai and Yuxuan Cai and Ruizhe Chen and Keqin Chen and Xionghui Chen and Zesen Cheng and Lianghao Deng and Wei Ding and Chang Gao and Chunjiang Ge and Wenbin Ge and Zhifang Guo and Qidong Huang and Jie Huang and Fei Huang and Binyuan Hui and Shutong Jiang and Zhaohai Li and Mingsheng Li and Mei Li and Kaixin Li and Zicheng Lin and Junyang Lin and Xuejing Liu and Jiawei Liu and Chenglong Liu and Yang Liu and Dayiheng Liu and Shixuan Liu and Dunjie Lu and Ruilin Luo and Chenxu Lv and Rui Men and Lingchen Meng and Xuancheng Ren and Xingzhang Ren and Sibo Song and Yuchong Sun and Jun Tang and Jianhong Tu and Jianqiang Wan and Peng Wang and Pengfei Wang and Qiuyue Wang and Yuxuan Wang and Tianbao Xie and Yiheng Xu and Haiyang Xu and Jin Xu and Zhibo Yang and Mingkun Yang and Jianxin Yang and An Yang and Bowen Yu and Fei Zhang and Hang Zhang and Xi Zhang and Bo Zheng and Humen Zhong and Jingren Zhou and Fan Zhou and Jing Zhou and Yuanzhi Zhu and Ke Zhu},
      year={2025},
      eprint={2511.21631},
      archivePrefix={arXiv},
      primaryClass={cs.CV},
}

@misc{vteam2026glm45vglm41vthinkingversatilemultimodal,
      title={GLM-4.5V and GLM-4.1V-Thinking: Towards Versatile Multimodal Reasoning with Scalable Reinforcement Learning}, 
      author={Wenyi Hong and Wenmeng Yu and Xiaotao Gu and Guo Wang and Guobing Gan and Haomiao Tang and Jiale Cheng and Ji Qi and Junhui Ji and Lihang Pan and Shuaiqi Duan and Weihan Wang and Yan Wang and Yean Cheng and Zehai He and Zhe Su and Zhen Yang and Ziyang Pan and Aohan Zeng and Baoxu Wang and Bin Chen and Boyan Shi and Changyu Pang and Chenhui Zhang and Da Yin and Fan Yang and Guoqing Chen and Haochen Li and Jiale Zhu and Jiali Chen and Jiaxing Xu and Jiazheng Xu and Jing Chen and Jinghao Lin and Jinhao Chen and Jinjiang Wang and Junjie Chen and Leqi Lei and Letian Gong and Leyi Pan and Mingdao Liu and Mingde Xu and Mingzhi Zhang and Qinkai Zheng and Ruiliang Lyu and Shangqin Tu and Sheng Yang and Shengbiao Meng and Shi Zhong and Shiyu Huang and Shuyuan Zhao and Siyan Xue and Tianshu Zhang and Tianwei Luo and Tianxiang Hao and Tianyu Tong and Wei Jia and Wenkai Li and Xiao Liu and Xiaohan Zhang and Xin Lyu and Xinyu Zhang and Xinyue Fan and Xuancheng Huang and Yadong Xue and Yanfeng Wang and Yanling Wang and Yanzi Wang and Yifan An and Yifan Du and Yiheng Huang and Yilin Niu and Yiming Shi and Yu Wang and Yuan Wang and Yuanchang Yue and Yuchen Li and Yusen Liu and Yutao Zhang and Yuting Wang and Yuxuan Zhang and Zhao Xue and Zhengxiao Du and Zhenyu Hou and Zihan Wang and Peng Zhang and Debing Liu and Bin Xu and Juanzi Li and Minlie Huang and Yuxiao Dong and Jie Tang},
      year={2026},
      eprint={2507.01006},
      archivePrefix={arXiv},
      primaryClass={cs.CV},
}

@misc{wang2025internvl35advancingopensourcemultimodal,
      title={InternVL3.5: Advancing Open-Source Multimodal Models in Versatility, Reasoning, and Efficiency}, 
      author={Weiyun Wang and Zhangwei Gao and Lixin Gu and Hengjun Pu and Long Cui and Xingguang Wei and Zhaoyang Liu and Linglin Jing and Shenglong Ye and Jie Shao and Zhaokai Wang and Zhe Chen and Hongjie Zhang and Ganlin Yang and Haomin Wang and Qi Wei and Jinhui Yin and Wenhao Li and Erfei Cui and Guanzhou Chen and Zichen Ding and Changyao Tian and Zhenyu Wu and Jingjing Xie and Zehao Li and Bowen Yang and Yuchen Duan and Xuehui Wang and Zhi Hou and Haoran Hao and Tianyi Zhang and Songze Li and Xiangyu Zhao and Haodong Duan and Nianchen Deng and Bin Fu and Yinan He and Yi Wang and Conghui He and Botian Shi and Junjun He and Yingtong Xiong and Han Lv and Lijun Wu and Wenqi Shao and Kaipeng Zhang and Huipeng Deng and Biqing Qi and Jiaye Ge and Qipeng Guo and Wenwei Zhang and Songyang Zhang and Maosong Cao and Junyao Lin and Kexian Tang and Jianfei Gao and Haian Huang and Yuzhe Gu and Chengqi Lyu and Huanze Tang and Rui Wang and Haijun Lv and Wanli Ouyang and Limin Wang and Min Dou and Xizhou Zhu and Tong Lu and Dahua Lin and Jifeng Dai and Weijie Su and Bowen Zhou and Kai Chen and Yu Qiao and Wenhai Wang and Gen Luo},
      year={2025},
      eprint={2508.18265},
      archivePrefix={arXiv},
      primaryClass={cs.CV},
}

@misc{gemmateam2025gemma3technicalreport,
      title={Gemma 3 Technical Report}, 
      author={Aishwarya Kamath and Johan Ferret and Shreya Pathak and Nino Vieillard and Ramona Merhej and Sarah Perrin and Tatiana Matejovicova and Alexandre Ramé and Morgane Rivière and Louis Rouillard and Thomas Mesnard and Geoffrey Cideron and Jean-bastien Grill and Sabela Ramos and Edouard Yvinec and Michelle Casbon and Etienne Pot and Ivo Penchev and Gaël Liu and Francesco Visin and Kathleen Kenealy and Lucas Beyer and Xiaohai Zhai and Anton Tsitsulin and Robert Busa-Fekete and Alex Feng and Noveen Sachdeva and Benjamin Coleman and Yi Gao and Basil Mustafa and Iain Barr and Emilio Parisotto and David Tian and Matan Eyal and Colin Cherry and Jan-Thorsten Peter and Danila Sinopalnikov and Surya Bhupatiraju and Rishabh Agarwal and Mehran Kazemi and Dan Malkin and Ravin Kumar and David Vilar and Idan Brusilovsky and Jiaming Luo and Andreas Steiner and Abe Friesen and Abhanshu Sharma and Abheesht Sharma and Adi Mayrav Gilady and Adrian Goedeckemeyer and Alaa Saade and Alex Feng and Alexander Kolesnikov and Alexei Bendebury and Alvin Abdagic and Amit Vadi and András György and André Susano Pinto and Anil Das and Ankur Bapna and Antoine Miech and Antoine Yang and Antonia Paterson and Ashish Shenoy and Ayan Chakrabarti and Bilal Piot and Bo Wu and Bobak Shahriari and Bryce Petrini and Charlie Chen and Charline Le Lan and Christopher A. Choquette-Choo and CJ Carey and Cormac Brick and Daniel Deutsch and Danielle Eisenbud and Dee Cattle and Derek Cheng and Dimitris Paparas and Divyashree Shivakumar Sreepathihalli and Doug Reid and Dustin Tran and Dustin Zelle and Eric Noland and Erwin Huizenga and Eugene Kharitonov and Frederick Liu and Gagik Amirkhanyan and Glenn Cameron and Hadi Hashemi and Hanna Klimczak-Plucińska and Harman Singh and Harsh Mehta and Harshal Tushar Lehri and Hussein Hazimeh and Ian Ballantyne and Idan Szpektor and Ivan Nardini and Jean Pouget-Abadie and Jetha Chan and Joe Stanton and John Wieting and Jonathan Lai and Jordi Orbay and Joseph Fernandez and Josh Newlan and Ju-yeong Ji and Jyotinder Singh and Kat Black and Kathy Yu and Kevin Hui and Kiran Vodrahalli and Klaus Greff and Linhai Qiu and Marcella Valentine and Marina Coelho and Marvin Ritter and Matt Hoffman and Matthew Watson and Mayank Chaturvedi and Michael Moynihan and Min Ma and Nabila Babar and Natasha Noy and Nathan Byrd and Nick Roy and Nikola Momchev and Nilay Chauhan and Noveen Sachdeva and Oskar Bunyan and Pankil Botarda and Paul Caron and Paul Kishan Rubenstein and Phil Culliton and Philipp Schmid and Pier Giuseppe Sessa and Pingmei Xu and Piotr Stanczyk and Pouya Tafti and Rakesh Shivanna and Renjie Wu and Renke Pan and Reza Rokni and Rob Willoughby and Rohith Vallu and Ryan Mullins and Sammy Jerome and Sara Smoot and Sertan Girgin and Shariq Iqbal and Shashir Reddy and Shruti Sheth and Siim Põder and Sijal Bhatnagar and Sindhu Raghuram Panyam and Sivan Eiger and Susan Zhang and Tianqi Liu and Trevor Yacovone and Tyler Liechty and Uday Kalra and Utku Evci and Vedant Misra and Vincent Roseberry and Vlad Feinberg and Vlad Kolesnikov and Woohyun Han and Woosuk Kwon and Xi Chen and Yinlam Chow and Yuvein Zhu and Zichuan Wei and Zoltan Egyed and Victor Cotruta and Minh Giang and Phoebe Kirk and Anand Rao and Kat Black and Nabila Babar and Jessica Lo and Erica Moreira and Luiz Gustavo Martins and Omar Sanseviero and Lucas Gonzalez and Zach Gleicher and Tris Warkentin and Vahab Mirrokni and Evan Senter and Eli Collins and Joelle Barral and Zoubin Ghahramani and Raia Hadsell and Yossi Matias and D. Sculley and Slav Petrov and Noah Fiedel and Noam Shazeer and Oriol Vinyals and Jeff Dean and Demis Hassabis and Koray Kavukcuoglu and Clement Farabet and Elena Buchatskaya and Jean-Baptiste Alayrac and Rohan Anil and Dmitry and Lepikhin and Sebastian Borgeaud and Olivier Bachem and Armand Joulin and Alek Andreev and Cassidy Hardin and Robert Dadashi and Léonard Hussenot},
      year={2025},
      eprint={2503.19786},
      archivePrefix={arXiv},
      primaryClass={cs.CL},
}

@misc{meta-llama4-multimodal-intelligence,
  author       = {{Meta AI}},
  title        = {The Llama 4 herd: The beginning of a new era of natively multimodal AI innovation},
  howpublished = {\url{https://ai.meta.com/blog/llama-4-multimodal-intelligence/}},
  year         = {2025},
  note         = {Accessed: 2026-01-10}
}

@misc{li2024llavaonevisioneasyvisualtask,
      title={LLaVA-OneVision: Easy Visual Task Transfer}, 
      author={Bo Li and Yuanhan Zhang and Dong Guo and Renrui Zhang and Feng Li and Hao Zhang and Kaichen Zhang and Peiyuan Zhang and Yanwei Li and Ziwei Liu and Chunyuan Li},
      year={2024},
      eprint={2408.03326},
      archivePrefix={arXiv},
      primaryClass={cs.CV},
}

@article{deng2025interactcomp,
  title={Interactcomp: Evaluating search agents with ambiguous queries},
  author={Deng, Mingyi and Huang, Lijun and Fan, Yani and Zhang, Jiayi and Ren, Fashen and Bai, Jinyi and Yang, Fuzhen and Miao, Dayi and Yu, Zhaoyang and Wu, Yifan and others},
  journal={arXiv preprint arXiv:2510.24668},
  year={2025}
}

@inproceedings{ma20253dsrbench,
  title={3dsrbench: A comprehensive 3d spatial reasoning benchmark},
  author={Ma, Wufei and Chen, Haoyu and Zhang, Guofeng and Chou, Yu-Cheng and Chen, Jieneng and de Melo, Celso and Yuille, Alan},
  booktitle={Proceedings of the IEEE/CVF International Conference on Computer Vision},
  pages={6924--6934},
  year={2025}
}

@inproceedings{wang2025spatial457,
  title={Spatial457: A diagnostic benchmark for 6d spatial reasoning of large mutimodal models},
  author={Wang, Xingrui and Ma, Wufei and Zhang, Tiezheng and de Melo, Celso M and Chen, Jieneng and Yuille, Alan},
  booktitle={Proceedings of the Computer Vision and Pattern Recognition Conference},
  pages={24669--24679},
  year={2025}
}

@article{jia2025omnispatial,
  title={Omnispatial: Towards comprehensive spatial reasoning benchmark for vision language models},
  author={Jia, Mengdi and Qi, Zekun and Zhang, Shaochen and Zhang, Wenyao and Yu, Xinqiang and He, Jiawei and Wang, He and Yi, Li},
  journal={arXiv preprint arXiv:2506.03135},
  year={2025}
}

@article{wang2025enact,
  title={Enact: Evaluating embodied cognition with world modeling of egocentric interaction},
  author={Wang, Qineng and Huang, Wenlong and Zhou, Yu and Yin, Hang and Bao, Tianwei and Lyu, Jianwen and Liu, Weiyu and Zhang, Ruohan and Wu, Jiajun and Fei-Fei, Li and others},
  journal={arXiv preprint arXiv:2511.20937},
  year={2025}
}
\bibliographystyle{icml2026}

\newpage
\appendix
\onecolumn

\section{Appendix Overview and Organization}
\label{app:appendix_overview}

This appendix provides supplementary details to support and extend the main paper, with an emphasis on transparency and reproducibility. The appendix is organized as follows.

\noindent\textbf{(1) Comparison with existing benchmarks (Section~\ref{app:ssi_comparison}).}
We provide detailed comparisons between \ssibench and prior benchmarks to clarify the positioning and distinctive challenges covered by our benchmark.

\noindent\textbf{(2) Structural constraint types (Section~\ref{app:structural_constraints}).}
We provide a detailed categorization of the structural constraints used in \ssibench, including geometric regularity, topological and connectivity constraints, and physical and semantic feasibility.

\noindent\textbf{(3) Dataset statistics (Section~\ref{app:data_statistics}).}
We report summary statistics of the dataset and the distributional properties of instances across tasks and categories.

\noindent\textbf{(4) Details of the Data Curation Process (Section~\ref{app:ssi_constr_details}).}
We provide additional details on the benchmark construction process beyond the main text, including data sources (Section~\ref{app:data_sources}), data annotation and quality control interfaces (Section~\ref{app:annotation_qc_interfaces}), general annotation guidelines (Section~\ref{app:general_guidelines}), the data format (Section~\ref{app:data_format_structure}), tie handling and evaluation fairness (Section~\ref{app:tie_handling}), semi-automatic expansion (Section~\ref{app:semi_auto_expansion}), and the difficulty annotation methodology (Section~\ref{app:difficulty_annotation}).

\noindent\textbf{(5) Additional implementation details and experimental results (Section~\ref{app:implementation_details}).}
We provide metric definitions (Section~\ref{app:eval_metrics}), evaluated models and inference settings (Section~\ref{app:benchmark_models}), implementation details for model evaluation (Section~\ref{app:model_eval_details}), and the human evaluation protocol (Section~\ref{app:human_eval_setup}), along with additional results such as pairwise accuracy (Section~\ref{app:pairwise_results}) and complete analyses of thinking-related settings (Section~\ref{app:thinking_results}).

\noindent\textbf{(6) Representative benchmark examples (Section~\ref{app:samples}).}
We present representative instances from each category with full prompts and visual annotations to illustrate the diversity of structure types and task demands.

\noindent\textbf{(7) Additional case studies (Section~\ref{app:additional_case}).}
We provide further qualitative analyses illustrating typical model behaviors and failure patterns on SCSR, complementing the examples discussed in the main text.

\noindent\textbf{(8) Limitations (Section~\ref{app:limitations}).}
We discuss limitations of \ssibench and our evaluation protocol, and outline directions for future work.

\noindent Overall, these sections provide additional transparency on dataset construction and evaluation, enabling reproducibility and broader contextualization of our findings.

\section{Comparison with Other Spatial Intelligence Benchmarks}
\label{app:ssi_comparison}

Table~\ref{tab:benchmark_comparison} compares \ssibench with representative spatial intelligence benchmarks along the axes used throughout our paper: (i) regime (unconstrained vs.\ structurally constrained), (ii) modality (single-/multi-image vs.\ video/multi-video), (iii) annotation (automatic vs.\ human), and (iv) task coverage (spatial understanding, motion understanding, and cross-video reasoning). Beyond cataloging differences, we emphasize a central theme of \ssibench: by operating in a \emph{structurally constrained} regime, SCSR places a \emph{higher demand on genuine 3D understanding}---models must infer a constraint-consistent 3D scene hypothesis rather than rely on weak 2D correlations. At the same time, these strong structural constraints narrow the space of feasible 3D configurations and make the queried relations more stable under plausible interpretations, enabling evaluation that is more \emph{quantifiable and comparable} across models and runs.

\textbf{Single-image benchmarks.}
Early spatial benchmarks largely evaluate spatial understanding from a single view in everyday environments where feasible 3D configurations are only weakly restricted.
SpatialRGPT~\cite{cheng2024spatialrgpt} and SpatialVLM~\cite{chen2024spatialvlm} focus on local metric or relational cues from one image, typically relying on automatic or mixed auto/human annotations.
CVBench~\cite{zhu2025cvbench} further scales single-image evaluation with automatically generated questions.
These benchmarks are valuable for tracking progress on local spatial cues, but their scenes are commonly weakly constrained: many different 3D states can plausibly explain the same observation.
As a result, models can sometimes achieve non-trivial performance via 2D correlations, appearance priors, or dataset regularities, without committing to a constraint-consistent 3D structural interpretation.

\textbf{Multi-image benchmarks.}
A second line improves over single-view ambiguity by providing multiple images and probing cross-view reasoning.
MultiSPA~\cite{xu2025multi} introduces multi-image settings with automatic generation; All-Angles-Bench~\cite{yeh2025seeing} emphasizes viewpoint changes with human annotations.
MMSI-Bench~\cite{yang2025mmsi}, ViewSpatial-Bench~\cite{li2025viewspatial}, and MindCube~\cite{yin2025spatial} expand multi-image evaluation in task diversity and scale, aiming to test whether models can fuse observations to infer more reliable 3D relations.
While adding views generally strengthens supervision and reduces under-determination, these benchmarks still predominantly operate in unconstrained everyday regimes, where feasible configurations span a broad space.
Consequently, even in multi-image settings, models may partially rely on multi-view correlations that do not require enforcing strong feasibility constraints or recovering a globally consistent structural state.

\textbf{Video benchmarks for spatio-temporal reasoning.}
Video-based benchmarks extend spatial intelligence to spatio-temporal understanding and often include motion-centric tasks.
VSI-Bench~\cite{yang2025thinking} tests spatio-temporal spatial understanding and planning-related components; OST-Bench~\cite{lin2025ost} targets object--state transitions; SPAR-Bench~\cite{zhang2025flatland} evaluates richer spatio-temporal reasoning.
STI-Bench~\cite{li2025sti} and DSI-Bench~\cite{zhang2025dsi} further emphasize dynamic instance states, with DSI-Bench incorporating human annotation.
These efforts broaden evaluation beyond static geometry, but most still focus on scenes with weak feasibility constraints, where strong performance may be attainable through temporal tracking heuristics or action priors, rather than constraint-consistent 3D reconstruction.

\textbf{Multi-video and cross-video benchmarks.}
More recent benchmarks investigate correspondence and reasoning across videos.
EgoExoBench~\cite{he2025egoexobench} evaluates alignment across egocentric/exocentric views, and MMSI-Video~\cite{lin2025mmsi} extends multi-video reasoning with human annotation and additional task coverage (e.g., planning, cross-video reasoning).
These datasets raise the difficulty through cross-episode generalization, yet their underlying environments are typically not governed by strong geometric/topological feasibility constraints.

\textbf{\ssibench: structural constraints enable well-defined ranking-based evaluation.}
In contrast to the above unconstrained benchmarks, \ssibench targets SCSR, which places a \emph{higher demand on genuine 3D understanding}: models must recover a constraint-consistent 3D structural hypothesis (geometry \emph{and} topology) from visual evidence and reason over it, rather than relying on weakly constrained 2D correlations or appearance priors. The strong geometric, topological, and physics-based constraints of real-world structures sharply narrow the space of feasible configurations, making many queried relations more stable under plausible interpretations and thereby enabling evaluation that is more \emph{quantifiable and comparable} across models. To leverage this property, \ssibench adopts a ranking formulation over 3--4 candidates under explicit geometric/topological criteria, which emphasizes robust comparative reasoning and converts constraint-induced stability into a directly measurable signal, especially in cluttered, occluded, and viewpoint-diverse structural scenes.

\begin{table*}[t]
\caption{Comparison of \ssibench with representative spatial reasoning benchmarks in terms of evaluation regime, modality, annotation type, and task coverage. SU/MU denote spatial/motion understanding; CV denotes cross-video reasoning. SC/UC indicate structurally constrained/unconstrained regimes. Human--AI Gap is the absolute performance difference (percentage points) between humans and the best reported AI system.}
  \label{tab:benchmark_comparison}
  \vspace{-3pt}
  \centering
  \begin{footnotesize}
    \setlength{\tabcolsep}{3.2pt}
    \renewcommand{\arraystretch}{1.06}

    \begin{tabular*}{\textwidth}{@{\extracolsep{\fill}} c c c c c c c @{}}
      \toprule
      \textbf{Benchmark} & \textbf{Regime} & \textbf{Modality} & \textbf{Annotation} & \textbf{Task} & \textbf{Samples} & \textbf{Human--AI Gap} \\
      \midrule
      SpatialRGPT~\cite{cheng2024spatialrgpt} & UC & Single-Image & Auto & SU & 1{,}406 & $<$42 \\
      SpatialVLM~\cite{chen2024spatialvlm}   & UC & Single-Image & Auto \& Human & SU & 546 & $<$30 \\
      CVBench~\cite{zhu2025cvbench}                               & UC & Single-Image & Auto & SU & 2{,}638 & -- \\
      MultiSPA~\cite{xu2025multi}                              & UC & Multi-Image & Auto & SU \& MU & 7{,}800 & -- \\
      All-Angles-Bench~\cite{yeh2025seeing}                      & UC & Multi-Image & Human & SU & 2{,}100 & 21.2 \\
      MMSI-Bench~\cite{yang2025mmsi}         & UC & Multi-Image & Human & SU \& MU & 1{,}000 & 55.3 \\
      ViewSpatial-Bench~\cite{li2025viewspatial}         & UC & Multi-Image & Auto \& Human & SU & 5{,}712 & -- \\
      MindCube~\cite{yin2025spatial}         & UC & Multi-Image & Auto \& Human & SU \& MU & 21{,}154 & -- \\
      VSI-Bench~\cite{yang2025thinking}      & UC & Video & Auto & SU & 5{,}000 & 33 \\
      OST-Bench~\cite{lin2025ost}            & UC & Video & Auto & SU \& MU & 10{,}000 & 29.3 \\
      SPAR-Bench~\cite{zhang2025flatland}    & UC & Video & Auto & SU \& MU & 7{,}207 & 27.8 \\
      STI-Bench~\cite{li2025sti}             & UC & Video & Auto & SU \& MU & 2{,}064 & -- \\
      DSI-Bench~\cite{zhang2025dsi}     & UC & Video & Human & SU \& MU & 1{,}000 & -- \\
      EgoExoBench~\cite{he2025egoexobench}                            & UC & Multi-Video & Auto \& Human & CV & 7{,}330 & 41.9 \\
      MMSI-Video~\cite{lin2025mmsi}  & UC & Video/Multi-Video & Human & SU \& MU \& CV & 1{,}106 & 58.4 \\
      \midrule
      \textbf{\ssibench (ours)} & \textbf{SC} & \textbf{Single-/Multi-Image} & \textbf{Human} & \textbf{SU} & \textbf{1{,}000} & \textbf{58.0} \\
      \bottomrule
    \end{tabular*}
  \end{footnotesize}
  \vskip -0.2in
\end{table*}

\section{Structural Constraint Types}
\label{app:structural_constraints}

We categorize the constraints used in \ssibench into three groups: geometric regularity, topological and connectivity constraints, and physical and semantic feasibility, as summarized in Table~\ref{tab:structural_constraints}. 
These constraint types are not provided to models during evaluation. 
Instead, they describe the implicit regularities used during benchmark construction to select suitable scenes, design unambiguous questions, and ensure that the target rankings can be inferred from visual evidence rather than from superficial 2D cues.

Geometric regularity constrains the metric and directional layout of components. 
Topological and connectivity constraints restrict how nodes and members can be connected and how graph-level relations are preserved under projection. 
Physical and semantic feasibility rules out implausible configurations according to support, gravity, material behavior, and the functional roles of structural components. 
Together, these constraints reduce the space of plausible 3D configurations and make the queried spatial relations more stable under reasonable interpretations.

\begin{table}[t]
    \centering
    \caption{Categorization and description of structural constraints in \ssibench.}
    \label{tab:structural_constraints}
    \vspace{-4pt}
    \small
    \renewcommand{\arraystretch}{1.1}
    \begin{tabularx}{\textwidth}{@{} 
        >{\raggedright\arraybackslash}p{1.8cm}
        >{\raggedright\arraybackslash}p{2cm}
        >{\raggedright\arraybackslash}X
        >{\raggedright\arraybackslash}X
        >{\raggedright\arraybackslash}X
        @{}}
        \toprule
        \textbf{Category} & \textbf{Constraint Type} & \textbf{Mechanism} & \textbf{Example} & \textbf{3D Inference Benefit} \\
        \midrule
        
        Geometric Regularity 
        & Symmetry 
        & Mirrored geometry and node positions across an axis or plane. 
        & A pitched house roof or a symmetrical stadium dome, where the left and right halves are mirrored. 
        & Helps infer occluded or heavily foreshortened parts from their visible counterparts. \\
        
        & Parallelism \& Orthogonality 
        & Members align parallel ($A \parallel B$) or orthogonal ($A \perp B$). 
        & A multi-story parking garage, where floor slabs are parallel and support columns are orthogonal to floors. 
        & Establishes a global coordinate frame to resolve perspective distortion. \\
        
        & Coplanarity \& Collinearity 
        & Nodes or members share a single 3D plane or line. 
        & The flat glass curtain wall of a modern commercial building facade. 
        & Restricts depth variation and reduces geometric degrees of freedom. \\
        
        & Periodicity \& Equidistance 
        & Structural units repeat with equal or regular spacing. 
        & Bleacher seating in a stadium or regular steps on a public staircase. 
        & Replaces absolute metric measurement with structural counting. \\
        
        \midrule
        
        Topological \& Connectivity 
        & Joint Sharing 
        & Converging members share exact 3D endpoint coordinates. 
        & A steel space-frame roof where multiple struts meet at a single spherical node. 
        & Provides hard equality constraints and prevents inconsistent floating endpoints. \\
        
        & Sequence \& Enclosure 
        & Fixed connection rules form ordered chains or closed invariant loops. 
        & The side rails of a pedestrian truss bridge, forming closed and stable triangular patterns. 
        & Preserves invariant topology under 2D projection and narrows feasible 3D layouts. \\
        
        & Anchoring 
        & Base nodes rest on a common datum plane. 
        & The concrete piers of a highway overpass, all anchored to the same ground-level foundation. 
        & Establishes an absolute zero-elevation reference. \\
        
        \midrule
        
        Physical \& Semantic 
        & Gravity Alignment 
        & Load-bearing members align with gravity or remain level relative to it. 
        & The main load-bearing columns of a shopping mall and its horizontal floor decks. 
        & Defines a global vertical axis and resolves up--down ambiguity under camera tilt. \\
        
        & Shape Physics 
        & Components deform into predictable, physics-governed shapes. 
        & The main cables of a large suspension bridge, forming a catenary curve under load. 
        & Injects physical priors to prevent arbitrary or implausible 3D interpretations. \\
        
        \bottomrule
    \end{tabularx}
  \vskip -0.1in
\end{table}

\section{Dataset Statistics}
\label{app:data_statistics}

Figure~\ref{fig:category_distribution} and Table~\ref{tab:ssi-taxonomy} in the main text summarize the category distribution and task taxonomy of \ssibench. 
This appendix provides additional statistics on sample counts, image usage, and prompt lengths. 
In total, the benchmark contains 1{,}000 multiple-choice ranking questions. 
Each question compares either member groups or node groups under an explicit geometric or topological criterion and requires selecting the correct permutation among 3 or 4 candidates.

Table~\ref{tab:ssi-taxonomy-stats} reports the number of questions, unique images, and template prompt length for each task category. 
Across the 1{,}000 questions, we use 1{,}160 unique images (\#UI), reflecting that the Multi-View setting consumes two views per instance and thus contributes more unique images than single-view categories. 
The average template prompt length (\#TPL) is 2{,}062 characters, measured after filling in the task-specific template text, including spaces and punctuation. 
This provides a coarse proxy for linguistic complexity and annotation density.

\begin{table}[t]
  \caption{Additional statistics of SCSR tasks in \ssibench. \#UI denotes the number of unique images. \#TPL denotes the template prompt length in characters, including spaces and punctuation.}
  \label{tab:ssi-taxonomy-stats}
  \vspace{-4pt}
  \centering
  \small
  \setlength{\tabcolsep}{12pt}
  \renewcommand{\arraystretch}{1.08}
  \begin{tabular}{llrrr}
    \toprule
    \textbf{Category} & \textbf{Sub-Category} & \textbf{Count} & \textbf{\#UI} & \textbf{\#TPL} \\
    \midrule
    \multirow[t]{7}{*}{Geometric}
      & Ground Height     & 105 & 105 & 1,662 \\
      & Ground Angle      & 101 & 101 & 1,892 \\
      & Dimension         & 100 & 100 & 1,820 \\
      & Relative Distance & 103 & 103 & 1,782 \\
      & Area              & 100 & 100 & 1,467 \\
      & Volume            & 97  & 97  & 1,489 \\
      & Multi-View        & 106 & 192 & 2,404 \\
    \midrule
    \multirow[t]{3}{*}{Topological}
      & Hop Distance      & 94 & 94  & 2,475 \\
      & Cycle Length      & 98 & 98  & 2,511 \\
      & Multi-View        & 96 & 170 & 3,158 / 3,225 \\
    \midrule
    & \textbf{Total / Avg.} & \textbf{1{,}000} & \textbf{1{,}160} & \textbf{2{,}062} \\
    \bottomrule
  \end{tabular}
  \vskip -0.2in
\end{table}

As shown in the main-text distribution figure, \ssibench spans two task families: Geometric and Topological. 
The Geometric family covers six single-view tasks plus a Multi-View variant, with 712 questions in total. 
The Topological family contains two single-view graph-based tasks and a Multi-View variant, with 288 questions in total. 
Overall, the benchmark emphasizes structure-centric geometric comparisons while preserving substantial coverage of connectivity reasoning.

The Multi-View setting is designed to test cross-view correspondence and structural consistency while keeping the queried relation well-defined. 
For the Geometric family, Multi-View is effectively the multi-view version of Relative Distance: it fuses two viewpoints to rank target members by their geometric relations to a reference member. 
For the Topological family, Multi-View includes two template variants because it covers both Hop Distance and Cycle Length under multi-view querying. 
Accordingly, Table~\ref{tab:ssi-taxonomy-stats} reports two \#TPL values for this row, corresponding to the two prompt templates used in the Multi-View topological setting. 
The full prompt templates, including the exact wording for each variant, are provided in Section~\ref{app:samples}.

The \#TPL column also shows that Topological tasks typically require longer prompts than Geometric tasks, because they must specify graph-based criteria and precisely identify queried members or groups within potentially cluttered connectivity patterns. 
Multi-View prompts are generally longer than single-view prompts due to explicit instructions for fusing information from two images and establishing correspondence to the same structural elements across views. 
These statistics reflect the intended challenge of the benchmark: solving the questions requires reconstructing a constraint-consistent 3D or graph structure rather than relying on simple 2D heuristics.

\section{Details of the Data Curation Process}
\label{app:ssi_constr_details}

\subsection{Dataset Sources}
\label{app:data_sources}

\textbf{Unsplash.}
Unsplash~\cite{unsplash-website} is a large repository of high-resolution, royalty-free photographs. We use it as a primary source to collect single-view images of real-world 3D structures with salient geometric regularities and viewpoint variation, filtering for diversity in structure types, scenes, illumination, and camera perspectives.

\textbf{Pexels.}
Pexels~\cite{pexels-website} provides curated royalty-free images with diverse compositions and environments. We use it to complement Unsplash with additional structural scenes featuring clutter, self-occlusion, and varied viewpoints, and curate candidates to reduce reliance on pixel-level shortcuts.

\textbf{Pixabay.}
Pixabay~\cite{pixabay-website} is a royalty-free media library with broad topical coverage. We use it to further increase diversity in structural instances and visual conditions (e.g., materials and lighting), and retain images that highlight constraint-governed 3D structure suitable for SCSR.

\textbf{Original Photography (Multi-View Subset).}
Because multi-view structural imagery is scarce in public sources, part of our Multi-View subset is captured by us. These paired views support evaluating cross-view correspondence and structural consistency relative to a reference member. Across all sources, selected images are high-resolution and uniformly compressed so that the longer side is at most 1920 pixels.

In total, we collect 1{,}160 candidate images from four sources: Pexels contributes 500 images (43.1\%), Unsplash contributes 474 (40.9\%), Pixabay contributes 102 (8.8\%), and our original photography contributes 84 (7.2\%).

\subsection{Data Annotation and Quality Control Interfaces}
\label{app:annotation_qc_interfaces}

\begin{figure*}
  \centering
  \includegraphics[width=0.75\textwidth]{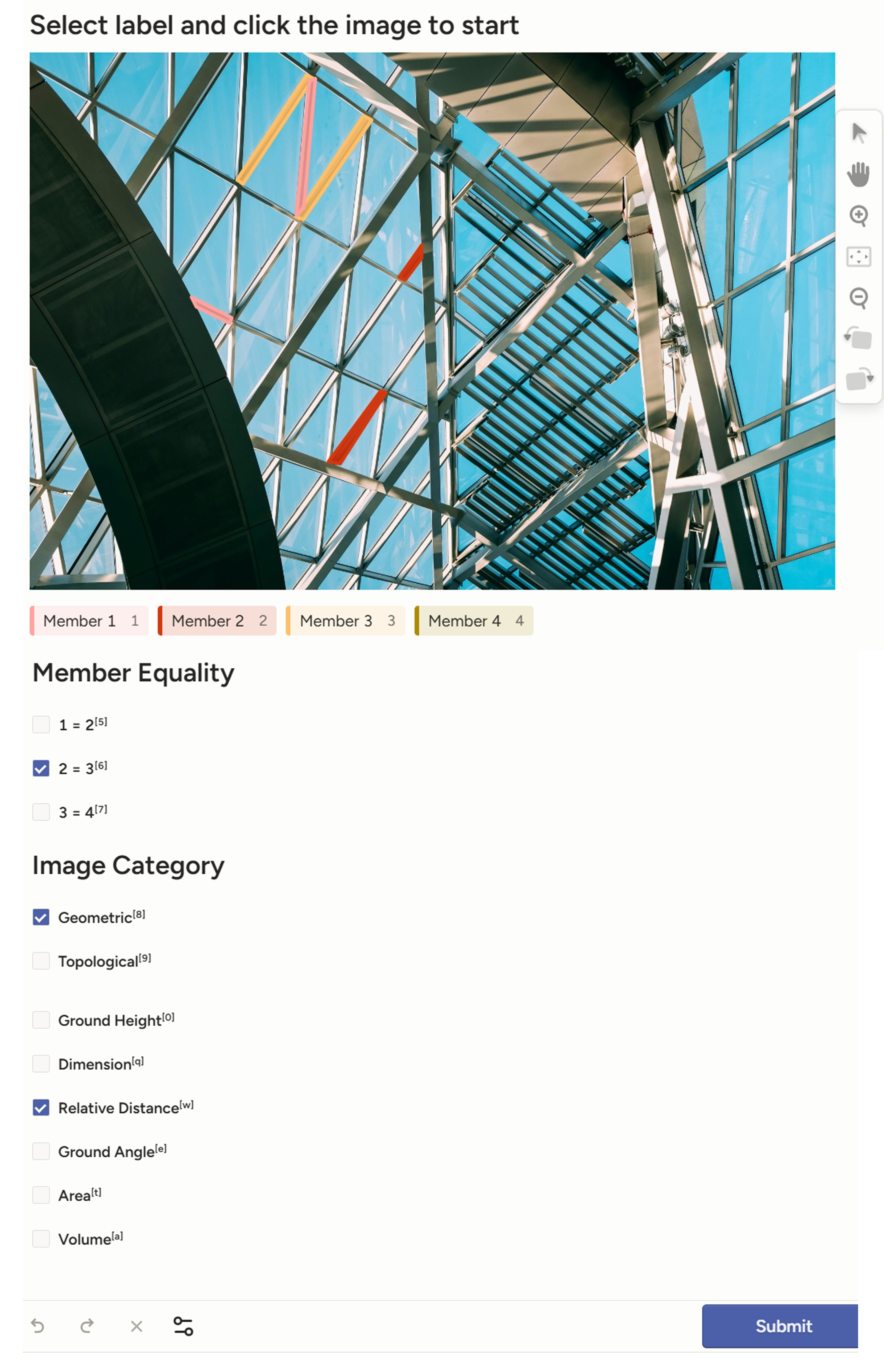}
  \caption{Screenshot of the data annotation interface, where annotators select members/groups, the task category, and specify equality relations.}
  \label{app_fig:annotation}
\end{figure*}

\begin{figure*}
  \centering
  \includegraphics[width=0.85\textwidth]{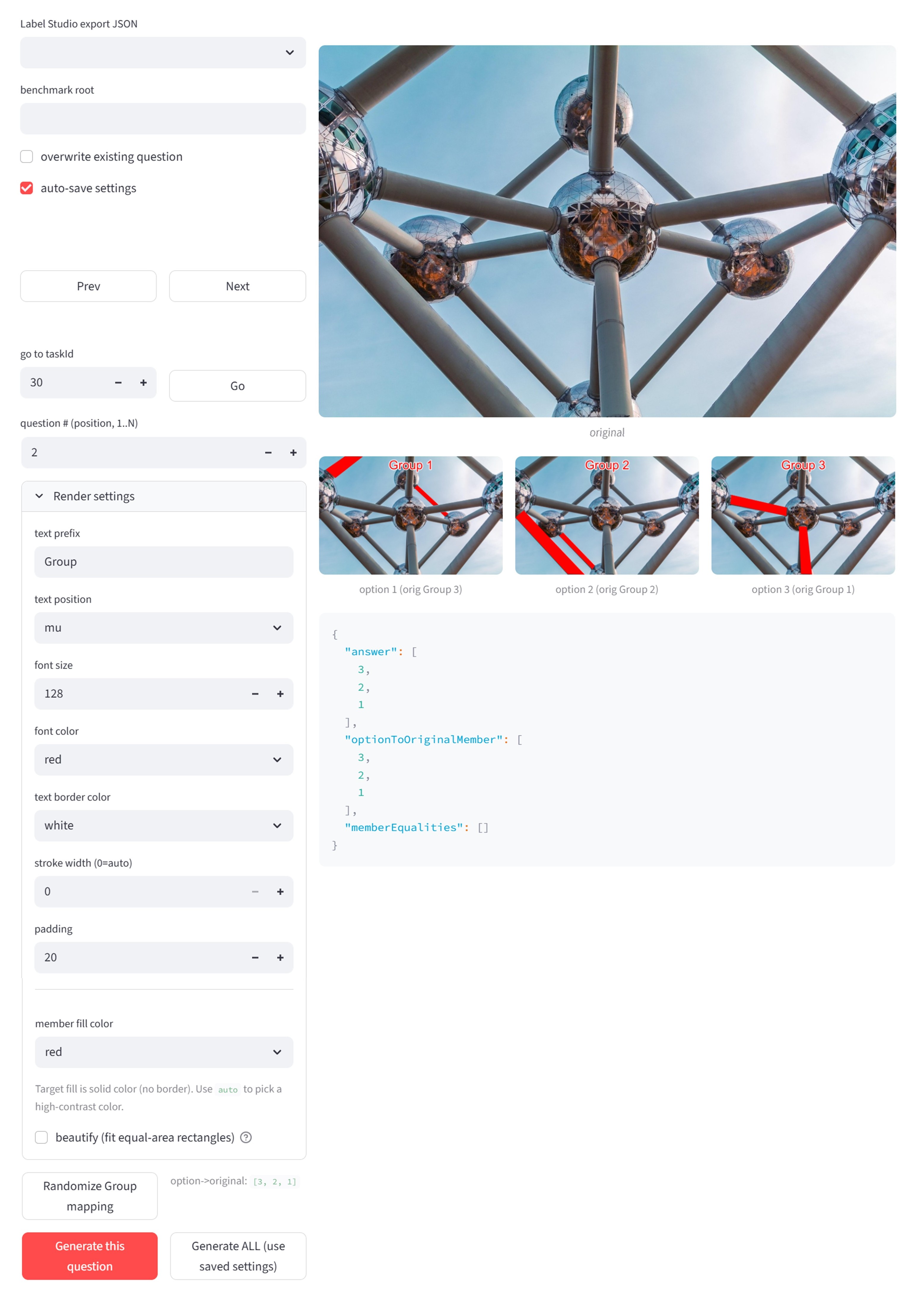}
  \caption{Screenshot of the manual review interface used during question generation, where annotators verify whether the annotated colors and text placements are appropriate.}
  \label{app_fig:generation_review}
\end{figure*}

\begin{figure*}
  \centering
  \includegraphics[width=0.85\textwidth]{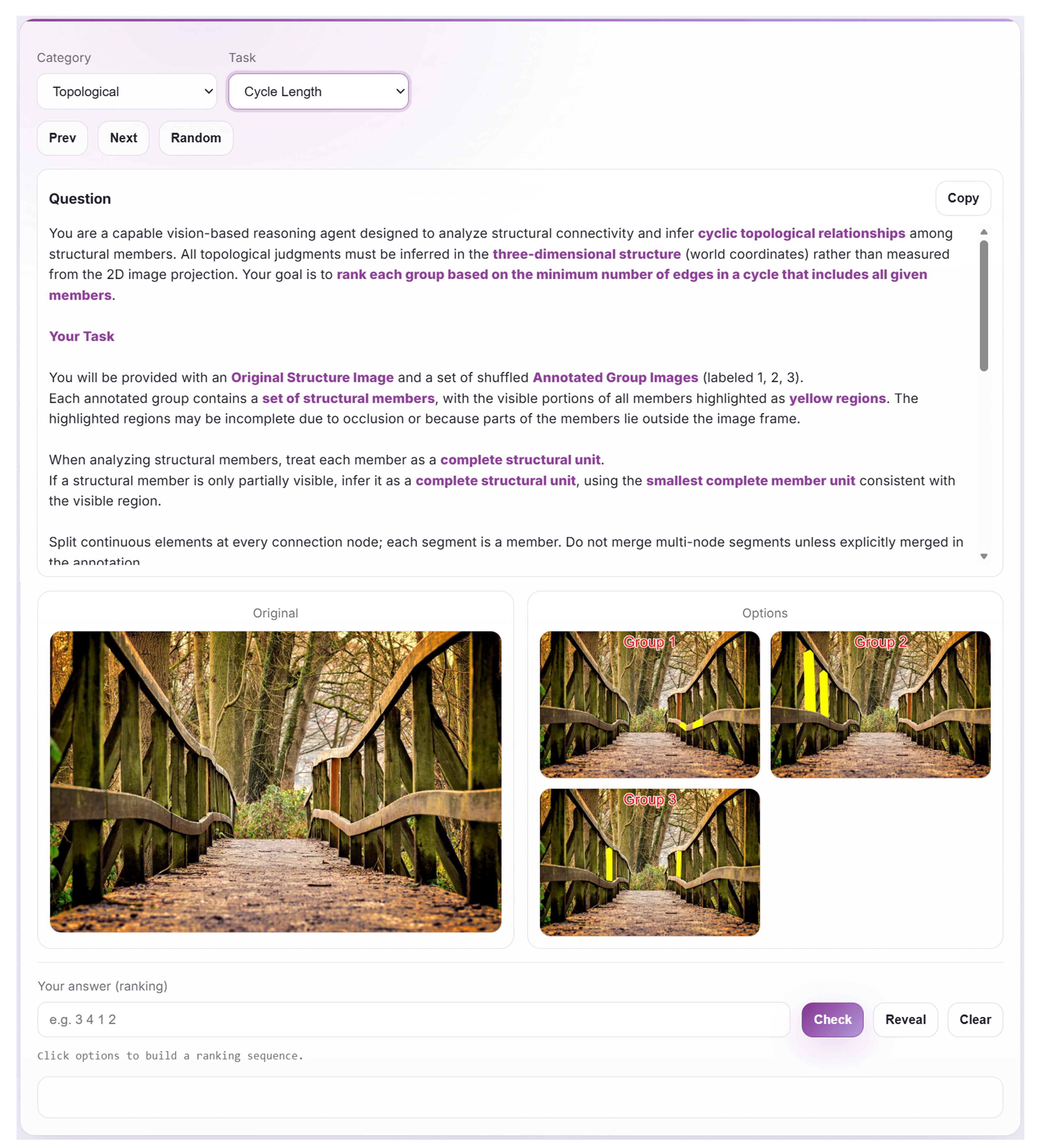}
  \caption{Screenshot of the independent review interface used for quality control, where annotators attempt the problem and provide their answers.}
  \label{app_fig:human_eval}
\end{figure*}

To ensure consistent metadata collection and rigorous quality control, we develop three lightweight, human-in-the-loop interfaces corresponding to (i) metadata annotation, (ii) manual review during question generation, and (iii) independent review for quality control. Figures~\ref{app_fig:annotation}--\ref{app_fig:human_eval} show representative screenshots.

\textbf{Metadata annotation interface.}
Figure~\ref{app_fig:annotation} shows the interface used for metadata annotation in Label Studio. For each image (or image pair in the Multi-View subset), annotators first specify the task category and then select the referenced entities (members or groups) according to the task definition. The interface supports recording the target ranking as an \emph{ascending order} under the task-specific criterion, and allows annotators to explicitly mark equality relations (ties) among candidates to avoid ambiguous supervision. In addition, annotators provide localization annotations for the referenced members/groups, which are later used to render option-specific highlighted images for question generation.

\textbf{Manual review interface for question generation.}
After instantiating questions from the annotated metadata, we perform a manual review step to verify the faithfulness and readability of visual annotations. As shown in Figure~\ref{app_fig:generation_review}, reviewers inspect each option-specific annotated image to check (1) whether highlighting colors maintain sufficient contrast against the background and other marks, and (2) whether text labels are placed appropriately (i.e., close to the intended targets, non-overlapping, and not occluded). When issues are found (e.g., low-contrast colors, labels covering key geometry, or confusing placement), reviewers revise the rendering configuration (color choice or label position) to ensure that the visualization is clear while avoiding unintended cues.

\textbf{Independent review interface for quality control.}
Figure~\ref{app_fig:human_eval} illustrates the interface used for independent quality control. For each question, a checker attempts the problem under the same format as the benchmark evaluation and provides their final answer (and, when applicable, notes about ambiguity or potential flaws). Questions with disagreements between the original annotation-derived answer key and the checker’s answer are escalated to a third reviewer for adjudication. This interface also supports structured flags for common failure modes (e.g., unclear target identification, excessive occlusion, or options that are trivially inferred from 2D cues), enabling systematic filtering and iterative refinement of the dataset.

\subsection{General Guidelines}
\label{app:general_guidelines}

We provide the following guidelines to ensure that \ssibench annotations are accurate, unambiguous, and consistently aligned with the intended evaluation of SCSR.

\begin{itemize}
    \item \textbf{Human-centric design.} All questions and annotations are created to be understandable to humans, using clear natural-language descriptions rather than technical or camera-parameter-based prompts.

    \item \textbf{Answer determinism and uniqueness.} Each instance must admit a \emph{single, well-defined} correct answer. Annotators should actively avoid ambiguous cases (e.g., unclear targets, visually indistinguishable options, or multiple plausible interpretations). This design choice differs from interaction-oriented benchmarks such as InteractComp~\cite{deng2025interactcomp}.

    \item \textbf{Moderate, meaningful difficulty.} Questions should require careful reasoning by a non-expert human, but remain solvable. Trivial shortcuts that rely on superficial 2D cues should be avoided (e.g., when the correct ordering matches the pixel-height ordering in the image without requiring 3D or structural reasoning).

    \item \textbf{Consistent candidate specification.} For member-level tasks, annotators label four candidates (1--4); for group-level tasks, annotators label three candidates (1--3). Candidate sets should be selected such that the intended comparison is non-trivial yet clearly resolvable.

    \item \textbf{Precise localization and occlusion handling.} Polygons should tightly fit the target components while respecting occlusion. If a component is occluded, annotators should mark the visible parts as separate polygons for the same target when necessary, rather than merging distinct components into one region. If a component is incomplete in the image, annotators default to the smallest identifiable unit that can be treated as an atomic element.

    \item \textbf{Strict ordering and tie annotation.} Annotations must follow the ground-truth ranking order \emph{from smallest to largest} under the specified criterion; annotators should double-check to avoid reversed ordering. If ties exist, annotators explicitly mark equality relations (e.g., selecting $1{=}2$ indicates $1{=}2{<}3{<}4$). During question generation, ties are resolved by a deterministic rule: if two candidates are indistinguishable, the smaller index is considered earlier in the ordering.

    \item \textbf{Style and rendering conventions.} Annotators use the default highlight colors in the interface; colors and label styling are standardized in a later post-processing step to ensure consistency across the dataset.

    \item \textbf{Task-specific criteria and special considerations.} Annotators must follow the task definitions and apply the following task-level notes to avoid systematic errors:
    (i) \emph{Relative Distance} is defined along the straight line of the member (i.e., the member’s axis), therefore curved components should not be selected as candidates;
    (ii) \emph{Ground Angle} measures the angle relative to the ground plane, where members parallel to the ground have the smallest angle and members perpendicular to the ground have the largest;
    (iii) \emph{Dimension} refers to the length along the member’s principal direction (i.e., its main axis);
    (iv) \emph{Cycle Length} is the number of edges in the smallest cycle that contains both members;
    (v) \emph{Hop Distance} should not be too small; candidates are chosen to ensure non-trivial graph distances so that the task remains challenging;
    (vi) \emph{Multi-View} instances require cross-view correspondence: one image labels the reference member as \texttt{0}, and the other image labels the target candidates as \texttt{1--3}. Annotators should first identify stable anchor regions visible in both views; if correspondence is unclear, the instance should be flagged for discussion.

\end{itemize}

\subsection{Data Format and Structure}
\label{app:data_format_structure}

To ensure consistency and facilitate reliable evaluation and analysis, \ssibench adopts a two-stage data organization: (i) per-instance storage for construction and inspection, and (ii) dataset-level exports for evaluation and release.

\noindent\textbf{Per-instance storage.}
\begin{itemize}
    \item \textbf{JSON metadata.} Each instance is stored as a JSON object containing the question text, answer options, the ground-truth answer, task/category labels, difficulty, and relative file paths to the associated images.
    \item \textbf{Image files.} Images are stored separately and referenced by paths in the JSON entries. For each instance, we retain both (i) the \emph{original} image(s) and (ii) the \emph{annotated} image(s) rendered with highlights and text labels for evaluation. This design preserves raw visual content while enabling reproducible benchmarking with standardized visual cues.
\end{itemize}

\noindent\textbf{Dataset-level export and release.}
\begin{itemize}
    \item \textbf{Evaluation-ready TSV.} To support automated benchmarking with VLMEvalKit \cite{duan2024vlmevalkit}, we aggregate all instances and export the benchmark into a TSV file tailored for evaluation. Each row corresponds to one instance and contains the minimal fields required for model inference and scoring, including the instance ID, question, answer label, category, and base64-encoded image content.
    \item \textbf{Hugging Face release format.} For convenient distribution and efficient loading, we additionally provide a packaged release on Hugging Face that stores the aggregated metadata and (binary-encoded) images in a Parquet-based format.
\end{itemize}

\subsection{Tie Handling and Evaluation Fairness}
\label{app:tie_handling}

Some ranking questions in \ssibench may contain candidates that are equal or visually indistinguishable under the task-defined criterion. 
These ties are not treated as annotation ambiguities; rather, they are anticipated cases handled by a deterministic rule during both annotation and evaluation. 
Specifically, when two candidates are tied, the candidate with the smaller index is placed earlier in the ranking. 
This rule is consistently applied when constructing the ground-truth answer.

To ensure fairness, the same tie-breaking rule is explicitly stated in the prompts provided to VLMs. 
Therefore, models are not penalized for correctly recognizing equality; they are expected to follow the specified output convention once equality is identified. 
We use strict exact-match evaluation instead of a tie-aware metric because accepting both possible orders for tied candidates, e.g., accepting both $A < B$ and $B < A$ when $A = B$, would make it difficult to distinguish genuine recognition of equality from arbitrary guessing. 
The deterministic rule thus provides a clear and reproducible scoring protocol while preserving the ranking-based formulation of \ssibench.

\subsection{Semi-Automatic Expansion}
\label{app:semi_auto_expansion}

Although \ssibench is constructed through a fully human-centered pipeline, its ranking-based design naturally supports semi-automatic expansion once candidate-level annotations are available. 
Instead of annotating each question from scratch, annotators can enrich an existing image with additional candidate members or groups, after which new ranking questions are generated by automatically sampling candidate combinations under the same task definitions, prompt templates, tie-breaking rule, and rendering procedure as the original benchmark.

We conduct a pilot expansion to examine this scalability. 
Starting from 100 existing images, we add two extra candidates per image and automatically sample candidate combinations, yielding 10 questions per image and 1,000 new questions in total. 
The generated questions are then checked to remove ambiguous cases, unclear visual annotations, and overly trivial instances. 
As shown in Table~\ref{tab:original-vs-new}, the newly generated split preserves the same model ranking as the original benchmark and yields highly consistent performance comparisons, suggesting that \ssibench can be scaled with substantially lower marginal annotation cost while maintaining its evaluation characteristics.

\begin{table}[t]
    \centering
    \caption{Accuracy on the original and newly generated benchmark splits. The expanded set, generated from 100 images by adding two extra candidates per image (yielding 10 questions/image), produces performance comparisons highly consistent with the original benchmark.}
    \label{tab:original-vs-new}
    \renewcommand{\arraystretch}{1.05}
    \setlength{\tabcolsep}{2pt}
    \scriptsize
    \begin{tabularx}{\textwidth}{@{}
        >{\centering\arraybackslash}p{0.9cm}
        >{\raggedright\arraybackslash}p{2.7cm}
        *{12}{Y}
        @{}}
        \toprule
        \multirow{2}{*}{\textbf{Split}} 
        & \multirow{2}{*}{\textbf{Models}} 
        & \multicolumn{7}{c}{\textbf{Geometric}} 
        & \multicolumn{3}{c}{\textbf{Topological}} 
        & \multirow{2}{*}{\textbf{Avg.}} 
        & \multirow{2}{*}{\textbf{Rank}} \\
        \cmidrule(lr){3-9} \cmidrule(lr){10-12}
        & 
        & {\tiny\textbf{Grd.\ Ht.}} 
        & {\tiny\textbf{Grd.\ Ang.}} 
        & {\tiny\textbf{Dim.}} 
        & {\tiny\textbf{Rel.\ Dist.}} 
        & {\tiny\textbf{Area}} 
        & {\tiny\textbf{Volume}} 
        & {\tiny\textbf{M-View}} 
        & {\tiny\textbf{Hop Dist.}} 
        & {\tiny\textbf{Cyc.\ Len.}} 
        & {\tiny\textbf{M-View}} 
        & & \\
        \midrule
        \multirow{3}{*}{Original}
        & Gemini-3-Flash        & 37.14 & 38.61 & 35.00 & 41.75 & 27.00 & 25.77 & 33.96 & 32.98 & 34.69 & 28.13 & 33.60 & 1 \\
        & GLM-4.6V              &  9.52 & 21.78 & 16.00 & 30.10 & 28.00 & 21.65 & 26.42 & 25.53 & 20.41 & 22.92 & 22.20 & 2 \\
        & InternVL3.5-241B-A28B &  3.81 &  6.93 &  6.00 & 28.16 & 25.00 & 16.49 & 21.70 & 28.72 & 23.47 & 23.96 & 18.30 & 3 \\
        \midrule
        \multirow{3}{*}{New}
        & Gemini-3-Flash        & 43.00 & 45.00 & 39.00 & 39.00 & 30.00 & 29.00 & 36.00 & 30.00 & 37.00 & 25.00 & 35.30 & 1 \\
        & GLM-4.6V              &  9.00 & 23.00 & 14.00 & 31.00 & 34.00 & 23.00 & 30.00 & 23.00 & 20.00 & 19.00 & 22.60 & 2 \\
        & InternVL3.5-241B-A28B &  7.00 & 12.00 &  7.00 & 23.00 & 28.00 & 18.00 & 23.00 & 24.00 & 26.00 & 23.00 & 19.10 & 3 \\
        \bottomrule
    \end{tabularx}
\end{table}

\subsection{Difficulty Annotation}
\label{app:difficulty_annotation}

To quantify instance difficulty in \ssibench in a scalable and objective manner, we use human \emph{lead time} as the primary proxy. Lead time is defined as the elapsed time (in seconds) that an annotator stays on the annotation interface to design, verify, and finalize a question instance, including selecting candidates, confirming the ordering (and ties, if any), and completing the required metadata fields. Intuitively, instances that require longer lead time tend to involve more subtle geometric/topological judgments, heavier occlusion, or more careful cross-checking to ensure answer uniqueness.

\textbf{Robustness to idle time.}
Raw lead time can be inflated by non-productive pauses (\emph{e.g.}, brief interruptions while the interface remains open). To reduce the impact of such noise, we apply robust outlier filtering based on the Interquartile Range (IQR) rule, which is less sensitive to extreme values than mean/standard-deviation-based filtering.

\textbf{Outlier filtering.}
Let $t$ denote the lead time of an annotation session. Over all collected sessions, we compute the first and third quartiles ($Q_1$ and $Q_3$) of $\{t\}$. In our data, $Q_1 = 101.46$ s and $Q_3 = 216.89$ s, yielding an interquartile range $\text{IQR} = Q_3 - Q_1 = 115.43$ s. Following the standard 1.5$\times$IQR heuristic, we define the upper outlier threshold as
\begin{equation}
    t_{\max} = Q_3 + 1.5 \times \text{IQR} = 390.03 \text{ s}.
\end{equation}
Any session with $t > t_{\max}$ is treated as an outlier and excluded from subsequent difficulty labeling, since such cases are likely dominated by idle time rather than annotation complexity. (We do not apply a lower outlier bound because extremely short sessions are typically valid and correspond to straightforward instances.)

\begin{figure*}[t]
  \centering
  \includegraphics[width=\textwidth]{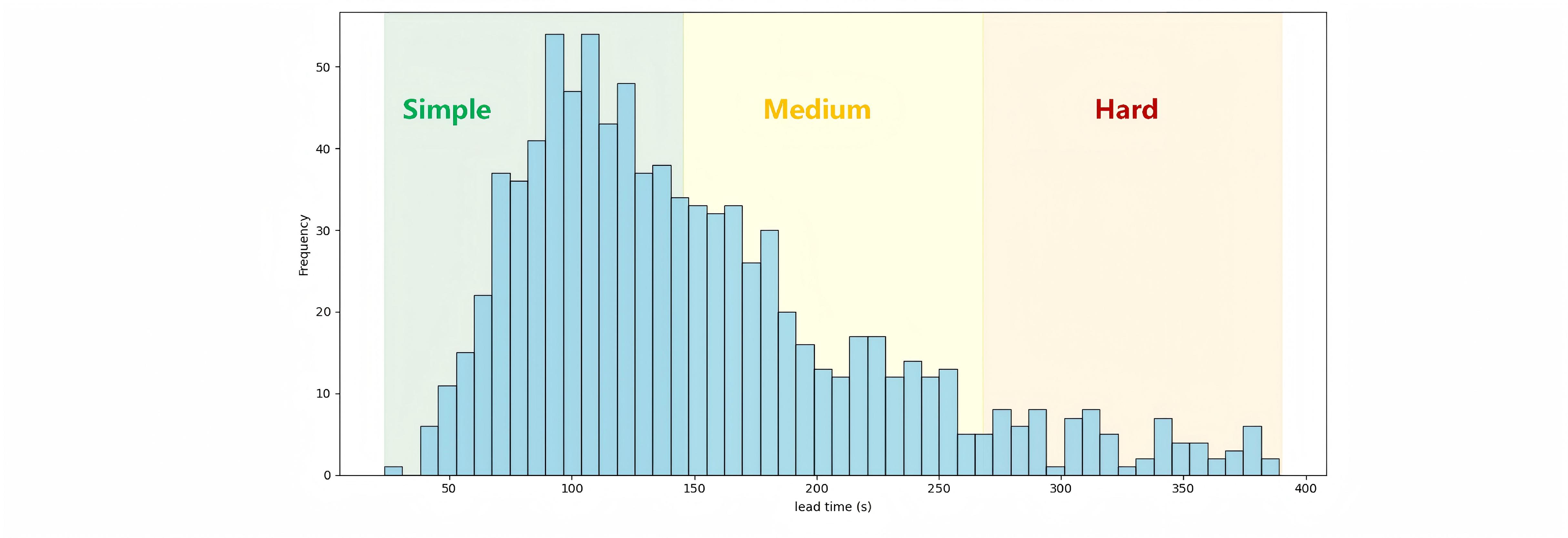}
  \caption{Distribution of human lead time for question annotation and the corresponding difficulty categorization.}
  \label{app_fig:difficult_distribution}
\end{figure*}

\textbf{Difficulty categorization.}
After removing outliers, we assign a three-level difficulty label---\textit{Easy}, \textit{Medium}, and \textit{Hard}---using a combination of (i) filtered lead time and (ii) empirical human performance during quality control. Figure~\ref{app_fig:difficult_distribution} shows the post-filtering distribution.

\begin{itemize}
    \item \textbf{Easy.} Instances with lead time in $[0, 150)$ seconds (515 instances). These questions are typically finalized quickly while still meeting our uniqueness and clarity constraints.
    \item \textbf{Medium.} Instances with lead time in $[150, 270)$ seconds (319 instances). These questions generally require additional verification, such as careful disambiguation under mild occlusion or more subtle comparisons.
    \item \textbf{Hard.} Instances that satisfy either of the following conditions (74 instances): (i) lead time in $[270, 400]$ seconds (bounded above by the outlier threshold), indicating sustained effort to ensure correctness and uniqueness; or (ii) the instance was answered incorrectly by at least one independent human reviewer during the quality control stage, suggesting that the instance is challenging even for humans under the same presentation format.
\end{itemize}

\section{Additional Implementation Details and Experimental Results} \label{app:implementation_details}

\subsection{Evaluation Metrics}
\label{app:eval_metrics}

Each benchmark instance is a ranking question with $n$ candidate options (e.g., panels / sub-images / items) indexed by $\{0,1,\dots,n-1\}$. The ground-truth answer is a permutation
$\pi^\star = [\pi^\star_1,\dots,\pi^\star_n]$,
where $\pi^\star_k$ denotes the index placed at rank $k$ (from best/most plausible to worst/least plausible, following the instruction of the corresponding QA type). The model prediction is a list
$\hat{\pi} = [\hat{\pi}_1,\dots,\hat{\pi}_n]$.

\textbf{Output parsing and validity.}
Models are instructed to output a \emph{parsable Python list} of integers. We parse the first valid list appearing in the output using a strict Python literal parser.
A prediction is considered \emph{valid} if and only if:
(i) its length equals $n$;
(ii) all entries are integers;
(iii) the set of entries equals $\{0,1,\dots,n-1\}$ (i.e., it is a permutation with no duplicates or missing indices).
If the output is not valid, we treat the instance as incorrect for all metrics (i.e., score $0$ for Taskwise Accuracy and Pairwise Accuracy). This choice avoids introducing heuristic post-processing that may unfairly benefit certain models.

\textbf{Taskwise Accuracy.}
Taskwise Accuracy measures exact match on the full permutation:
\begin{equation}
\mathrm{Acc}_{\mathrm{task}} \;=\; \frac{1}{N}\sum_{i=1}^{N} \mathbf{1}\big[\hat{\pi}^{(i)} = \pi^{\star (i)}\big],
\end{equation}
where $N$ is the number of instances, and $\mathbf{1}[\cdot]$ is the indicator function.
This metric is strict: any deviation at any position leads to $0$ for the instance.

\textbf{Pairwise Accuracy.}
Taskwise exact match can be overly harsh when a prediction is close to the correct ordering.
We therefore report Pairwise Accuracy, which evaluates \emph{pairwise ordering consistency} between the predicted ranking and the ground truth.
For any two distinct items $(a,b)$, define the induced order by a permutation $\pi$ as:
\begin{equation}
a \prec_{\pi} b \;\;\Leftrightarrow\;\; \mathrm{pos}_{\pi}(a) < \mathrm{pos}_{\pi}(b),
\end{equation}
where $\mathrm{pos}_{\pi}(x)$ returns the rank position of item $x$ in permutation $\pi$.
Pairwise Accuracy for an instance is the fraction of item pairs whose relative order matches the ground truth:
\begin{equation}
\mathrm{Acc}_{\mathrm{pair}}(\hat{\pi},\pi^\star)
\;=\;
\frac{1}{\binom{n}{2}}
\sum_{0\le a < b \le n-1}
\mathbf{1}\Big[
(a \prec_{\hat{\pi}} b \;\wedge\; a \prec_{\pi^\star} b)\;\;\vee\;\;
(b \prec_{\hat{\pi}} a \;\wedge\; b \prec_{\pi^\star} a)
\Big].
\end{equation}
We report the dataset-level Pairwise Accuracy as the average over instances:
\begin{equation}
\mathrm{Acc}_{\mathrm{pair}} \;=\; \frac{1}{N}\sum_{i=1}^{N} \mathrm{Acc}_{\mathrm{pair}}\!\left(\hat{\pi}^{(i)},\pi^{\star (i)}\right).
\end{equation}

\textbf{Random baseline.}
We report the expected performance of a uniformly random ranking over $n$ items.

\textbf{(i) Taskwise Accuracy.}
Since there are $n!$ possible permutations, the probability of exactly matching the ground-truth permutation is:
\begin{equation}
\mathbb{E}\left[\mathrm{Acc}_{\mathrm{task}}^{\mathrm{rand}}\right] \;=\; \frac{1}{n!}.
\end{equation}

\textbf{(ii) Pairwise Accuracy.}
For any pair $(a,b)$, a uniformly random permutation puts $a$ before $b$ with probability $1/2$.
Therefore, the expected fraction of correctly ordered pairs is:
\begin{equation}
\mathbb{E}\left[\mathrm{Acc}_{\mathrm{pair}}^{\mathrm{rand}}\right] \;=\; \frac{1}{2}.
\end{equation}

\textbf{Reporting.}
All metrics are computed with temperature $0$ and the same prompt template per QA type, as described in Section~\ref{evel_setting}. We report both Taskwise Accuracy and Pairwise Accuracy to capture strict end-to-end correctness and softer ordering quality, respectively.

\subsection{Benchmark Models}
\label{app:benchmark_models}
We conduct a comprehensive evaluation across 31 VLMs, covering both proprietary and open-source families, broad parameter scales, and recent architectural advances. For proprietary systems, we evaluate Google DeepMind’s Gemini series (Gemini-3-Pro and Gemini-3-Flash; Gemini-2.5-Pro and Gemini-2.5-Flash) \cite{google-gemini3-website, google-gemini25-thinking-updates-website}, OpenAI’s GPT family (GPT-5.2, GPT-5 mini, GPT-4.1, and GPT-4o) \cite{openai-gpt5.2-website, openai-gpt5-website, openai-gpt4.1-website, openai-gpt4o-website}, Anthropic’s Claude-Sonnet-4.5 \cite{anthropic-claude-sonnet45-website}, and ByteDance Seed-1.8 \cite{bytedance-seed18-website}. On the open-source side, we systematically assess 21 state-of-the-art models spanning multiple families and sizes: the GLM-V series (GLM-4.6V, GLM-4.6V-Flash, GLM-4.5V) \cite{vteam2026glm45vglm41vthinkingversatilemultimodal}, Qwen3-VL across MoE and dense variants (Qwen3-VL-235B-A22B, 30B-A3B, 8B/4B/2B) \cite{bai2025qwen3vltechnicalreport}, InternVL3.5 across a wide range of scales (InternVL3.5-241B-A28B, 30B-A3B, 38B/14B/8B/4B/2B) \cite{wang2025internvl35advancingopensourcemultimodal}, Meta’s Llama-4-Scout-17B-16E (Meta, 2025) \cite{meta-llama4-multimodal-intelligence}, Google’s Gemma-3 family (27B/12B/4B) \cite{gemmateam2025gemma3technicalreport}, and LLaVA-OneVision in 72B and 7B settings \cite{li2024llavaonevisioneasyvisualtask}.

\subsection{Implementation Details for Model Evaluation}
\label{app:model_eval_details}

\textbf{Execution setup.} Proprietary models are accessed via their official APIs to ensure standardized inference behavior and fair comparison across providers. In contrast, open-source models are deployed locally on our in-house inference stack on a single node equipped with 8$\times$ NVIDIA A100 80GB GPUs.

\textbf{Image preprocessing.} To ensure a consistent input budget across models, all images are resized such that the longer side is at most 512 pixels while preserving the aspect ratio. We apply the same preprocessing pipeline for all evaluated VLMs. This resolution is chosen to control inference cost and maintain a comparable visual-token budget across models. To verify that it does not remove essential visual information for human spatial reasoning, we conduct an additional human study on 100 randomly sampled questions, comparing the original high-resolution images with their 512px resized versions. As shown in Table~\ref{tab:human-high-vs-low-res}, human accuracy only decreases slightly from 92\% to 90\%, indicating that the 512px preprocessing preserves sufficient detail for solving \ssibench questions.

\begin{table}[t]
    \centering
    \caption{Human performance comparison on high- vs.\ low-resolution images. For each task category, 10 questions are randomly sampled, resulting in 100 questions in total.}
    \label{tab:human-high-vs-low-res}
    \renewcommand{\arraystretch}{0.95}
    \setlength{\tabcolsep}{2pt}
    \scriptsize
    \begin{tabularx}{\linewidth}{@{}>{\raggedright\arraybackslash}p{1.25cm}*{11}{Y}@{}}
        \toprule
        \multirow{2}{*}{\textbf{Resolution}} & \multicolumn{7}{c}{\textbf{Geometric}} & \multicolumn{3}{c}{\textbf{Topological}} & \multirow{2}{*}{\textbf{Avg.}} \\
        \cmidrule(lr){2-8} \cmidrule(lr){9-11}
        & {\scriptsize\textbf{Grd.\ Ht.}} 
        & {\scriptsize\textbf{Grd.\ Ang.}} 
        & {\scriptsize\textbf{Dim.}} 
        & {\scriptsize\textbf{Rel.\ Dist.}} 
        & {\scriptsize\textbf{Area}} 
        & {\scriptsize\textbf{Volume}} 
        & {\scriptsize\textbf{M-View}} 
        & {\scriptsize\textbf{Hop Dist.}} 
        & {\scriptsize\textbf{Cyc.\ Len.}} 
        & {\scriptsize\textbf{M-View}} 
        & \\
        \midrule
        1920px & 90 & 90 & 100 & 90 & 90 & 100 & 80 & 100 & 90 & 90 & 92 \\
        512px  & 100 & 90 & 90 & 90 & 80 & 100 & 80 & 90 & 90 & 90 & 90 \\
        \bottomrule
    \end{tabularx}
\end{table}

\textbf{Evaluation framework.} We implement the evaluation pipeline using \textsc{VLMEvalKit}, which standardizes prompt construction, model invocation, output parsing, and metric computation across different backends (API-based and locally deployed).

\begin{table*}[t]
    \caption{Evaluation on \ssibench (Pairwise Accuracy). We highlight the \shadeword{bestbg}{best} and \shadeword{secondbg}{second best} results within each category (Proprietary or Open-source Models). Abbreviations: Grd. Ht. = Ground Height; Grd. Ang. = Ground Angle; Dim. = Dimension; Rel. Dist. = Relative Distance; Area = Area; Volume = Volume; M-View = Multi-View; Hop Dist. = Hop Distance; Cyc. Len = Cycle Length.}
  \label{tab:geom-topo-pair}
  \vspace{-9pt}
  \setlength{\fboxsep}{1.45pt}

  \setlength{\tabcolsep}{0pt}
  \renewcommand{\arraystretch}{1}

  \begin{center}
    \begin{footnotesize}
      \begin{sc}
        \begin{tabular}{ l *{11}{>{\centering\arraybackslash}p{0.07\textwidth}} }
        \toprule
        \multirow{2}{*}[-0.5ex]{\textbf{Models}} & \multicolumn{7}{c}{\textbf{Geometric}} & \multicolumn{3}{c}{\textbf{Topological}} & \multirow{2}{*}[-0.5ex]{\textbf{Avg.}} \\
        \cmidrule(lr){2-8} \cmidrule(lr){9-11} & {\scriptsize\normalfont\textbf{Grd. Ht.}} & {\scriptsize\normalfont\textbf{Grd. Ang.}} & {\scriptsize\normalfont\textbf{Dim.}} & {\scriptsize\normalfont\textbf{Rel. Dist.}} & {\scriptsize\normalfont\textbf{Area}} & {\scriptsize\normalfont\textbf{Volume}} & {\scriptsize\normalfont\textbf{M-View}} & {\scriptsize\normalfont\textbf{Hop Dist.}} & {\scriptsize\normalfont\textbf{Cyc. Len.}} & {\scriptsize\normalfont\textbf{M-View}} & \\
        \midrule

          \rowcolor{groupbg}
          \multicolumn{12}{l}{\textit{Proprietary Models}} \\
          Gemini-3-Pro       & 69.52 & \colorbox{secondbg}{78.38} & 74.00 & 66.99 & \colorbox{secondbg}{56.67} & \colorbox{bestbg}{59.79} & 61.64 & \colorbox{secondbg}{61.35} & 62.93 & 59.38 & 65.15 \\
          Gemini-3-Flash     & \colorbox{bestbg}{76.03} & 78.22 & \colorbox{secondbg}{74.50} & 68.93 & 56.33 & \colorbox{secondbg}{58.08} & \colorbox{secondbg}{62.26} & \colorbox{bestbg}{64.54} & \colorbox{secondbg}{66.67} & \colorbox{secondbg}{60.76} & \colorbox{bestbg}{66.73} \\
          Gemini-2.5-Pro     & 66.98 & 76.24 & 64.17 & 65.70 & 50.33 & 55.67 & \colorbox{secondbg}{62.26} & 56.74 & 62.24 & 58.33 & 61.98 \\
          Gemini-2.5-Flash   & 67.30 & 70.13 & 66.00 & 55.02 & 52.67 & 46.39 & 59.43 & 54.96 & 58.84 & 58.68 & 59.05 \\
          GPT-5.2            & \colorbox{secondbg}{72.86} & \colorbox{bestbg}{78.88} & \colorbox{bestbg}{77.67} & \colorbox{bestbg}{72.17} & 56.00 & 51.20 & \colorbox{bestbg}{63.52} & \colorbox{secondbg}{61.35} & 65.99 & 57.29 & \colorbox{secondbg}{65.85} \\
          GPT-5 mini         & 69.68 & 74.09 & 67.67 & \colorbox{secondbg}{71.52} & \colorbox{bestbg}{58.67} & 49.83 & 55.03 & 53.55 & 57.82 & 58.68 & 61.80 \\
          GPT-4.1            & 63.33 & 59.74 & 68.17 & 52.43 & \colorbox{secondbg}{56.67} & 50.52 & 49.37 & 58.51 & 56.80 & 58.33 & 57.37 \\
          GPT-4o             & 66.67 & 63.70 & 65.33 & 57.93 & 54.00 & 52.92 & 50.00 & 57.45 & 54.42 & \colorbox{secondbg}{60.76} & 58.33 \\
          Claude-Sonnet-4.5  & 53.49 & 59.90 & 64.17 & 61.49 & 49.67 & 47.42 & 53.46 & 53.19 & 55.78 & 57.29 & 55.62 \\
          Seed-1.8           & 67.46 & 74.09 & 69.33 & 69.58 & 48.67 & 50.86 & 57.55 & 60.99 & \colorbox{bestbg}{67.69} & \colorbox{bestbg}{61.11} & 62.80 \\

          \midrule
        \rowcolor{groupbg}\multicolumn{12}{l}{\textit{Open-source Models}} \\
          GLM-4.6V           & 63.65 & \colorbox{secondbg}{69.47} & 64.83 & \colorbox{bestbg}{62.78} & 60.33 & \colorbox{secondbg}{50.86} & \colorbox{bestbg}{58.18} & 58.16 & 55.10 & 55.90 & \colorbox{secondbg}{60.02} \\
          GLM-4.6V-Flash     & \colorbox{secondbg}{64.92} & 65.02 & 62.00 & 59.55 & 59.00 & \colorbox{bestbg}{51.55} & 51.89 & 59.22 & 54.42 & \colorbox{bestbg}{60.42} & 58.82 \\
          GLM-4.5V           & \colorbox{bestbg}{66.67} & \colorbox{bestbg}{73.60} & \colorbox{secondbg}{67.33} & \colorbox{secondbg}{59.87} & 59.67 & 49.83 & 53.46 & 58.16 & 57.14 & 57.99 & \colorbox{bestbg}{60.43} \\
          Qwen3-VL-235B-A22B & 62.22 & 67.49 & \colorbox{bestbg}{67.83} & \colorbox{secondbg}{59.87} & 55.33 & 46.05 & 54.72 & \colorbox{bestbg}{62.41} & 53.40 & 57.99 & 58.77 \\
          Qwen3-VL-30B-A3B   & 54.60 & 53.47 & 64.17 & 58.58 & 61.33 & \colorbox{secondbg}{50.86} & 54.09 & 59.57 & 60.54 & \colorbox{secondbg}{59.38} & 57.62 \\
          Qwen3-VL-8B        & 54.29 & 50.83 & 61.00 & 58.25 & 56.00 & \colorbox{bestbg}{51.55} & \colorbox{secondbg}{57.55} & 56.38 & 60.54 & 58.33 & 56.47 \\
          Qwen3-VL-4B        & 55.56 & 52.31 & 59.33 & 58.58 & 57.67 & 50.52 & 55.97 & 59.57 & \colorbox{bestbg}{61.56} & 57.99 & 56.88 \\
          Qwen3-VL-2B        & 53.17 & 52.81 & 59.17 & 59.22 & \colorbox{secondbg}{63.33} & 50.52 & 44.65 & \colorbox{secondbg}{61.70} & 60.54 & 26.04 & 53.13 \\
          InternVL3.5-241B-A28B & 50.63 & 52.31 & 60.00 & 59.55 & 59.00 & 48.45 & 52.52 & 60.28 & 54.76 & 52.08 & 54.93 \\
          InternVL3.5-30B-A3B & 54.60 & 52.64 & 60.00 & 58.90 & \colorbox{secondbg}{63.33} & 50.17 & 38.68 & 60.28 & 60.54 & 58.68 & 55.65 \\
          InternVL3.5-38B    & 55.56 & 52.15 & 66.00 & 56.63 & \colorbox{secondbg}{63.33} & 50.17 & 55.66 & 59.93 & 55.44 & 58.33 & 57.30 \\
          InternVL3.5-14B    & 53.33 & 50.83 & 59.50 & 50.81 & 60.33 & \colorbox{secondbg}{50.86} & 56.60 & 53.90 & 55.44 & 57.99 & 54.95 \\
          InternVL3.5-8B     & 50.95 & 52.15 & 54.67 & 53.72 & \colorbox{bestbg}{63.67} & 50.52 & \colorbox{bestbg}{58.18} & 58.16 & 60.54 & 57.99 & 56.02 \\
          InternVL3.5-4B     & 54.29 & 51.49 & 55.67 & 54.37 & 61.00 & 50.17 & 30.82 & 52.84 & 59.18 & 50.35 & 51.90 \\
          InternVL3.5-2B     & 50.63 & 46.53 & 55.50 & 49.84 & 58.67 & 42.96 & 16.35 & 49.65 & 34.69 & 1.74 & 40.70 \\
          Llama-4-Scout-17B-16E & 56.98 & 68.48 & 64.50 & 57.61 & 62.67 & 47.08 & 56.60 & 56.74 & 59.86 & 59.03 & 58.98 \\
          Gemma-3-27B        & 51.75 & 51.98 & 57.33 & 58.25 & \colorbox{secondbg}{63.33} & 50.17 & 55.97 & 59.57 & \colorbox{secondbg}{61.22} & 58.33 & 56.75 \\
          Gemma-3-12B        & 46.83 & 50.00 & 47.67 & 47.90 & 58.67 & 48.80 & 56.60 & 57.45 & \colorbox{secondbg}{61.22} & 59.03 & 53.33 \\
          Gemma-3-4B         & 54.92 & 52.64 & 58.50 & 56.31 & \colorbox{bestbg}{63.67} & 50.52 & 56.92 & 59.93 & 54.42 & 56.60 & 56.43 \\
          LLaVA-Onevision-72B & 49.68 & 51.98 & 50.50 & 50.81 & 62.00 & 49.48 & 56.92 & 54.26 & 55.10 & \colorbox{secondbg}{59.38} & 53.98 \\
          LLaVA-Onevision-7B & 52.22 & 50.66 & 45.83 & 49.84 & 58.00 & 48.45 & 57.23 & 58.51 & 54.76 & 56.25 & 53.15 \\

          \midrule
          \rowcolor{groupbg}
          \multicolumn{12}{l}{\textit{Baselines}} \\
          Random Guessing     & 50.00 & 50.00 & 50.00 & 50.00 & 50.00 & 50.00 & 50.00 & 50.00 & 50.00 & 50.00 & 50.00 \\
        \textbf{Human Performance}
        & \textbf{99.05} & \textbf{98.18} & \textbf{97.83} & \textbf{97.09}
        & \textbf{99.00} & \textbf{95.88} & \textbf{96.23}
        & \textbf{97.16} & \textbf{95.58} & \textbf{90.97} & \textbf{96.73} \\

          \bottomrule
        \end{tabular}
      \end{sc}
    \end{footnotesize}
  \end{center}
  \vskip -0.2in
\end{table*}

\subsection{Human Evaluation Setup}
\label{app:human_eval_setup}

Human performance is measured as the average accuracy of six adult participants who were not involved in the data annotation process. The six participants collectively completed all 1,000 benchmark questions. During evaluation, participants are presented with the question and the corresponding images simultaneously (including the original image and all annotated variants, identical to the visual inputs provided to the VLMs). They are given unlimited time to answer each question to the best of their ability and may revisit the images as many times as needed. To facilitate answer submission, participants provide their rankings by clicking the options in order to form a permutation.

\subsection{Generality Analysis}
\label{app:generality_analysis}

Although \ssibench focuses on structure-centric spatial reasoning, it is not intended to evaluate specialized engineering expertise. 
We examine this from two perspectives: human performance across different backgrounds and model-performance consistency with general spatial benchmarks.

\paragraph{Non-expert human performance.}
To examine whether \ssibench requires specialized engineering expertise, we further conduct a pilot study with non-expert participants who do not have structural-engineering backgrounds. 
Before testing, they receive only a brief introduction to basic structural common sense, lasting less than 10 minutes. 
We randomly sample 10 questions from each task category, resulting in 100 questions in total. 
As shown in Table~\ref{tab:human-background-comparison}, non-expert participants achieve 91\% accuracy, which is close to the 92\% accuracy of participants with basic engineering background on the same subset. 
This suggests that \ssibench primarily evaluates spatial reasoning rather than domain-specific engineering knowledge. 
In particular, Area and Volume questions do not require advanced engineering computation. 
They are designed so that humans can solve them through intuitive 3D comparison after identifying the relevant structural components.

\begin{table}[t]
    \centering
    \caption{Performance of human evaluators on \ssibench. The non-expert group received a brief introduction under 10 minutes to structural common sense prior to testing. For each task category, 10 questions were randomly sampled, resulting in 100 questions in total.}
    \label{tab:human-background-comparison}
    \renewcommand{\arraystretch}{0.95}
    \setlength{\tabcolsep}{2pt}
    \scriptsize
    \begin{tabularx}{\linewidth}{@{}>{\raggedright\arraybackslash}p{2.5cm}*{11}{Y}@{}}
        \toprule
        \multirow{2}{*}{\textbf{Participant Group}} 
        & \multicolumn{7}{c}{\textbf{Geometric}} 
        & \multicolumn{3}{c}{\textbf{Topological}} 
        & \multirow{2}{*}{\textbf{Avg.}} \\
        \cmidrule(lr){2-8} \cmidrule(lr){9-11}
        & {\scriptsize\textbf{Grd.\ Ht.}} 
        & {\scriptsize\textbf{Grd.\ Ang.}} 
        & {\scriptsize\textbf{Dim.}} 
        & {\scriptsize\textbf{Rel.\ Dist.}} 
        & {\scriptsize\textbf{Area}} 
        & {\scriptsize\textbf{Volume}} 
        & {\scriptsize\textbf{M-View}} 
        & {\scriptsize\textbf{Hop Dist.}} 
        & {\scriptsize\textbf{Cyc.\ Len.}} 
        & {\scriptsize\textbf{M-View}} 
        & \\
        \midrule
        Basic Eng.\ Background & 90 & 90 & 100 & 90 & 90 & 100 & 80 & 100 & 90 & 90 & 92 \\
        Non-Expert (Layperson) & 90 & 80 & 100 & 90 & 80 & 90 & 90 & 100 & 100 & 90 & 91 \\
        \bottomrule
    \end{tabularx}
\end{table}

\paragraph{Consistency with general spatial benchmarks.}
We further compare model performance on \ssibench with reported results on representative spatial benchmarks that cover broader scene-centric or video-based settings. 
As shown in Table~\ref{tab:correlation}, models with stronger performance on \ssibench generally also perform better on these existing benchmarks when overlapping results are available. 
This trend suggests that \ssibench captures a relevant dimension of general spatial intelligence, while still testing a complementary structure-centric capability that is underrepresented in prior evaluations. 
Because public reports cover different model subsets across benchmarks, we treat this comparison as a consistency analysis rather than a definitive correlation estimate.

\begin{table}[t]
    \centering
    \scriptsize
    \caption{Performance of various models on \ssibench and other general spatial benchmarks. The models are sorted from left to right in descending order based on their performance on \ssibench. ``--'' indicates that the model's performance on the corresponding benchmark was not reported.}
    \label{tab:correlation}
    \renewcommand{\arraystretch}{1.05}
    \setlength{\tabcolsep}{2pt}
    \begin{tabularx}{\textwidth}{@{}l *{5}{>{\centering\arraybackslash}X}@{}}
        \toprule
        \textbf{Benchmark} & \textbf{Gemini-3-Pro} & \textbf{Gemini-2.5-Pro} & \textbf{GPT-4o} & \textbf{InternVL3.5-8B} & \textbf{LLaVA-Onevision-7B} \\
        \midrule
        SSI-Bench         & 29.50 & 26.10 & 22.60 & 20.20 & 16.50 \\
        DSI-Bench         & --    & 46.90 & 37.23 & 36.41 & --    \\
        MMSI-Video-Bench  & 38.00 & --    & 31.60 & --    & --    \\
        MMSI-Bench        & --    & 36.90 & 30.30 & --    & 24.50 \\
        ViewSpatial-Bench & --    & --    & 34.98 & --    & 27.49 \\
        CVBench           & --    & --    & 69.10 & --    & 52.60 \\
        VSI-Bench         & --    & --    & 34.00 & --    & 32.40 \\
        \bottomrule
    \end{tabularx}
\end{table}

\subsection{Pairwise Accuracy Results}
\label{app:pairwise_results}

We additionally report \textbf{Pairwise Accuracy} results on \ssibench in Table~\ref{tab:geom-topo-pair}. Pairwise Accuracy evaluates models in a two-choice, comparative setting using the same geometric and topological attributes as in the main table. We retain the same model grouping (Proprietary vs.\ Open-source) and highlight the best and second-best performance within each group to enable fair, within-group comparisons.

Overall, \textbf{Pairwise Accuracy and Taskwise Accuracy lead to highly consistent conclusions about model capability}. Within the proprietary group, models that perform strongly under Taskwise Accuracy remain strong under Pairwise Accuracy (e.g., Gemini-3-Flash continues to achieve the best average performance), while weaker models remain near the bottom (e.g., Claude-Sonnet-4.5). This consistency suggests that the two evaluation protocols largely agree on global model rankings, and that Pairwise Accuracy provides a reliable complementary perspective on the same underlying skills.

At the same time, we observe \textbf{localized discrepancies}, primarily among mid-tier models. A key reason is that Taskwise Accuracy typically requires a model to resolve all relevant comparisons within a problem correctly (e.g., producing a globally consistent ordering across multiple pairs), whereas Pairwise Accuracy scores each comparison independently. As a result, some mid-tier models can achieve reasonably high Pairwise Accuracy by getting many individual pairwise judgments right, yet still underperform on Taskwise Accuracy because they fail to handle specific “trap” cases or corner conditions embedded in the full problem (e.g., subtle ambiguities, distractors, or exceptions that break global consistency). In this sense, Taskwise Accuracy places greater emphasis on robust pitfall recognition and end-to-end consistency, while Pairwise Accuracy more directly reflects local comparative reasoning. Taken together, the results suggest that the two metrics are broadly aligned yet complementary: Taskwise Accuracy is stricter due to its holistic correctness requirement, whereas Pairwise Accuracy better isolates comparison-based geometric/topological intuition.

\subsection{Complete Evaluation Results of Thinking}
\label{app:thinking_results}

\begin{figure*}[t]
  \centering
  \includegraphics[width=\textwidth]{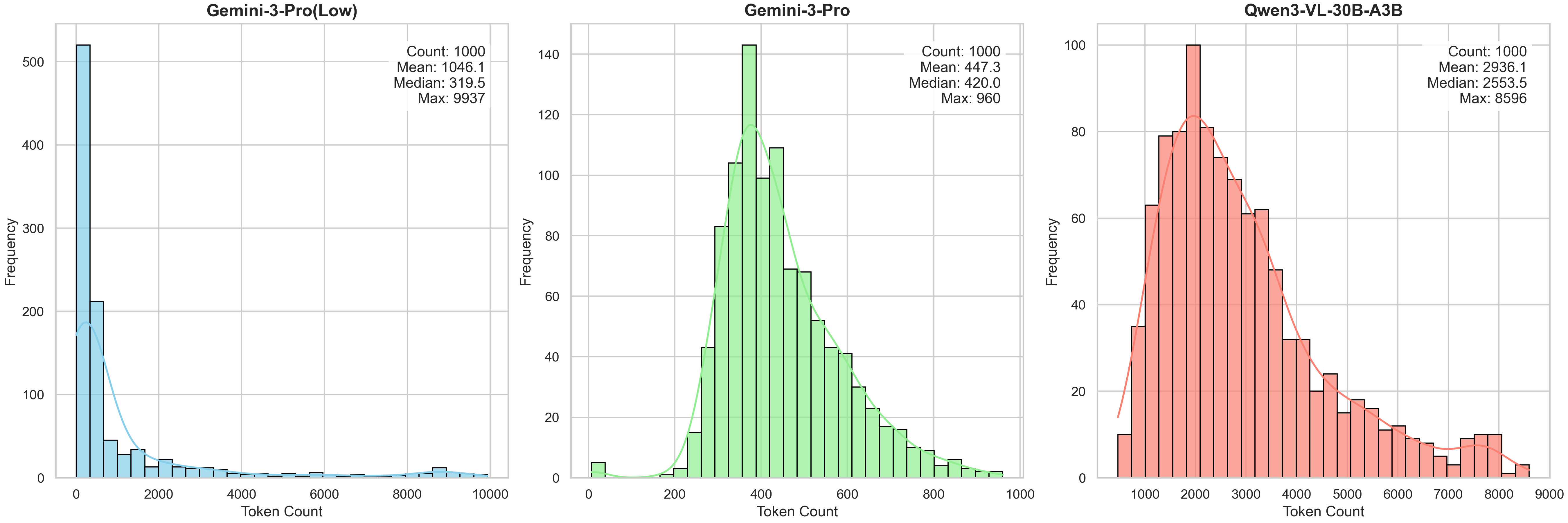}
  \caption{Distributions of the number of tokens used for explicit thinking for different models.}
  \label{app_fig:token_distribution}
\end{figure*}

As shown in Figure~\ref{app_fig:token_distribution}, different models exhibit markedly different distributions in the number of tokens used for explicit thinking, indicating diverse reasoning budgets and generation behaviors.

Table~\ref{tab:complete_think_results} reports the complete task accuracy on \ssibench for a set of representative thinking-enabled and instruct-style variants. We organize the results into two groups and highlight the best and second-best scores \emph{within each group} for every sub-task and the overall average. For Gemini-3-Pro, the \textit{high} setting yields higher accuracy on most geometric and topological dimensions, while the \textit{low} setting remains competitive and even performs better on \textit{Volume}, \textit{Geometric M-View}, and \textit{Topological M-View}. For Qwen3-VL-30B-A3B, the \textit{Thinking} variant provides consistent gains over the \textit{Instruct} variant on the majority of sub-tasks as well as the overall average, whereas \textit{Instruct} shows advantages on \textit{Area}, \textit{Geometric M-View}, and \textit{Hop Dist.}.

\begin{table*}[t]
    \caption{Complete evaluation results of thinking on \ssibench (Task Accuracy). Abbreviations: Grd. Ht. = Ground Height; Grd. Ang. = Ground Angle; Dim. = Dimension; Rel. Dist. = Relative Distance; Area = Area; Volume = Volume; M-View = Multi-View; Hop Dist. = Hop Distance; Cyc. Len = Cycle Length.}
  \label{tab:complete_think_results}
  \vspace{-9pt}
  \setlength{\fboxsep}{1.45pt}

  \setlength{\tabcolsep}{0pt}
  \renewcommand{\arraystretch}{1}

  \begin{center}
    \begin{scriptsize}
      \begin{sc}
        \begin{tabular}{ l *{11}{>{\centering\arraybackslash}p{0.07\textwidth}} }
        \toprule
        \multirow{2}{*}[-0.5ex]{\textbf{Models}} & \multicolumn{7}{c}{\textbf{Geometric}} & \multicolumn{3}{c}{\textbf{Topological}} & \multirow{2}{*}[-0.5ex]{\textbf{Avg.}} \\
        \cmidrule(lr){2-8} \cmidrule(lr){9-11} & {\scriptsize\normalfont\textbf{Grd. Ht.}} & {\scriptsize\normalfont\textbf{Grd. Ang.}} & {\scriptsize\normalfont\textbf{Dim.}} & {\scriptsize\normalfont\textbf{Rel. Dist.}} & {\scriptsize\normalfont\textbf{Area}} & {\scriptsize\normalfont\textbf{Volume}} & {\scriptsize\normalfont\textbf{M-View}} & {\scriptsize\normalfont\textbf{Hop Dist.}} & {\scriptsize\normalfont\textbf{Cyc. Len.}} & {\scriptsize\normalfont\textbf{M-View}} & \\
        \midrule

        Gemini-3-Pro (high) & \colorbox{bestbg}{25.71} & \colorbox{bestbg}{37.62} & \colorbox{bestbg}{28.00} & \colorbox{bestbg}{33.01} & \colorbox{bestbg}{24.00} & \colorbox{secondbg}{27.84} & \colorbox{secondbg}{31.13} & \colorbox{bestbg}{35.11} & \colorbox{bestbg}{30.61} & \colorbox{secondbg}{21.88} & \colorbox{bestbg}{29.50} \\
        Gemini-3-Pro (low)  & \colorbox{secondbg}{22.86} & \colorbox{secondbg}{30.69} & \colorbox{secondbg}{27.00} & \colorbox{secondbg}{32.04} & \colorbox{secondbg}{19.00} & \colorbox{bestbg}{30.93} & \colorbox{bestbg}{32.08} & \colorbox{secondbg}{21.28} & \colorbox{secondbg}{27.55} & \colorbox{bestbg}{27.08} & \colorbox{secondbg}{27.10} \\

        \midrule

        Qwen3-VL-30B-A3B-Thinking  & \colorbox{bestbg}{5.71} & \colorbox{bestbg}{12.87} & \colorbox{bestbg}{14.00} & \colorbox{bestbg}{33.01} & \colorbox{secondbg}{28.00} & \colorbox{bestbg}{21.65} & \colorbox{secondbg}{21.70} & \colorbox{secondbg}{27.66} & \colorbox{bestbg}{30.61} & \colorbox{bestbg}{30.21} & \colorbox{bestbg}{22.50} \\
        Qwen3-VL-30B-A3B-Instruct  & \colorbox{secondbg}{5.71} & \colorbox{secondbg}{7.92} & \colorbox{secondbg}{12.00} & \colorbox{secondbg}{28.16} & \colorbox{bestbg}{29.00} & \colorbox{secondbg}{17.53} & \colorbox{bestbg}{22.64} & \colorbox{bestbg}{28.72} & \colorbox{secondbg}{28.57} & \colorbox{secondbg}{27.08} & \colorbox{secondbg}{20.60} \\

        \bottomrule
        \end{tabular}
      \end{sc}
    \end{scriptsize}
  \end{center}
  \vskip -0.2in
\end{table*}

\section{Representative \ssibench Samples from Each Category}
\label{app:samples}
\begin{figure*}[!b]
  \centering
  \includegraphics[width=\textwidth]{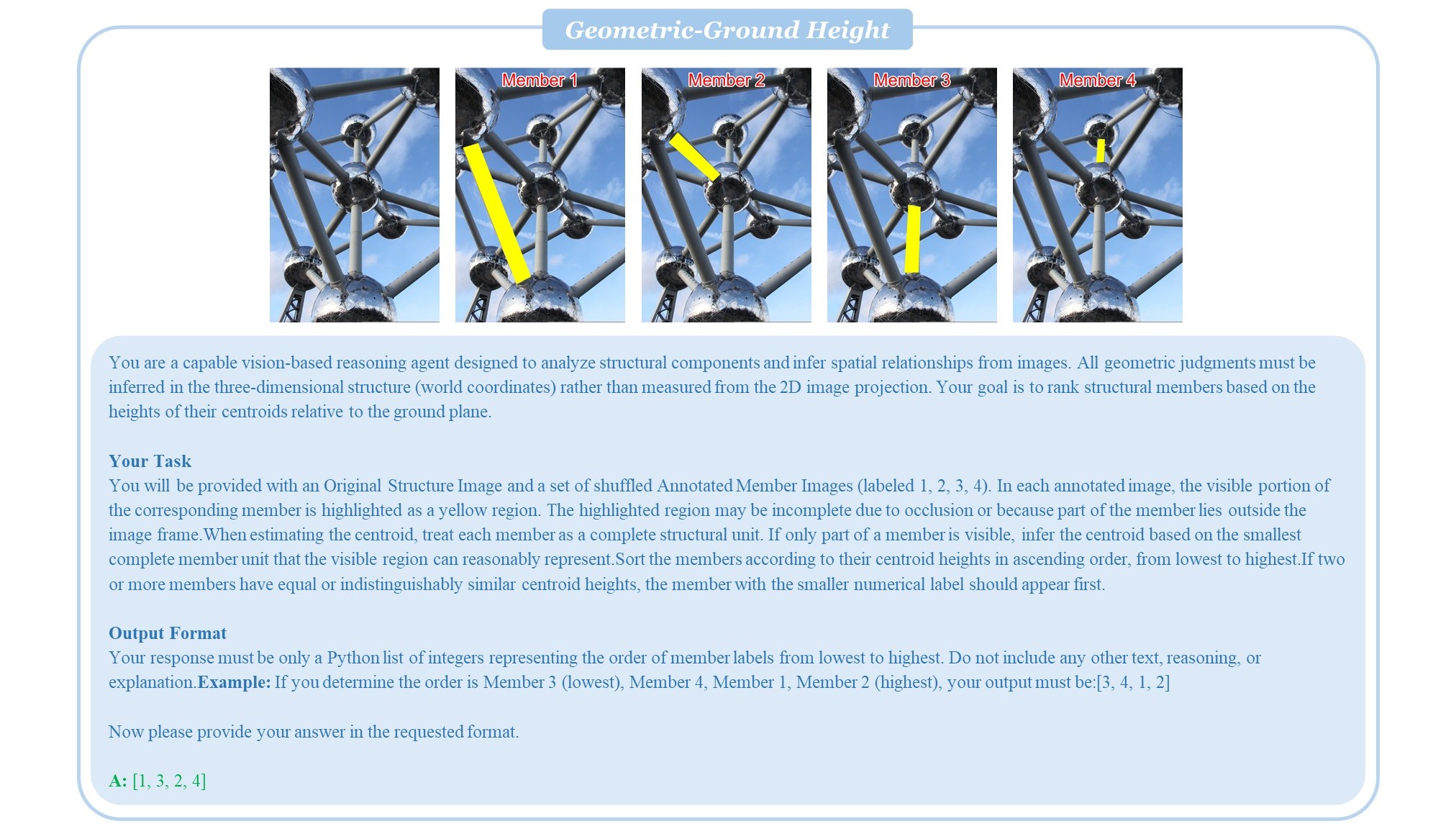}
  \caption{Full example for geometric-ground height category.}
  \label{app_fig:sample8}
\end{figure*}

\begin{figure*}
  \centering
  \includegraphics[width=\textwidth]{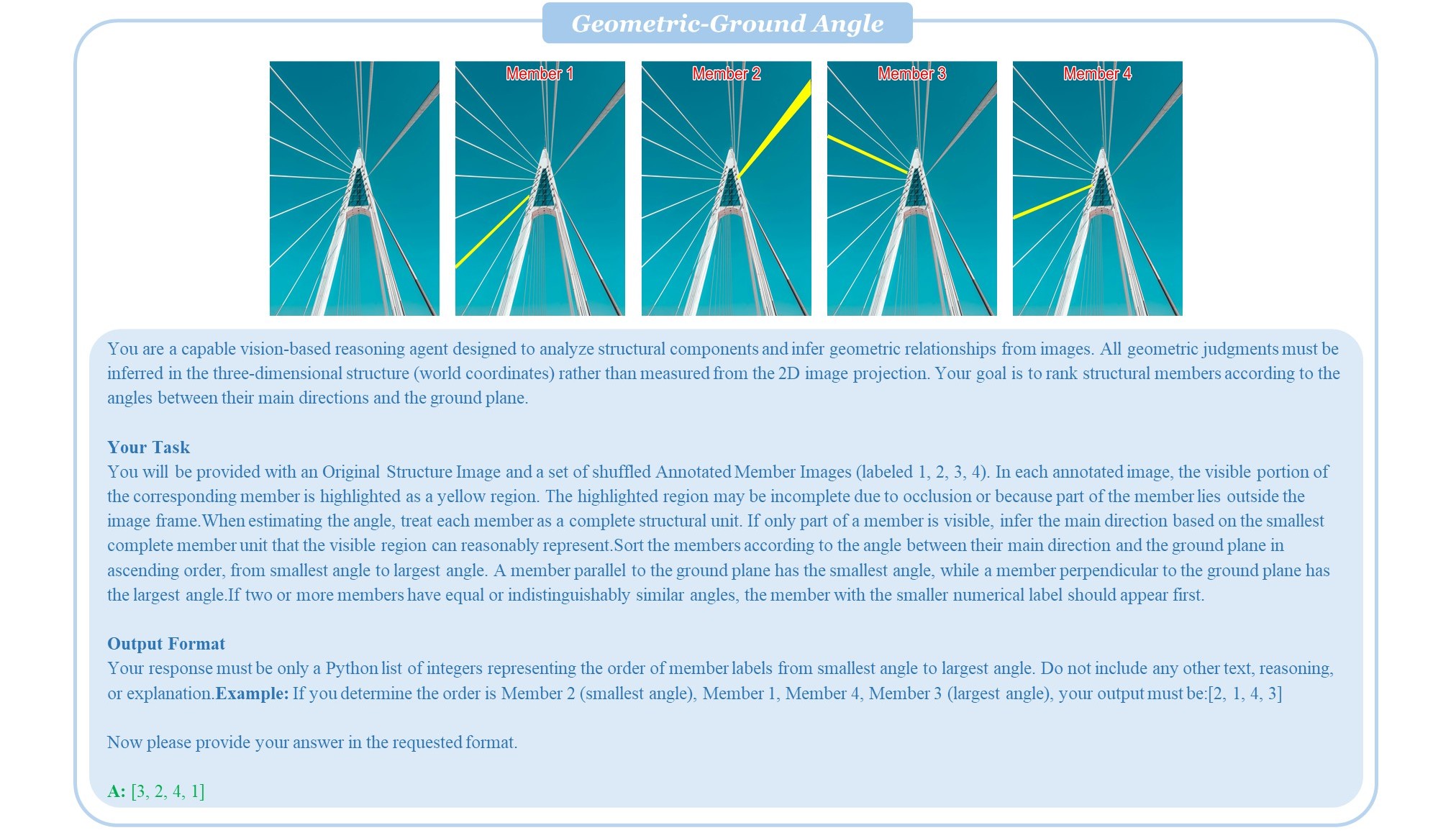}
  \caption{Full example for geometric-ground angle category.}
  \label{app_fig:sample7}
\end{figure*}

\begin{figure*}
  \centering
  \includegraphics[width=\textwidth]{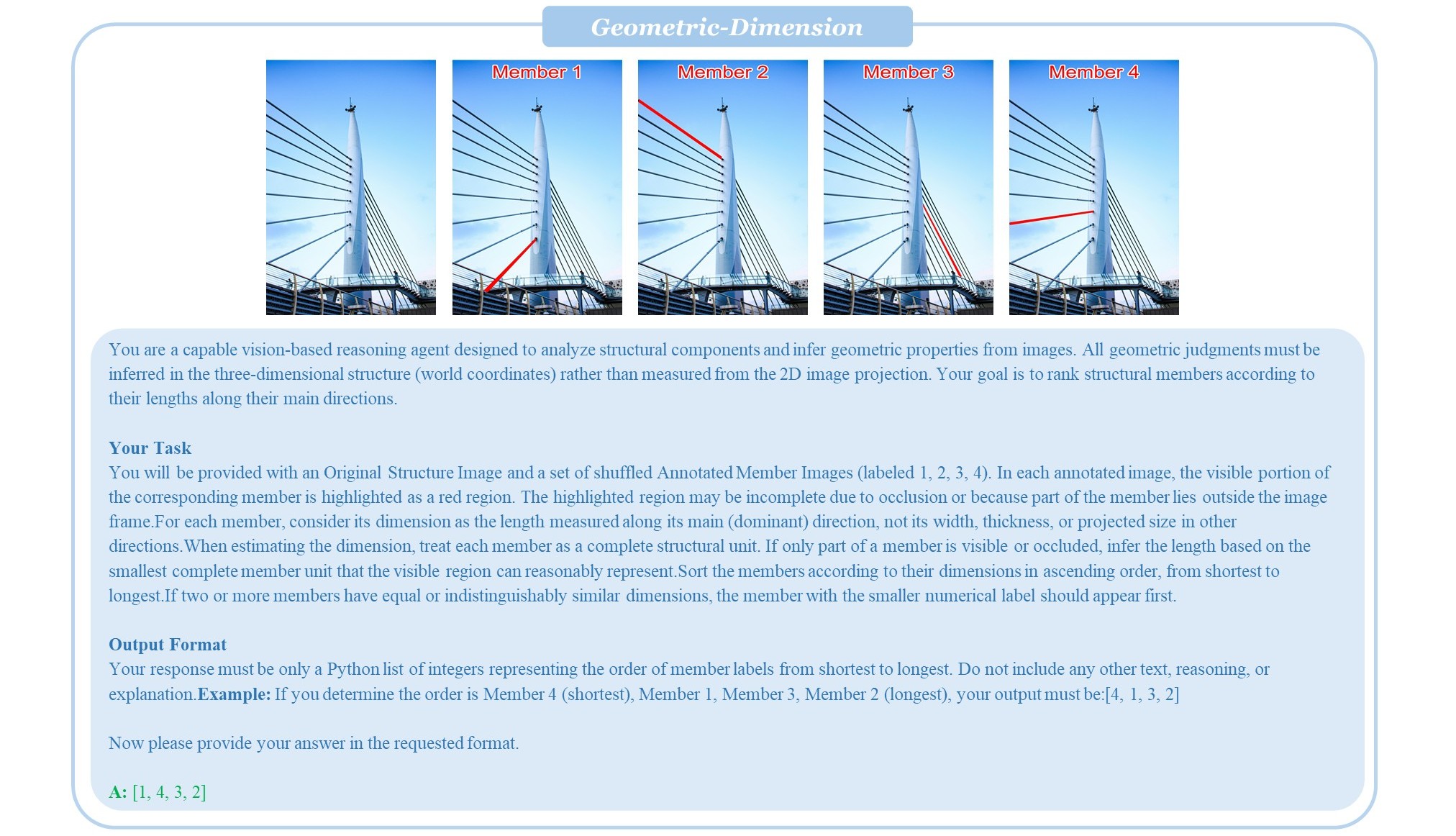}
  \caption{Full example for geometric-dimension category.}
  \label{app_fig:sample6}
\end{figure*}

\begin{figure*}
  \centering
  \includegraphics[width=\textwidth]{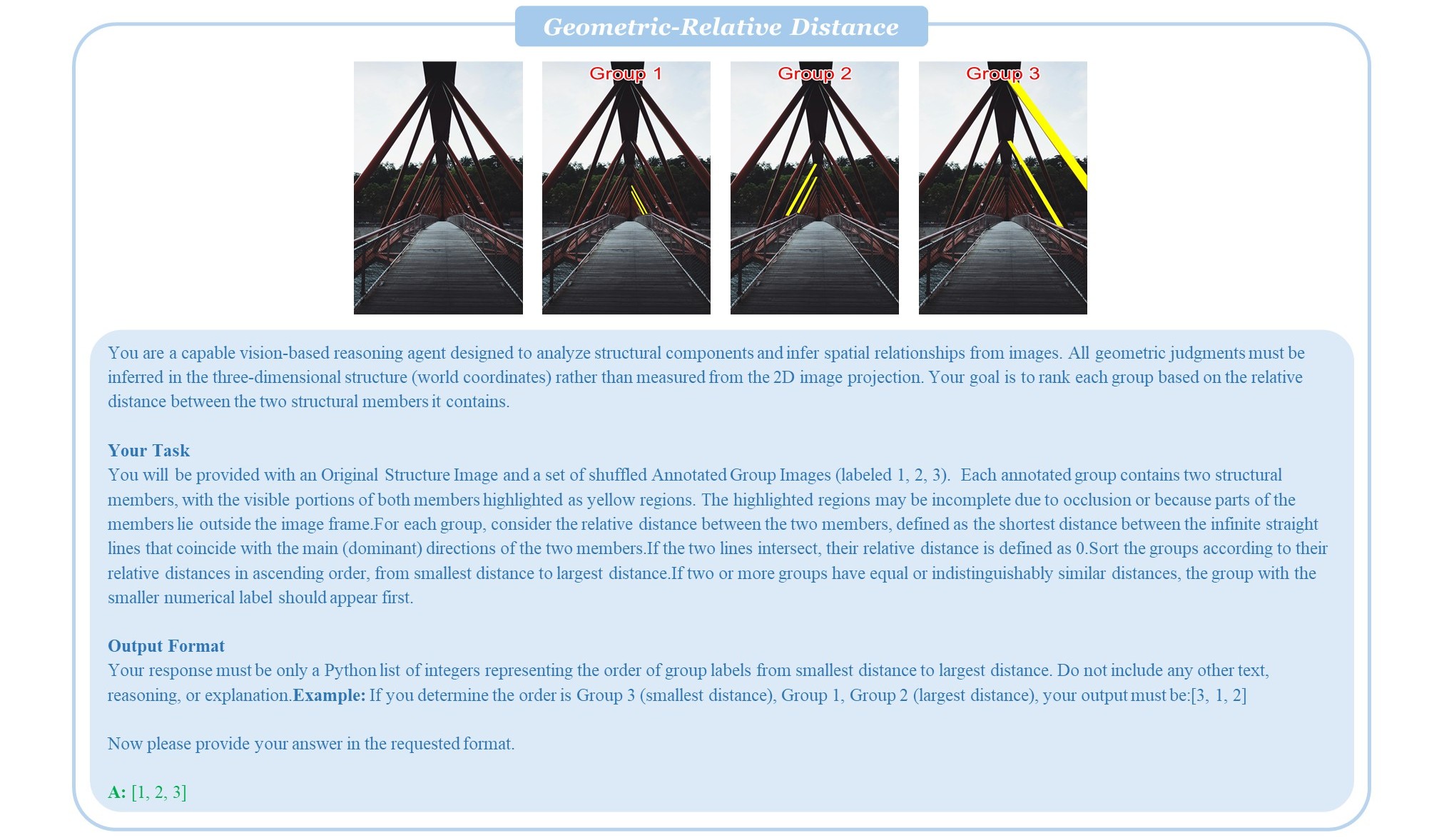}
  \caption{Full example for geometric-relative distance category.}
  \label{app_fig:sample5}
\end{figure*}

\begin{figure*}
  \centering
  \includegraphics[width=\textwidth]{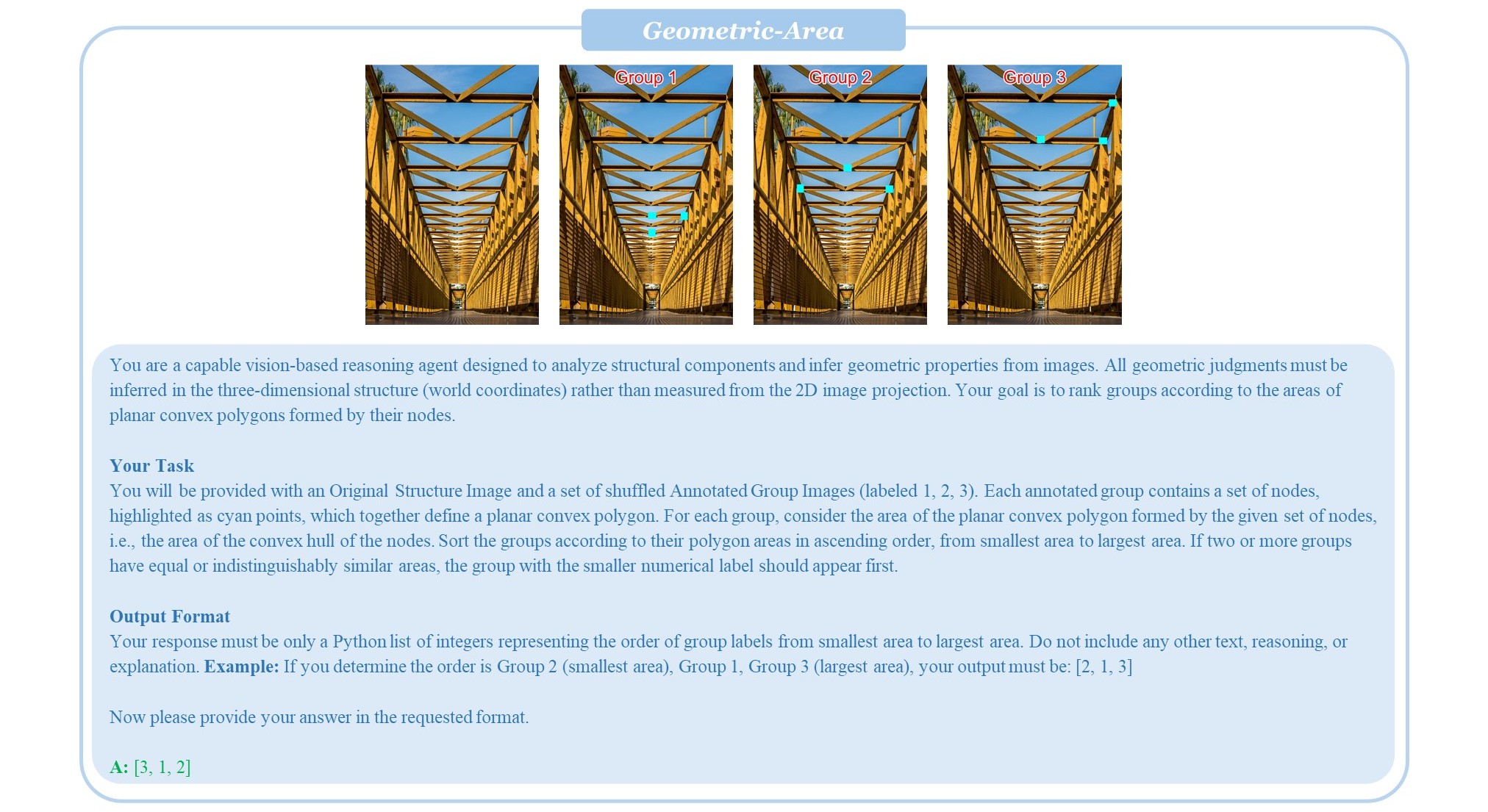}
  \caption{Full example for geometric-area category.}
  \label{app_fig:sample1}
\end{figure*}

\begin{figure*}
  \centering
  \includegraphics[width=\textwidth]{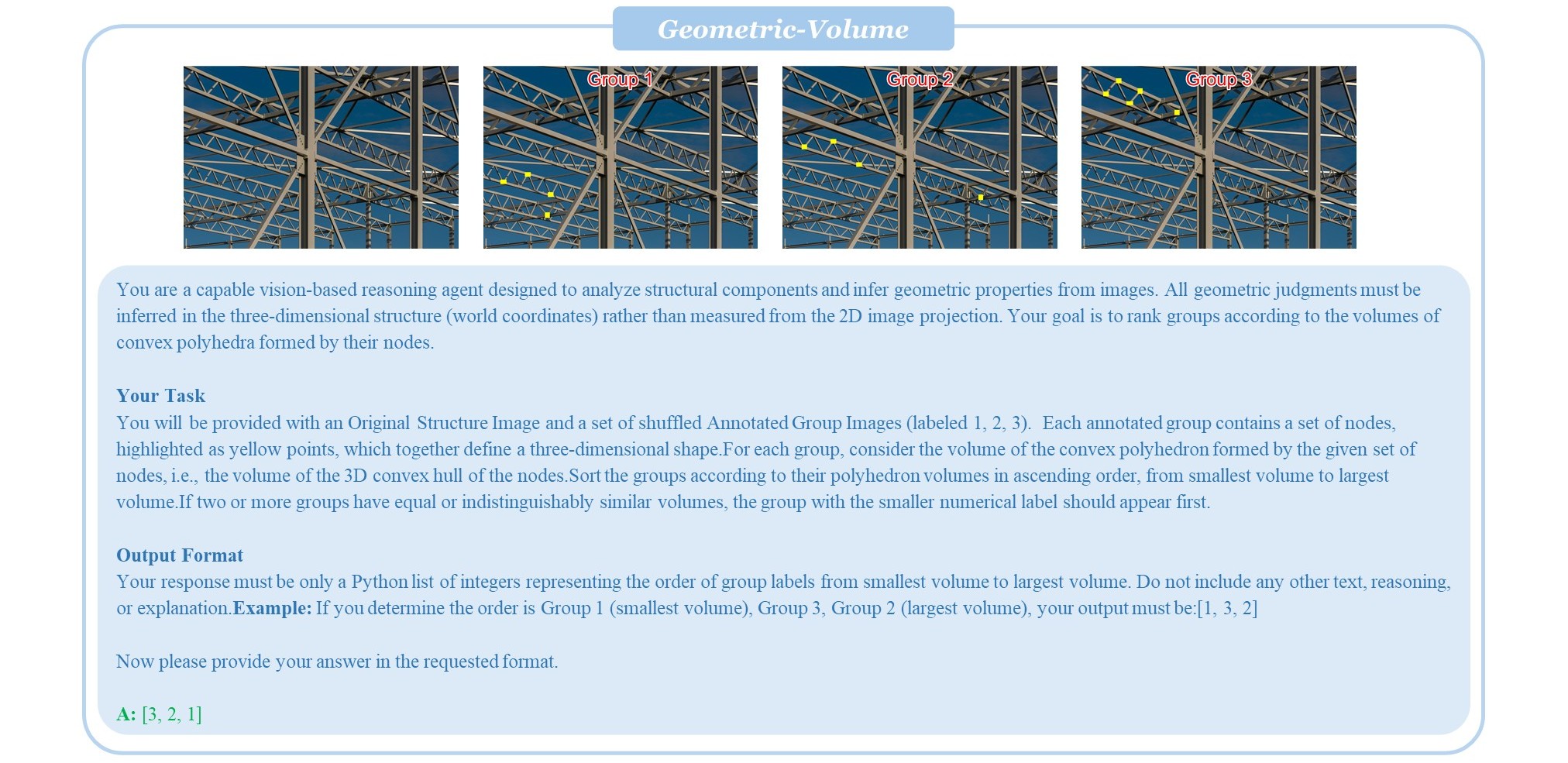}
  \caption{Full example for geometric-volume category.}
  \label{app_fig:sample3}
\end{figure*}

\begin{figure*}
  \centering
  \includegraphics[width=\textwidth]{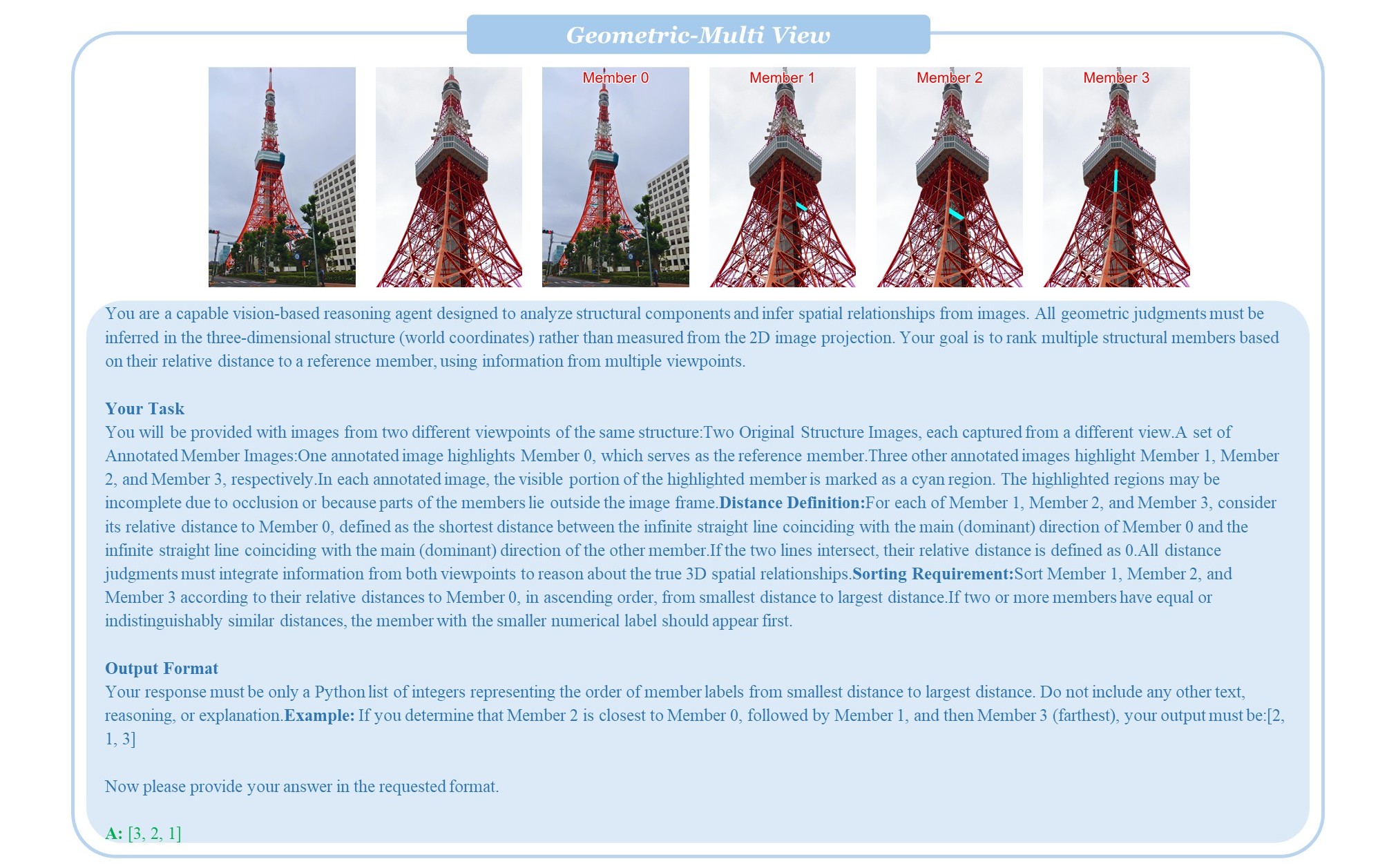}
  \caption{Full example for geometric-multi view category.}
  \label{app_fig:sample2}
\end{figure*}

\begin{figure*}
  \centering
  \includegraphics[width=\textwidth]{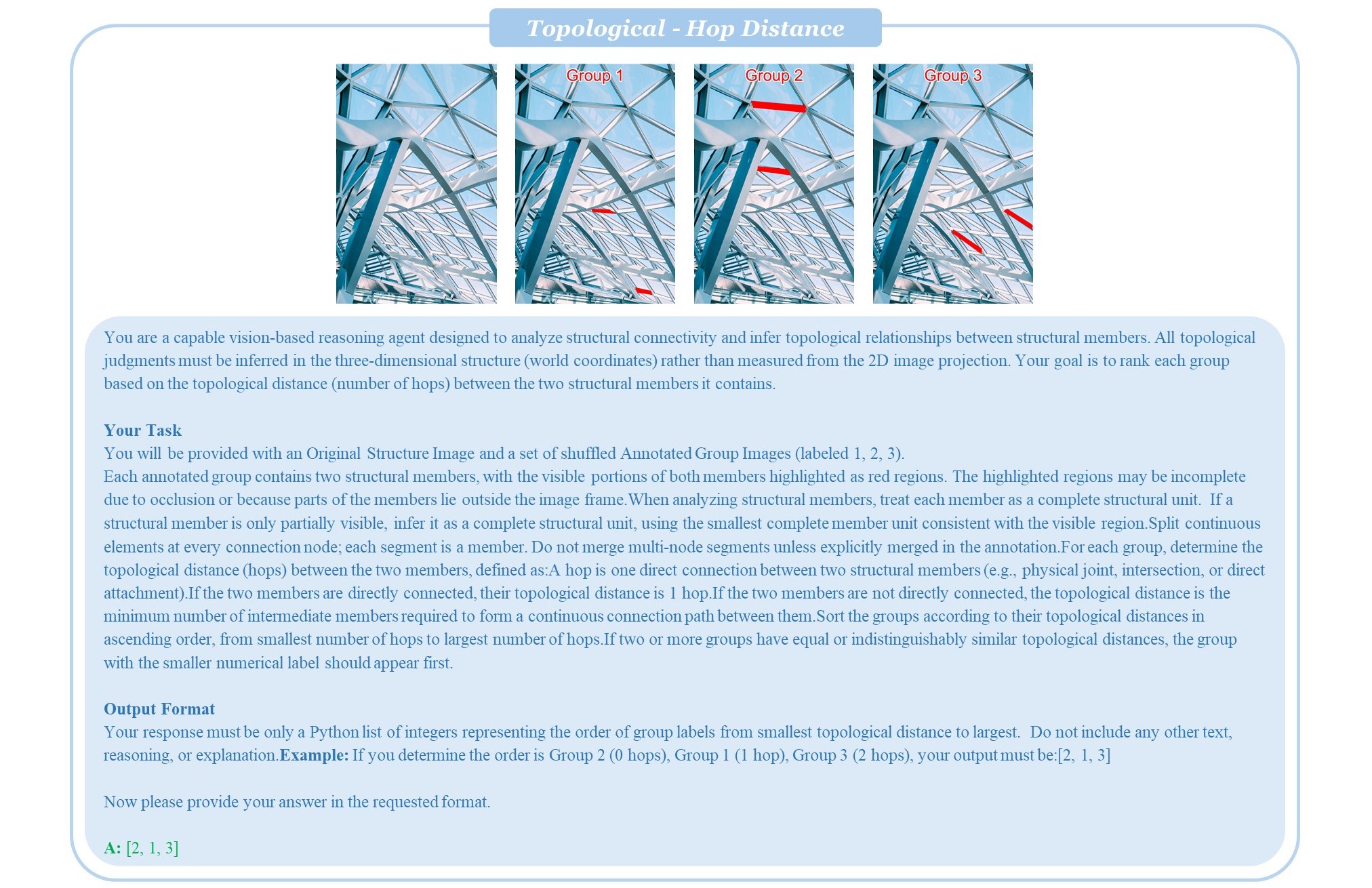}
  \caption{Full example for topological-hop distance category.}
  \label{app_fig:sample10}
\end{figure*}

\begin{figure*}
  \centering
  \includegraphics[width=\textwidth]{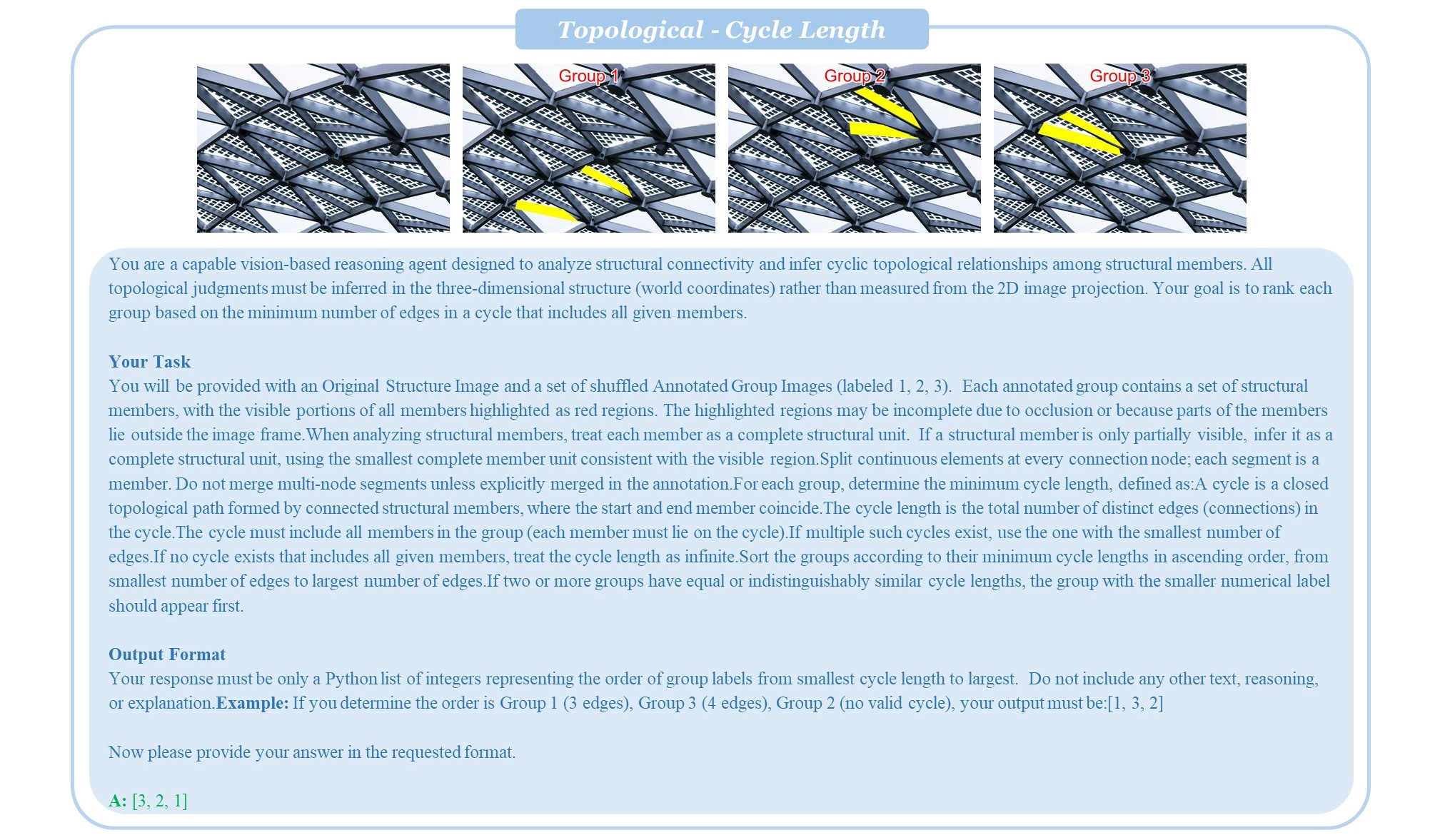}
  \caption{Full example for topological-cycle length category.}
  \label{app_fig:sample9}
\end{figure*}

\begin{figure*}
  \centering
  \includegraphics[width=0.85\textwidth]{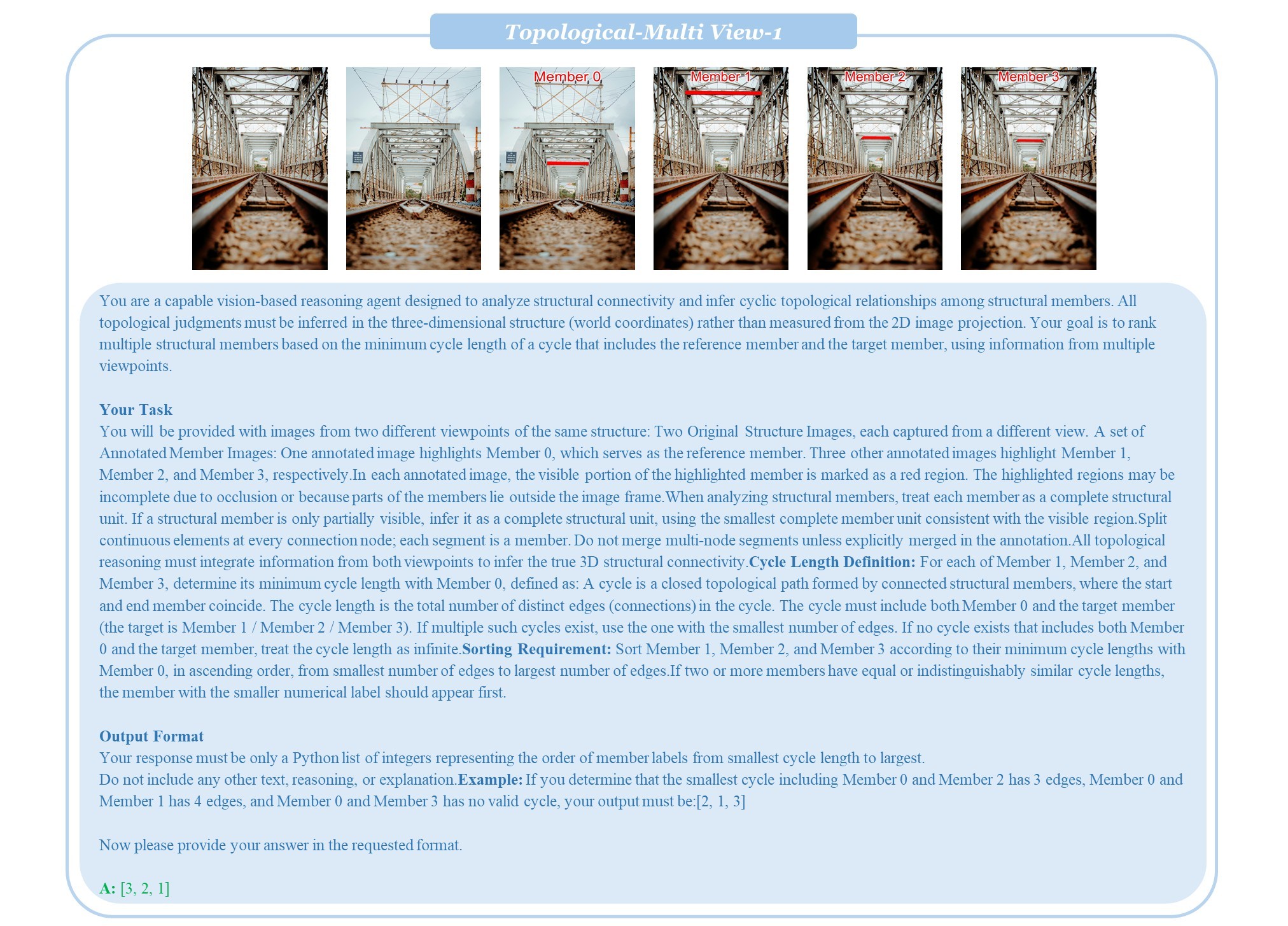}
  \caption{Full example for topological-multi view-1 category.}
  \label{app_fig:sample4}
\end{figure*}

\begin{figure*}
  \centering
  \includegraphics[width=0.85\textwidth]{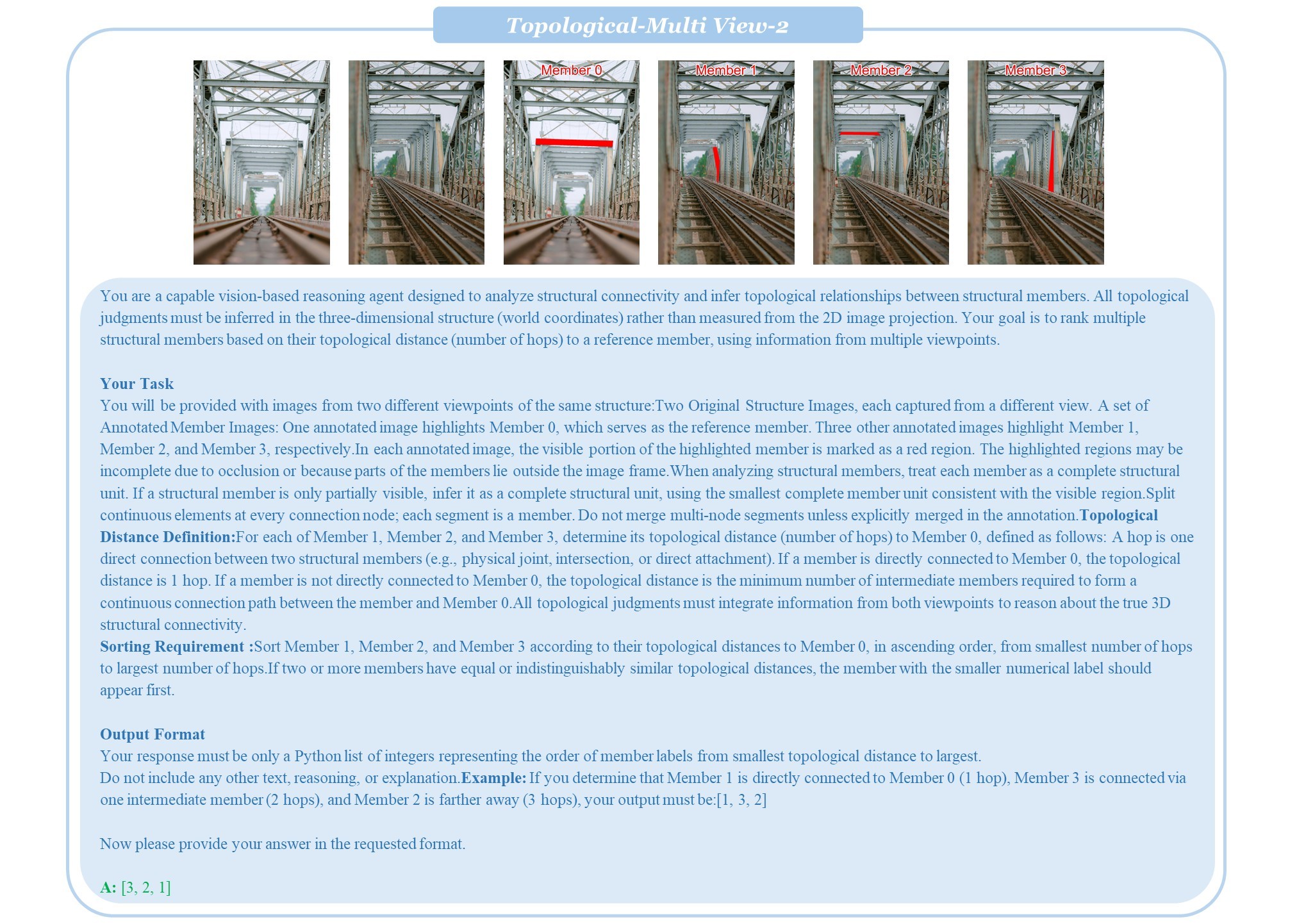}
  \caption{Full example for topological-multi view-2 category.}
  \label{app_fig:sample11}
\end{figure*}

For clarity and readability, the questions and rationales shown in the main text are simplified; this section provides the complete versions for each category. We include one full example for each of the 11 sub-categories: Ground Height, Ground Angle, Dimension, Relative Distance, Area, Volume, Multi-View (Geo.), Hop Distance, Cycle Length, and Multi-View (Topo.). Note that Multi-View (Geo.) is the multi-view variant of Relative Distance. In contrast, Multi-View (Topo.) comprises two prompt types, corresponding to the multi-view variants of Hop Distance and Cycle Length, respectively. These representative samples are illustrated in Figure~\ref{app_fig:sample8}--Figure~\ref{app_fig:sample11}.


\section{Case Studies}
\label{app:additional_case}
\begin{figure*}
  \centering
  \includegraphics[width=0.9\textwidth]{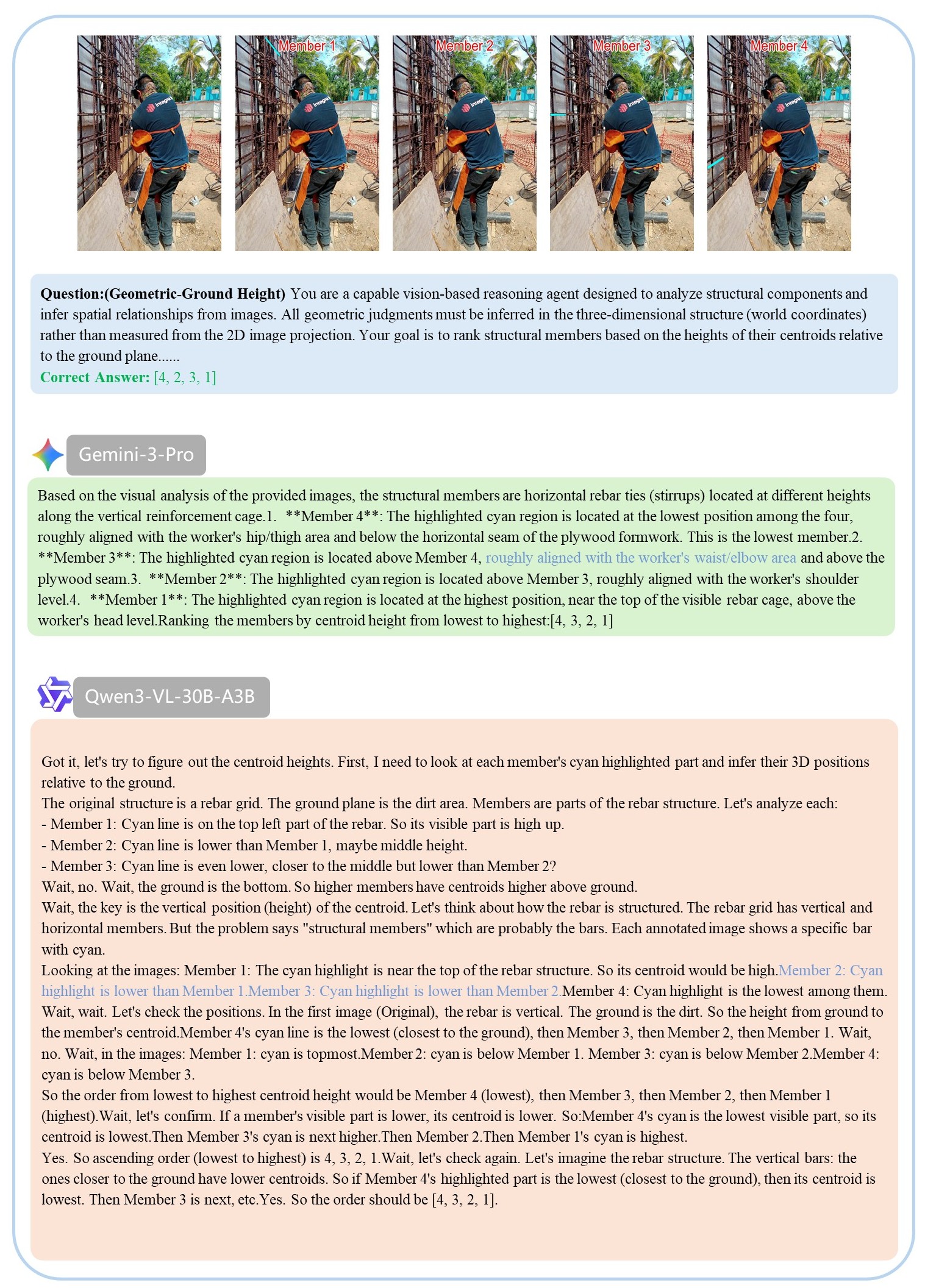}
  \caption{Illustration of 3D spatial-logic errors (highlighted in blue).}
  \label{app_fig:answer1}
\end{figure*}

\begin{figure*}
  \centering
  \includegraphics[width=0.9\textwidth]{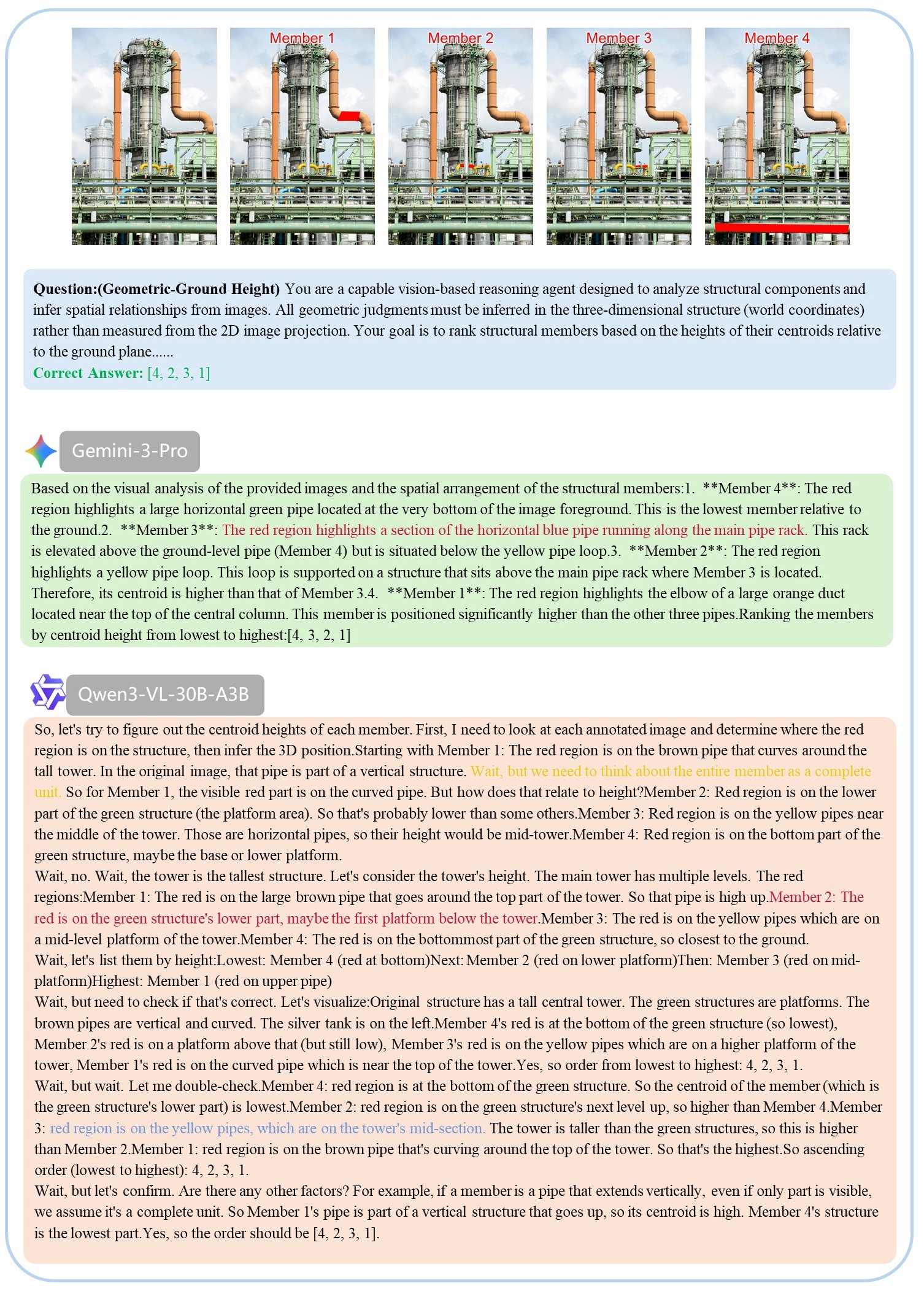}
  \caption{Illustration of 3D spatial-logic errors (highlighted in blue), Member-extent errors (highlighted in orange), Object-recognition errors (highlighted in red).}
  \label{app_fig:answer2}
\end{figure*}

\begin{figure*}
  \centering
  \includegraphics[width=0.9\textwidth]{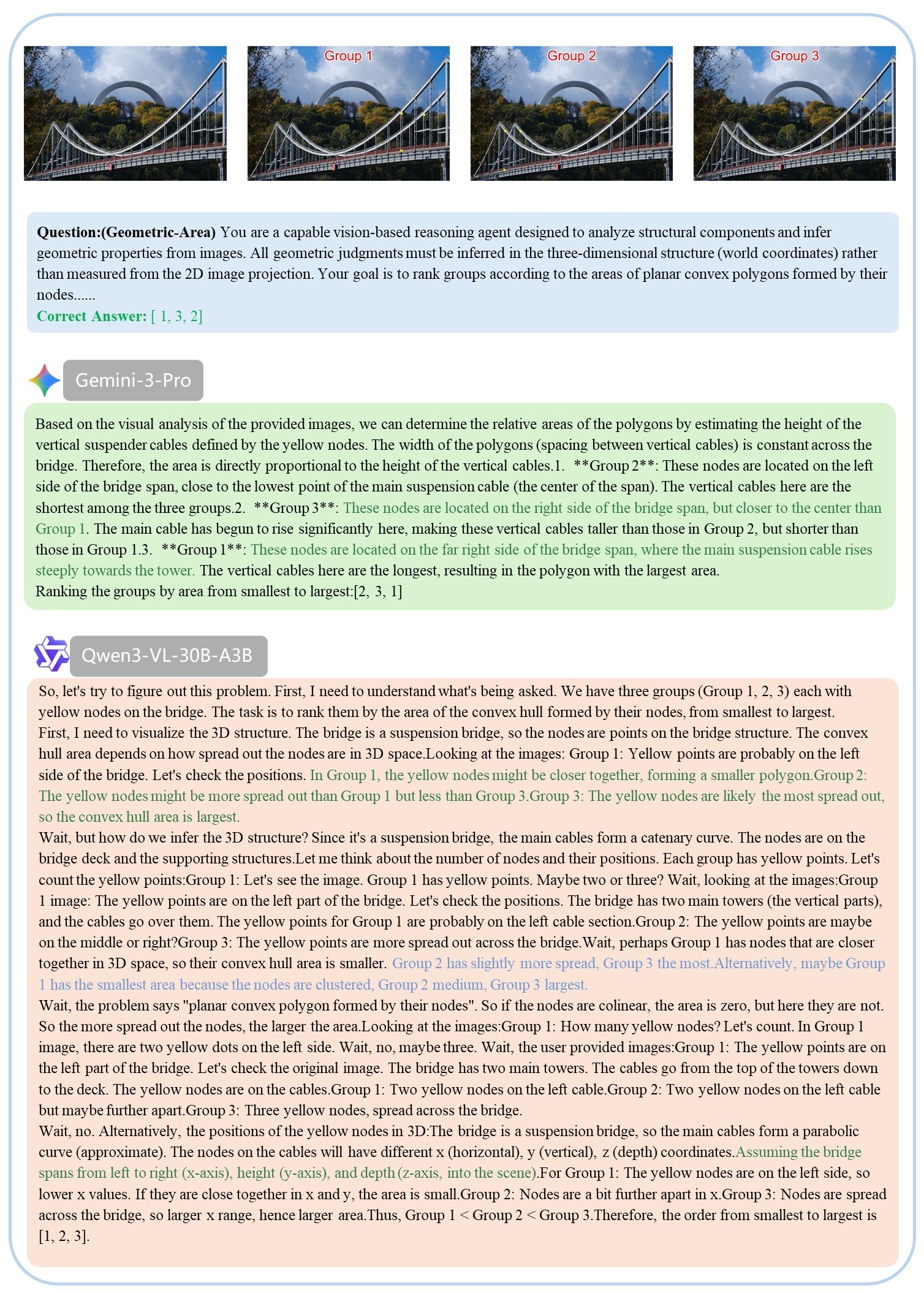}
  \caption{Illustration of 3D spatial-logic errors (highlighted in blue), Computational and comparison-logic errors (highlighted in green).}
  \label{app_fig:answer3}
\end{figure*}

In this section, we present additional complete reasoning processes of current VLMs to illustrate their spatial reasoning deficiencies more intuitively. As shown in Figure~\ref{app_fig:answer1}, Figure~\ref{app_fig:answer2}, and Figure~\ref{app_fig:answer3}, we highlight four types of errors that identify dominant model bottlenecks: Member-Extent Errors(orange), Object-Recognition Errors(red), Computational and Comparison-Logic Errors(green), and 3D Spatial-Logic Errors(blue). 

\section{Limitations}
\label{app:limitations}

To ensure high data quality, clear supervision, and reduced shortcut cues, we adopted a fully human-centered pipeline to curate images and manually annotate question--answer pairs (including rankings and ties). This design choice inevitably limits scalability: constructing and verifying instances requires substantial expert time and careful quality control, making it less scalable than fully automated generation pipelines. Nevertheless, we believe the current benchmark scale (1,000 samples) is sufficient for meaningful assessment at the present stage, since today’s VLMs still exhibit a large gap to human performance on \ssibench and do not appear close to saturation. As models improve and performance begins to saturate, expanding the benchmark---potentially with more diverse structures and harder cases---may become necessary, but reaching that point will likely require either further research breakthroughs or additional human annotation effort. Importantly, the metadata annotation and question generation stages also support semi-automatic expansion: once richer candidate annotations are available for an image, new ranking questions can be generated by sampling candidate combinations and then verified by humans, as discussed in Appendix~\ref{app:semi_auto_expansion}. This reduces the marginal cost of scaling, although human review remains necessary to preserve answer uniqueness, visual clarity, and resistance to superficial shortcut cues. For test-time evaluation, we prioritize data quality and reliability, which is why we opted for manual curation over fully automated pipelines.


\end{document}